\documentclass[]{xiaomiev}

\usepackage[toc,page,header]{appendix}
\usepackage{minitoc}
\usepackage{bbding}
\usepackage{solarized-light}
\usepackage{amsfonts}
\usepackage{booktabs}
\usepackage{multirow}
\usepackage{graphicx}
\usepackage{array}
\usepackage{siunitx,array}
\usepackage[table]{xcolor}
\usepackage{makecell,array,booktabs}
\usepackage{makecell}
\usepackage{framed}
\usepackage{amsthm}
\usepackage{bbding}
\usepackage{hyperref}
\usepackage{amsmath} 
\usepackage{placeins}
\usepackage[colorinlistoftodos]{todonotes}
\usepackage{longtable}
\usepackage{hhline}
\usepackage{fancyvrb}
\usepackage{graphicx}
\usepackage[table]{xcolor}
\usepackage{booktabs}
\usepackage{float}
\usepackage{fvextra}
\usepackage{CJKutf8}
\usepackage{multicol}
\usepackage{float}
\usepackage{placeins}
\usepackage{cleveref}
\usepackage{tablefootnote}
\usepackage{threeparttable}
\usepackage{tabularx}
\usepackage{mdframed}
\usepackage{subcaption}
\usepackage[usestackEOL]{stackengine}
\usepackage[numbers]{natbib}
\newcommand{\commentout}[1]{}
\renewcommand{\paragraph}[1]{\noindent\textbf{#1.}\hspace*{1em}}
\usepackage{enumitem}
\usepackage{hyperref}
\setlist[itemize]{leftmargin=15pt}
\usepackage{siunitx,array}

\usepackage{booktabs}
\usepackage{pifont}
\usepackage{amsmath}

\usepackage{wrapfig}

\usepackage{caption}
\usepackage{graphicx}

\RequirePackage{xspace}
\makeatletter
\DeclareRobustCommand\onedot{\futurelet\@let@token\@onedot}
\def\@onedot{\ifx\@let@token.\else.\null\fi\xspace}

\makeatother
\newcommand{\XFM}{Discrete-WAM}

\title{Discrete-WAM: Unified Discrete Vision-Action Token Editing for World-Policy Learning
}

\author{Xiaomi EV}
\vspace{-11pt}

\contribution{See \hyperref[sec:contributions]{Contributions and Acknowledgments} section for a full author list.}

\abstract{

Autonomous driving requires reasoning about how ego actions shape future world evolution, rather than merely mapping observations to actions. 
However, most end-to-end methods rely on direct state-to-action imitation, while existing world models often remain weakly aligned with downstream policy generation. 
We introduce \textbf{\XFM{}}, a unified discrete vision-action world-policy framework that represents visual observations, future states, high-level decisions, and ego actions within a shared token space. 
Built on this discrete alignment, \XFM{} jointly trains world modeling, world-policy modeling, and policy modeling through multi-task and multi-stage pretraining, allowing action-conditioned future prediction to directly support policy generation. 
For downstream planning, \XFM{} further decomposes policy generation into hierarchical decision prediction and parallel action-token editing, where the decision token provides a high-level planning skeleton and confidence-based scheduling refines dense future actions efficiently. 
Experiments on large-scale autonomous-driving benchmarks show that \XFM{} achieves strong planning performance while supporting controllable future generation, counterfactual evaluation, surprise-based world-model analysis, and efficient parallel policy decoding. 
These results suggest that discrete representation alignment, unified world-policy training, and hierarchical token editing provide a promising design paradigm for physical AI.

}

\begin{document}
\maketitle

\begin{figure*}[!b]
    \centering
    \vspace{-30pt}
    \centering
    \includegraphics[width=\linewidth]{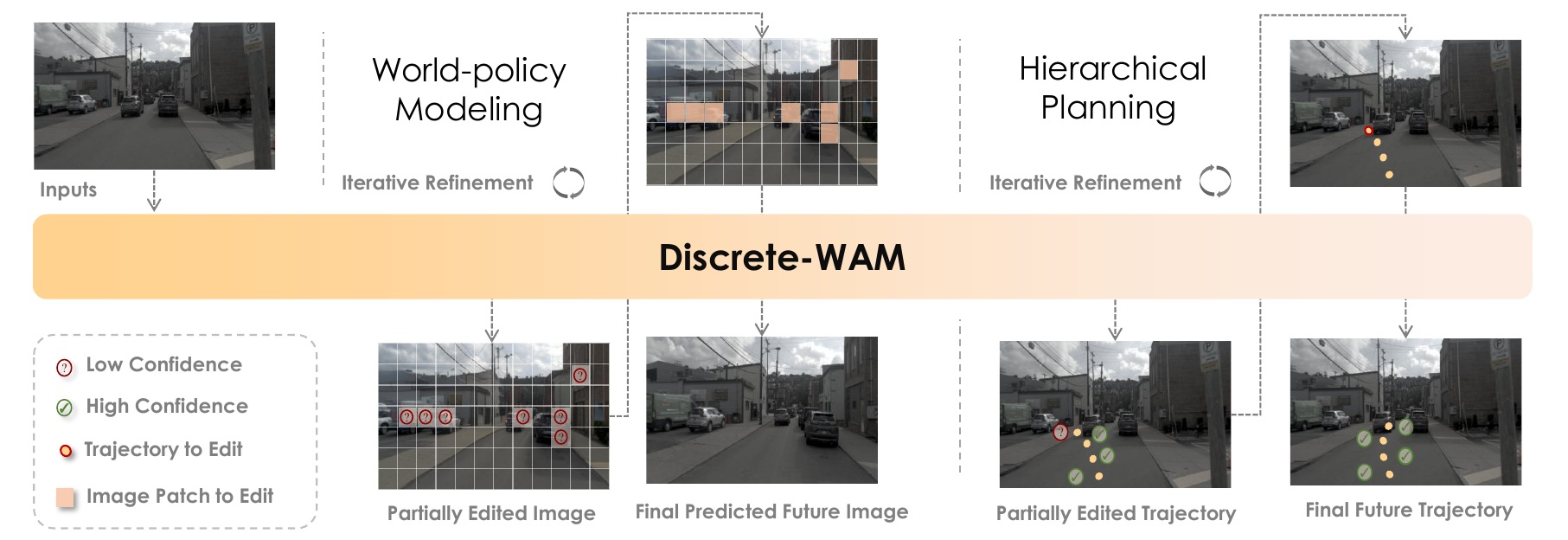}
    \caption{\textbf{Overview of Discrete-WAM.} Discrete-WAM jointly edits visual, decision, and action tokens in a unified discrete space, offering editable generation of future observations and planning trajectories through unified pretraining and reward-guided post-training.
    }
    \label{fig:overview}
\end{figure*}



\section{Introduction}

Autonomous driving is a representative physical AI problem: an agent must reason about how its actions shape future world evolution, rather than merely react to instantaneous observations~\cite{chen2022milestones,kong20253d,lecun2022path}. 
Existing end-to-end (E2E) autonomous driving systems~\cite{chen2024vadv2,hu2023planning,li2024hydra,liao2025diffusiondrive,liu2025hybrid} often formulate driving as direct vision-to-action mapping via behavior cloning~\cite{chen2024end}, capturing statistical correlations without explicitly modeling action-conditioned dynamics. 
Prediction-augmented planning introduces future reasoning~\cite{hagedorn2024integration}, but many methods still rely on predefined intermediate annotations or task-specific representations~\cite{chen2024ppad,huang2023gameformer,wei2024occllama}, which limits generalization across diverse scenes and counterfactual futures. 
Recent vision-language-action (VLA) models~\cite{li2025drivevla,li2025recogdrive,xu2024drivegpt4,zhou2025autovla} incorporate broader semantic priors through language supervision, yet their reasoning capability is still constrained by annotations, providing a low-bandwidth interface for fine-grained spatial-temporal dynamics~\cite{jiang2025survey}. 
These suggest that reliable physical AI requires unified formulation, where world policy prediction and generation are aligned within the same representation and optimization space.

A key bottleneck is representation alignment. 
Many autonomous driving systems operate in continuous latent spaces~\cite{hu2023gaia,lin2025futurex,zeng2026latent}, where representations provide smooth interpolation but remain highly entangled and lack explicit compositional semantics~\cite{locatello2020object}. 
As a result, observations, actions, and future states are only weakly aligned, making it difficult to perform reliable action-conditioned reasoning or compare alternative futures. 
Discrete representations provide compositional semantic units that can serve as shared anchors across vision, action, and future evolution~\cite{li2025discrete,ma2025dvlm,ye2025dap}. 
More importantly, a shared discrete token space enables observations and actions sequences to be optimized with a unified sequence modeling interface, shared framework, and token-level objectives for tighter multi-modal alignment. 
For action tokens, however, deterministic discretization error are inevitable under hard quantization. 


Beyond representation learning, physical AI requires world modeling to directly support policy optimization. Predicting future observations is valuable not for reconstruction itself, but for evaluating the consequences of candidate actions. Recent generative driving models have demonstrated strong capabilities in future generation and action-conditioned world modeling~\cite{yang2024genad,azzolini2025cosmos,liao2025diffusiondrive,zhang2025epona}, while autoregressive formulations introduce discrete compositional generation for unified world-action prediction~\cite{bartoccioni2025vavim,chen2026last,zhou2025autovla}. However, future prediction and policy learning are still commonly optimized as separate modules or auxiliary objectives~\cite{li2025drivevla,li2025end,wang2026drive,huang2026automot,huang2026mindvla,li2026unidrivevla}, creating a gap between understanding future dynamics and selecting optimal behaviors. Moreover, driving policies are inherently hierarchical: high-level decisions determine the feasible space of low-level trajectories, while route constraints, interactions, and vehicle dynamics induce strong dependencies among future action tokens. As a result, fully parallel action generation can suffer from dependency mismatch, motivating a unified formulation that jointly learns world dynamics and hierarchical policy construction.

Taken together, these observations suggest that physical AI requires a unified world-policy formulation rather than isolated advances in prediction or control. We identify three key properties: (1) representation alignment across observations, actions, and futures; (2) world-policy alignment between future prediction and action generation; (3) and hierarchical policy construction linking high-level decisions with low-level actions. Based on this perspective, we propose \textbf{\XFM{}}, a unified discrete vision-action world-policy framework for autonomous driving. \XFM{} formulates driving as sequence modeling over a shared discrete token space, where visual observations, ego actions, future states, and decision variables are jointly represented and modeled through a unified generative process. Rather than treating actions merely as conditioning signals for future prediction, \XFM{} models observations, actions, decisions, and future evolution as coupled variables, enabling bidirectional interaction between world dynamics and policy learning. To address representation alignment, we introduce a unified discrete representation together with soft-label action interpolation, reducing deterministic discretization error for continuous control. To bridge the gap between future prediction and policy optimization, \XFM{} jointly learns world modeling, world-policy modeling, and policy generation within a shared Transformer architecture, allowing future prediction and action generation to mutually benefit from a common training objective. To account for the hierarchical structure of driving behavior, \XFM{} further decomposes policy generation into high-level decision prediction and low-level action-token refinement, enabling efficient parallel generation while preserving decision consistency and long-horizon behavioral structure. Training and inference are designed accordingly through progressive world-policy alignment and confidence-guided parallel refinement, providing an efficient framework for both future reasoning and policy generation. Overall, \XFM{} provides a unified framework for representation alignment, world-policy learning, and hierarchical decision making, advancing physical AI from reactive imitation toward decision-oriented world modeling.

Our contributions are summarized as follows:
\begin{itemize}
\item We propose a multi-task world-policy framework that jointly models world dynamics, world-policy, and policy generation within a unified training paradigm. The framework is enabled by a shared discrete token space that aligns observations, decisions and actions across temporal horizons.

\item We formulate planning as hierarchical decision-conditioned action-token editing and develop post-training strategies with confidence-guided parallel editing, improving the suitability of parallel discrete diffusion for autonomous driving.

\item Comprehensive evaluations across planning benchmarks, world-model surprise analysis, attention-based grounding, scheduling dynamics, and inference-efficiency studies validates the effectiveness of the proposed physical-AI design paradigm.

\end{itemize}

\section{Architecture}
\label{sec:arch}

\begin{figure*}[t]
    \centering
    \includegraphics[width=1.0\linewidth, trim={0cm 8cm 6cm 0cm},clip]{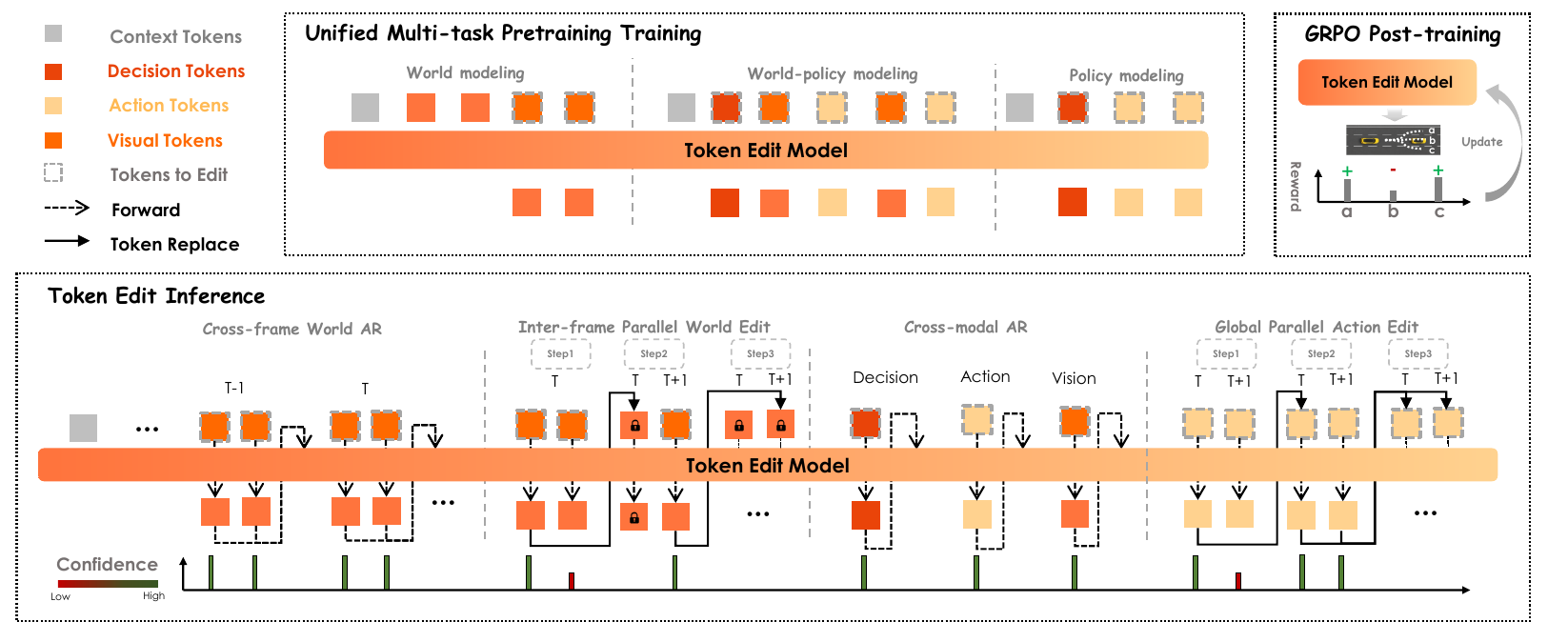}
    \caption{\textbf{Model architecture of \XFM{}.}
    \XFM{} is a unified vision-action world-policy model for autonomous driving. The architecture converts camera observations into discrete visual tokens, represents future ego motion with discrete action tokens, and injects ego-state, navigation, and high-level decision information as conditioning tokens. Built on a shared Transformer backbone, \XFM{} supports three complementary training modes: world modeling for action-conditioned vision prediction, world-policy modeling for joint action and future-vision prediction, and policy modeling for decision-conditioned action generation. This unified token interface enables visual observations, driving decisions, and future actions to be modeled and edited within the same discrete sequence space.
    }
    \label{main_structure}
\end{figure*}

\subsection{Base architecture of \XFM{}}
\label{sec2.1}

The \XFM{} architecture consists of four main components: (1) a vision VQ Tokenizer that encodes visual observations into discrete semantic tokens and a projector that align vision feature with transformer hidden dimension; (2) a context encoder that injects ego-state and navigation commands into the latent sequences; (3) a decoder-only Transformer backbone that jointly models observations, actions, and future evolution within a unified token space; and (4) multi-task prediction heads for world modeling, policy generation, and world-action sequence generation.

\subsection{Vision Tokenization}

The vision tokenizer converts continuous camera observations into compact discrete visual tokens, providing a token-level interface between raw images and the Transformer backbone. This discrete representation enables visual observations to be handled in the same sequence format as action tokens, while preserving the scene semantics required for downstream world and policy modeling. To obtain compact visual representations, we follow the previous work~\cite{yao2026unifieddrivingtokensrepresentation} and pretrain a VQ-VAE-based tokenizer~\cite{vanDenOord2017vqvae} to encode front-view camera images into discrete visual tokens.

\subsection{Action Tokenization}
\label{sec:action_tokenization}

For action representation, we convert continuous future motion into a discrete-token-compatible representation over an acceleration vocabulary. Given a future trajectory over ${H}$ time steps, we first fit the discrete trajectory with a cubic spline to obtain a smooth continuous curve. We then compute the ego-centric 2D acceleration at each future step using second-order finite differences, denoted as $(a_x, a_y)$, where $a_x$ and $a_y$ correspond to longitudinal and lateral acceleration, respectively. The spline fitting step improves the continuity of the derived acceleration sequence and ensures that integrating the recovered accelerations can closely match the original discrete trajectory.

We construct a uniformly distributed 2D acceleration vocabulary by independently partitioning the valid ranges of $a_x$ and $a_y$ into $N_x$ and $N_y$ bins. The Cartesian product of these bins forms a grid-structured action vocabulary with size $N_x \times N_y$. Instead of assigning each continuous acceleration vector to a single nearest bin, which would introduce deterministic hard-quantization error, we represent it with a soft target over its neighboring vocabulary entries. Specifically, for each acceleration component, we find the two adjacent bin centers that bracket the continuous value and assign interpolation weights according to its relative position between them. In the 2D acceleration vocabulary, this produces a soft label over the four neighboring prototypes around $(a_x,a_y)$. Under this grid interpolation, the continuous acceleration admits an exact interpolation representation within each acceleration grid cell.

During training, the action head is optimized with cross-entropy against this soft target distribution rather than a one-hot label. At inference time, the predicted action distribution can be mapped back to continuous acceleration by taking the weighted sum over the acceleration vocabulary. Under exact recovery of the soft target distribution, the continuous acceleration can be reconstructed exactly as the weighted sum of the neighboring vocabulary prototypes. Therefore, the proposed soft-label representation removes deterministic hard-assignment quantization error, while the remaining reconstruction error is attributed to distribution prediction mismatch. The corresponding derivation and error bound are provided in Appendix.

As a result, a future action sequence is represented as $\mathbf{A}_{t+1:t+H}$ over the discrete acceleration vocabulary, while continuous control values are preserved through soft-label interpolation. This action tokenization enables visual tokens and action tokens to be modeled jointly in the same Transformer token space while avoiding the deterministic error introduced by hard quantization of continuous actions.

\subsection{Unified World Policy}
\label{sec:unified_world_policy}

\XFM{} formulates autonomous driving as a unified world-policy modeling problem over discrete visual, decision, and action tokens. At time step $t$, we denote the scene context as $\mathbf{C}_t$, which contains historical visual observations, ego-state information, and navigation commands. The future visual observations are represented as discrete visual token sequences $\mathbf{V}_{t+1:t+H}$, and the future action policy is represented as discrete action token sequences $\mathbf{A}_{t+1:t+H}$, where each action token corresponds to a quantized acceleration prototype defined in Sec.~\ref{sec:action_tokenization}. We further denote the high-level decision condition as $\mathbf{D}_t$, which captures sparse low-frequency driving structure, such as maneuver intent, target lane, coarse waypoint, speed trend, or interaction priority.

Under this notation, \XFM{} integrates three training modes under a shared token-editing interface. These modes differ in which token streams are used as conditioning inputs and which token streams are treated as prediction targets.

The first mode is \textbf{world modeling}, whose objective is vision prediction. Given the scene context $\mathbf{C}_t$ and future action tokens $\mathbf{A}_{t+1:t+H}$, the model predicts future visual tokens $\mathbf{V}_{t+1:t+H}$. This task trains the model to understand how the scene evolves under a specified action sequence, and provides action-conditioned world dynamics for downstream policy learning.

    The second mode is \textbf{world-policy modeling}, which jointly trains vision prediction and action prediction. In this mode, the model reasons about future actions and their induced visual consequences within the same token sequence. Action tokens represent the future policy, while visual tokens represent the corresponding future world states. This formulation encourages the model to couple policy generation with world evolution, rather than learning them as independent objectives.

The third mode is \textbf{policy modeling}, whose objective is decision prediction followed by action prediction. The model first predicts the high-level decision condition $\mathbf{D}_t$ from the scene context $\mathbf{C}_t$. Conditioned on this decision, the action-token planner then predicts the future action sequence $\mathbf{A}_{t+1:t+H}$. This hierarchical decomposition separates policy learning into two levels: the high-level decision task captures multi-modal driving choices, while the low-level action prediction task focuses on generating smooth and temporally consistent trajectories conditioned on the selected decision.

Together, these three modes provide a unified architecture for action-conditioned world prediction, joint world-policy learning, and decision-conditioned action generation. The mathematical task formulations are described in Sec.~\ref{sec:method}.

\section{Method}
\label{sec:method}


\subsection{Unified Pretraining}
\label{sec:unified_pretraining}
Following the unified world-policy formulation in Sec.~\ref{sec:unified_world_policy}, pretraining instantiates the conditional token-editing objective with multiple task families. Each task uses the same discrete visual and action token interface, but differs in which token streams are treated as conditioning inputs and which corrupted tokens are supervised as editing targets. This design allows world modeling, policy prediction, and joint world-policy modeling to share one training framework while preserving their task-specific conditioning structure.

\paragraph{Training tasks}
We instantiate unified pretraining with three task families: world modeling, policy modeling, and joint world-policy modeling.

\paragraph{Tokenization}
We tokenize both visual observations and future actions into discrete token sequences so that they can be jointly modeled within a unified Transformer architecture.

For image tokenization, we employ the pretrained tokenizer. Following the previous work~\cite{yao2026unifieddrivingtokensrepresentation} whose tokenizer is aligned with~\cite{simeoni2025dinov3}, the quantizer contains a codebook of size $K_V$ . Each input image is divided into non-overlapping $H_V \times W_V$ patches and encoded into a sequence of discrete visual tokens. This process is applied independently to each of the $H$ input frames, producing the visual token sequence $\mathbf{V}_{t+1:t+H}$.

For action tokenization, we discretize the continuous future trajectory using the acceleration-based quantization described above. Specifically, the smoothed trajectory is converted into ego-centric longitudinal and lateral accelerations, which are then quantized into a grid-structured acceleration vocabulary. This yields a discrete action token sequence $\mathbf{A}_{t+1:t+H}$, where each token corresponds to a 2D acceleration prototype.

To incorporate action tokens into downstream discrete diffusion modeling, we associate each action token with a learnable embedding through an action embedding table. During diffusion training, corrupted or partially masked action tokens are embedded and fed into the Transformer together with visual tokens, while the model is trained to predict action tokens under the discrete diffusion objective. Through joint optimization, the action embeddings learn compact representations of discrete motion prototypes and are implicitly aligned with visual token representations in the shared hidden space. As a result, visual and action tokens can be processed jointly by a unified Transformer, enabling visually conditioned future action generation through discrete diffusion. Detailed token design are referred in Appendix~\ref{token_design}.

\paragraph{World modeling}
The world modeling task learns action-conditioned future visual prediction. Given the scene context $\mathbf{C}_t$ and a future action sequence $\mathbf{A}_{t+1:t+H}$, the model predicts the future visual token sequence $\mathbf{V}_{t+1:t+H}$. This can be written as
\begin{equation}
p_\theta(\mathbf{V}_{t+1:t+H} \mid \mathbf{C}_t, \mathbf{A}_{t+1:t+H}).
\end{equation}
During training, future action tokens are provided as conditioning inputs through teacher forcing, and the supervision is applied to future visual tokens. This task trains the model to capture how different action sequences induce different future world evolutions.

\paragraph{Policy modeling}
The policy modeling task learns hierarchical decision-conditioned action generation. 
Following the notation in Sec.~\ref{sec:unified_world_policy}, the model first predicts a high-level decision skeleton $\mathbf{D}_t$ from the scene context $\mathbf{C}_t$, and then predicts the future action sequence $\mathbf{A}_{t+1:t+H}$ conditioned on both the context and the decision tokens:
\begin{equation}
p_{\psi,\theta}(\mathbf{A}_{t+1:t+H}, \mathbf{D}_t \mid \mathbf{C}_t)
=
p_\psi(\mathbf{D}_t \mid \mathbf{C}_t)
p_\theta(\mathbf{A}_{t+1:t+H} \mid \mathbf{C}_t, \mathbf{D}_t).
\end{equation}

\paragraph{Theoretical analysis} This hierarchical decomposition separates policy learning into two levels: the high-level decision task captures multi-modal driving choices, while the low-level action prediction task generates smooth and temporally consistent trajectories under the selected decision condition. 
Here, $\mathbf{D}_t$ is treated as a sparse latent skeleton that encodes low-frequency planning structure, such as coarse maneuver intent, reference motion trend, or other decision-level constraints.

The motivation for introducing $\mathbf{D}_t$ is that future action tokens are not conditionally independent when only the scene context is given. 
Let $U\subseteq\{1,\ldots,H\}$ denote a subset of future steps, and let $\mathbf{A}_U=\{\mathbf{A}_{t+h}:h\in U\}$ denote the corresponding action-token group. 
We use $\mathrm{TC}(\cdot\mid\cdot)$ to denote conditional total correlation, which measures the residual statistical dependence among a group of tokens under a given condition. 
As derived in Appendix~\ref{app:analytical_latent_decomp}, conditioning on an upstream decision skeleton changes the residual dependence according to
\begin{equation}
\mathbb{E}_{\mathbf{D}_t}
\mathrm{TC}(\mathbf{A}_U\mid \mathbf{C}_t,\mathbf{D}_t)
=
\mathrm{TC}(\mathbf{A}_U\mid \mathbf{C}_t)
-
R_{\mathbf{D}}(U\mid \mathbf{C}_t),
\end{equation}
where $R_{\mathbf{D}}(U\mid \mathbf{C}_t)$ is the redundancy gain brought by the decision skeleton. 
This identity shows that decision conditioning reduces residual action-token dependence only when $R_{\mathbf{D}}(U\mid \mathbf{C}_t)>0$.

Appendix~\ref{app:analytical_positive_gain} further gives a sufficient condition for positive redundancy gain. 
Under the residual mixing assumption, the skeleton-conditioned dependence between future action tokens decays with their temporal distance:
\begin{equation}
I(\mathbf{A}_{t+i};\mathbf{A}_{t+j}\mid \mathbf{C}_t,\mathbf{D}_t)
\le
\beta
\exp\left(-\frac{d(i,j)}{\ell_D}\right),
\end{equation}
where $\beta$ measures the remaining local coupling strength and $\ell_D$ denotes the residual correlation length after conditioning on the decision skeleton. 
This assumption means that once the low-frequency driving intent is explained by $\mathbf{D}_t$, the remaining fine action tokens mainly encode local residual corrections. 
If the original action-token group has a dependence lower bound $\mathrm{TC}(\mathbf{A}_U\mid \mathbf{C}_t)\ge\kappa(U)$, then
\begin{equation}
\kappa(U)>
\mathbb{E}_{\mathbf{D}_t}
\left[
\sum_{\{i,j\}\subset U}
\beta
\exp\left(-\frac{d(i,j)}{\ell_D}\right)
\right]
\Longrightarrow
R_{\mathbf{D}}(U\mid \mathbf{C}_t)>0.
\end{equation}
Therefore, a valid decision skeleton reduces residual total correlation when it explains stronger group-level low-frequency dependence than the remaining skeleton-conditioned local dependence.

The same analysis also yields a schedule-level KL upper bound when decision prediction error and token-level model error are considered. 
Let $\pi$ denote the token-editing schedule, $A_r$ the active edit set at round $r$, and $S_r$ the editing state before round $r$. 
Appendix~\ref{app:analytical_kl_upper_bound} shows that
\begin{equation}
D_{\mathrm{KL}}
\left(
q(\mathbf{A}\mid \mathbf{C}_t)
\Vert
p_{\psi,\theta,\pi}(\mathbf{A}\mid \mathbf{C}_t)
\right)
\le
\delta_D
+
\delta_{\mathrm{init}}
+
\mathcal{B}_{\mathrm{model}}(\pi)
+
\mathcal{U}_{\mathrm{dep}}(\pi),
\end{equation}
where $\delta_D$ is the decision prediction error, $\delta_{\mathrm{init}}$ measures the mismatch of the initial editing proposal, $\mathcal{B}_{\mathrm{model}}(\pi)$ accumulates token-level model errors over edit rounds, and $\mathcal{U}_{\mathrm{dep}}(\pi)$ bounds the residual dependence within each active edit set. 
This bound suggests that hierarchical decision modeling is beneficial when the decision skeleton is predictable and reduces residual action-token dependence enough to offset its own prediction cost.

\paragraph{World-policy modeling}
The world-policy modeling task jointly learns action prediction and action-conditioned world prediction without using explicit decision tokens. Given the scene context $\mathbf{C}_t$, future action tokens and future visual tokens are arranged as an interleaved sequence along the prediction horizon. At each future step, action prediction is conditioned on the available previous action and visual tokens, while visual prediction is conditioned on the available previous action and visual tokens as well as the current action token. This matches the task-specific attention mask, where each prediction block can attend to its permitted historical action-vision context but cannot access its corresponding clean target tokens.

Let $\mathbf{Y}_{t+1:t+H}$ denote the interleaved future token sequence composed of action and visual tokens:
\begin{equation}
\mathbf{Y}_{t+1:t+H}
=
[\mathbf{A}_{t+1}, \mathbf{V}_{t+1}, \ldots, \mathbf{A}_{t+H}, \mathbf{V}_{t+H}].
\end{equation}
Then the world-policy objective can be described as prefix-conditioned token prediction over this interleaved sequence:
\begin{equation}
p_\theta(\mathbf{Y}_{t+1:t+H}\mid \mathbf{C}_t)
=
\prod_{h=1}^{H}
p_\theta(\mathbf{A}_{t+h}\mid \mathbf{C}_t,\mathbf{Y}_{t+1:t+h-1})
p_\theta(\mathbf{V}_{t+h}\mid \mathbf{C}_t,\mathbf{Y}_{t+1:t+h-1},\mathbf{A}_{t+h}).
\end{equation}
During training, the corresponding attention mask implements this dependency pattern under the token-editing formulation: noisy action and visual tokens are edited using the permitted historical action-vision context, while clean target tokens are kept isolated from their noisy counterparts.

Meanwhile, the world-policy model always applies token-editing supervision to future visual tokens, encouraging the model to recover future world states from corrupted visual tokens under action-conditioned context. In addition, the action stream can be trained with the same token-editing strategy, where corrupted future action tokens are edited toward the ground-truth action sequence. This joint supervision enables the model to learn not only plausible policy generation from the current context, but also how the generated actions shape subsequent visual evolution. As a result, world-policy modeling provides a unified objective for action generation, world prediction, and action-conditioned counterfactual reasoning.

\begin{figure*}[t]
    \centering
    \includegraphics[width=1.0\linewidth]{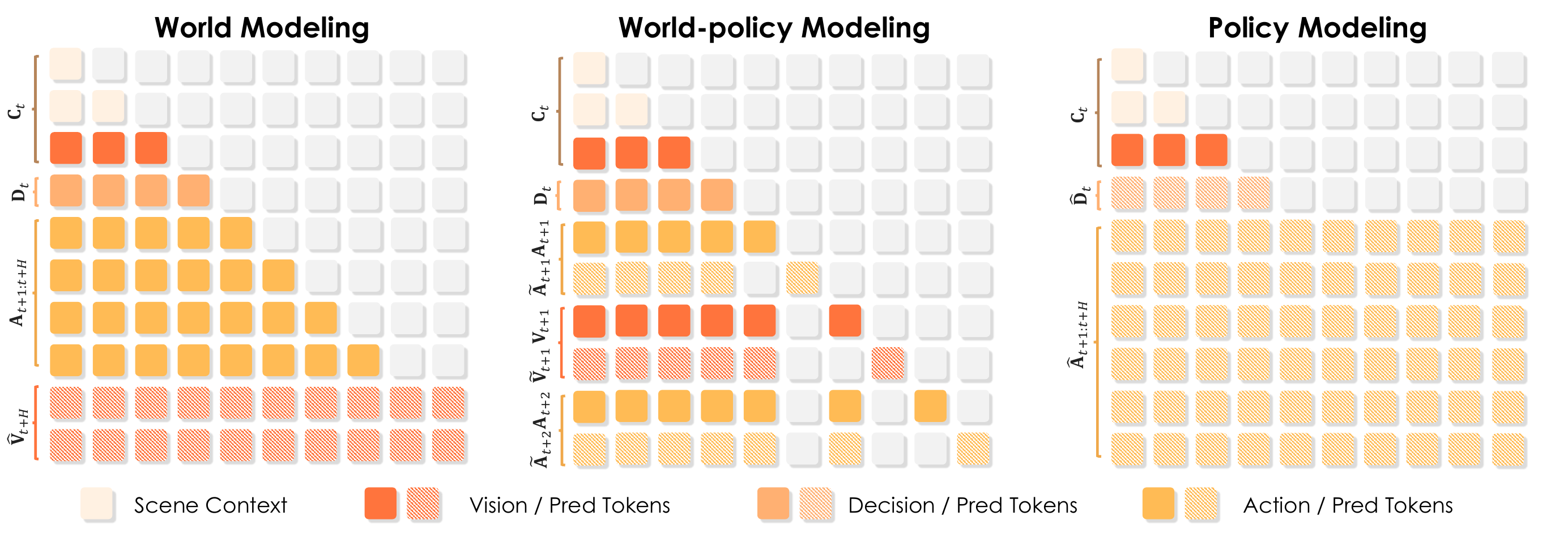}
    \caption{\textbf{Attention masking strategies for \XFM{} pretraining.}
     \XFM{} features three different types of attention masking that controls respective task families during unified pretraining. 
    }
    \label{fig3}
\end{figure*}

\paragraph{Task-specific attention masks}
We use task-specific attention masks to match the information flow required by each training objective. The general principle is that historical context tokens are visible as teacher-forced conditioning information, while current or future corrupted tokens are predicted through token editing without leaking their corresponding clean targets.

For world modeling, future action tokens are treated as clean conditioning inputs rather than editing targets. Therefore, the action stream does not use dual clean-noisy filling in this task. The future visual stream is constructed with the token-editing format, and the model predicts corrupted future visual tokens conditioned on historical observations, ego-state and navigation tokens, and the provided future action sequence.

For policy modeling, the context tokens follow a causal attention structure, so the model can condition on historical observations and available state information without accessing future context. The future action tokens form the editing target block and are allowed to attend bidirectionally within the action block. This bidirectional action attention allows the model to refine a complete future action sequence jointly, while the causal context mask prevents future information leakage from the observation side.

For world-policy modeling, both future visual tokens and future action tokens can be trained with token-editing supervision. We adopt a dual-path filling order for each editable stream, where clean tokens and noisy tokens are placed as separate blocks. The attention mask enforces bidirectional isolation between the clean target block and the corresponding noisy prediction block: noisy tokens can use the permitted context and task-specific visible conditions, but cannot directly attend to their clean targets. During training, teacher forcing provides historical ground-truth information as context, while the model learns to edit noisy visual and action tokens into their clean targets.

\paragraph{Training objectives}
The unified pretraining objective combines token-level classification losses with continuous motion reconstruction losses. For visual token editing, the model is supervised with a cross-entropy loss over the discrete visual vocabulary. Unlike objectives that only supervise corrupted positions, we apply the token classification loss to all editable visual token positions, including both clean and corrupted tokens. For corrupted positions, the loss trains the model to recover the original clean targets from noisy inputs. For clean positions, the same loss encourages an identity mapping, requiring the model to preserve tokens that are already close to the ground truth rather than unnecessarily editing them. This all-position supervision provides an implicit stopping signal for token editing: the model learns not only how to correct noisy tokens, but also when no edit is needed. When latent visual embeddings are available, we further apply a latent reconstruction loss to preserve fine-grained visual semantics beyond discrete token indices.

For action token editing, the model is supervised with a cross-entropy loss over the discrete action vocabulary. In addition to token classification, we impose motion-level supervision to ensure that the discrete action distribution remains physically consistent with continuous future motion. First, we supervise the decoded acceleration associated with the predicted action distribution, encouraging the discrete action tokens to preserve accurate low-level motion semantics. Second, we convert the predicted action distribution into continuous accelerations and integrate them twice over time to reconstruct the future ego trajectory, on which a trajectory-level regression loss is applied. In addition, we introduce an auxiliary factorized position classification loss, where the ground-truth future $x$ and $y$ positions are discretized into separate position vocabularies and supervised independently. 
The trajectory obtained by integrating decoded accelerations is used only for the trajectory-level regression loss. The detailed vocabulary configuration and loss formulation are provided in Appendix~\ref{implementation_details}.

A naive way to obtain continuous accelerations for motion reconstruction is to multiply the predicted action probabilities with the acceleration vocabulary and use the resulting full-distribution expectation. However, action prediction is often inherently multi-modal: different acceleration modes may correspond to different plausible driving decisions. Directly taking the expectation over the full predicted distribution can average incompatible modes and produce physically implausible intermediate accelerations. We refer to this issue as decoding-induced mode averaging.

To mitigate this issue, we use a mode-aware decoding strategy only for constructing the continuous acceleration, trajectory reconstruction, and auxiliary position supervision losses. We fit the predicted categorical distribution over the acceleration vocabulary with a multi-modal Gaussian mixture. Specifically, we consider GMMs with one, two, and three components and select the one with the lowest fitting error. We then apply top-$p$ sampling to select one Gaussian mode, re-normalize the action probabilities within the selected mode, and compute the expected acceleration from the re-normalized distribution and the acceleration vocabulary. The resulting mode-aware acceleration is integrated over time for trajectory reconstruction.

This mode-aware decoding is independent of the soft-label interpolation used for action tokenization. The latter removes deterministic hard-assignment quantization error in the construction of the action target, whereas the former is introduced to prevent continuous reconstruction losses from averaging mutually incompatible action modes. The mode-selection step introduces an additional approximation, whose role and error decomposition are discussed in Appendix~\ref{token_design}.

The final objective is a weighted sum of task-dependent loss terms:
\begin{equation}
\mathcal{L}
=
\lambda_v \mathcal{L}_{v}^{\mathrm{cls}}
+
\lambda_a \mathcal{L}_{a}^{\mathrm{cls}}
+
\lambda_{\mathrm{acc}} \mathcal{L}_{\mathrm{acc}}
+
\lambda_{\mathrm{traj}} \mathcal{L}_{\mathrm{traj}}
+
\lambda_{\mathrm{pos}} \mathcal{L}_{\mathrm{pos}}^{\mathrm{cls}}
+
\lambda_s \mathcal{L}_{s}^{\mathrm{cls}}
+
\lambda_\mathrm{dec} \mathcal{L}_\mathrm{dec}.
\end{equation}
Here, $\mathcal{L}_{v}^{\mathrm{cls}}$ denotes the cross-entropy loss over the visual token vocabulary, and $\mathcal{L}_{a}^{\mathrm{cls}}$ denotes the cross-entropy loss over the action token vocabulary. $\mathcal{L}_{\mathrm{acc}}$ is the acceleration-level regression loss, while $\mathcal{L}_{\mathrm{traj}}$ is the trajectory-level regression loss obtained after integrating predicted accelerations into future ego trajectories. $\mathcal{L}_{\mathrm{pos}}^{\mathrm{cls}}$ denotes the auxiliary factorized position classification loss over the integrated future positions, with separate classification heads for longitudinal and lateral coordinates. $\mathcal{L}_{s}^{\mathrm{cls}}$ denotes the classification loss for auxiliary special tokens, such as task or control tokens used to organize different training sequences. $\mathcal{L}_{\mathrm{dec}}$ terms for decision classification loss. The coefficients $\lambda_v$, $\lambda_a$, $\lambda_{\mathrm{acc}}$, $\lambda_{\mathrm{traj}}$, $\lambda_{\mathrm{pos}}$, $\lambda_{\mathrm{dec}}$, and $\lambda_s$ balance the relative contribution of each loss term. Different training tasks activate different subsets of these losses according to their prediction targets.

\paragraph{Training schedule}
\label{training-schedule}
We adopt a multi-stage training schedule to progressively align visual world modeling and action generation. In the first stage, we perform visual pretraining with both world-policy modeling and world modeling tasks. Only the vision prediction loss is applied in this stage, while future action tokens are provided through teacher forcing as conditioning inputs. This stage trains discrete visual representations that are predictive of future scene evolution and aligned with action-conditioned dynamics.

In the second stage, we jointly train visual and action prediction. In addition to world modeling, the world-policy modeling task also activates the action prediction loss, so the model learns to recover both future visual tokens and future action tokens under the shared token-editing framework. This stage strengthens the coupling between policy generation and action-conditioned world evolution.

In the third stage, we perform action finetuning with a LoRA adapter. This stage focuses on the discrete diffusion policy model, where the model is finetuned specifically for future action generation and refinement while preserving the visual-world representations learned in the earlier stages.

\subsection{Post Training}
To further capture rare or safety-critical behaviors by limited coverage in the dataset, we apply a post-training fine-tuning stage that leverages model-based trajectory sampling and reinforcement learning to refine the policy on challenging scenarios while preserving previously learned behavior. 

\paragraph{Policy sampling} For each driving scene context $\mathbf{C}_t$, \XFM{} generates a group of candidate trajectories $\{\tau_1, \dots, \tau_G\}$ via token-edit planner for efficient exploration. Each trajectory is iteratively edited through $r$ rounds by: $\tau_i\sim p_{\theta}(\mathbf{\hat{A}}^{(r)}_{t+1:t+H} \mid \mathbf{\tilde{A}}^{(r-1)}_{t+1:t+H},\mathbf{\hat{D}}_t,\mathbf{C}_t)p_\psi(\mathbf{\hat{D}}_t\mid\mathbf{C}_t)$. This produces a diverse set of high-quality rollouts evaluated using online reward function $R(\tau_i)$, i.e., (E)/PDMS metrics. The hierarchical modeling for decision of $\mathbf{D}_t$ further embraces two type of sampling strategies. 1) Group sampling under the most probable decision $\arg\max p_\psi(\mathbf{\hat{D}}_t\mid\mathbf{C}_t)$. 2) A parallel line of sampling strategy that \XFM{} adopts directly leverage the full distribution of decisions $\mathbf{\hat{D}}_t$, and conduct group sampling specific for each decision token. This further offers decision-level post training update for $\log p_\psi(\mathbf{\hat{D}}_t\mid\mathbf{C}_t)$.

\paragraph{Training objectives} Following the Grouped Relative Policy Optimization (GRPO) paradigm, we compute per-token log-probabilities under the current policy $\pi^{A}_{\theta,i}\sim \frac{1}{H}\sum^H_{h=1}\log_{\mathbf{A}\sim\tau_i} p_\theta(\mathbf{\hat{A}}_{t+h} \mid \mathbf{\hat{A}}_{<t+h},\mathbf{\hat{D}}^i_t,\mathbf{C}_t)$ using a one-step reconstruction estimator following~\cite{zhao2026d1}. Decision distribution are directly gathered as $\pi^D_{\theta,i}\sim\log p_\psi(\mathbf{\hat{D}}^i_t\mid\mathbf{C}_t)$. The advantage is computed by $A_i=R(\tau_i)-\sum^G_{i=1}R(\tau_i)$, the overall objective becomes:
\begin{equation}
\mathbf{E}_{\tau\sim\pi_{\theta,\psi}} \sum_{k\in(A,D)}\Bigg[ 
\frac{1}{G} \sum_{i=1}^G  
\min \Big( \rho^k_i A_i, \text{clip}(\rho^k_i, 1-\epsilon, 1+\epsilon) A_i \Big)- \text{KL}(\pi^k_{\theta,\psi}||\pi^k_{\text{ref}})
\Bigg],
\end{equation}
where $\rho^k_i=\pi_{\theta,\psi,i}^k/\pi_{\text{ref},i}^k$ terms for the importance sampling ratio. 

\section{Evaluation}
\subsection{Setup}
\begin{table}[t]
\centering
\resizebox{\linewidth}{!}{
\begin{tabular}{l |c  c c c c c c c c| c c}
\toprule
\rowcolor[HTML]{FFE0CC}
\textbf{Method} & 
\textbf{NC$\uparrow$} & \textbf{DAC$\uparrow$} & \textbf{DDC$\uparrow$} &
\textbf{TLC$\uparrow$} & \textbf{EP$\uparrow$} & \textbf{TTC$\uparrow$} &
\textbf{LK$\uparrow$} & \textbf{HC$\uparrow$} & \textbf{EC$\uparrow$} &
\textbf{EPDMS$^*\uparrow$} & \textbf{EPDMS$\uparrow$} \\
\midrule
Transfuser~\cite{chitta2022transfuser}  & 96.9 & 89.9 & 97.8 & 99.7 & 87.1 & 95.4 & 92.7 & \textbf{98.3} & \underline{87.2} & 76.7 & - \\
ReCogDrive~\cite{li2025recogdrive} & 98.3 &95.2 &\underline{99.5} &\underline{99.8} &87.1 &97.5 &96.6 &98.3 &86.5&83.6 &-\\
WAM-Flow~\cite{xu2025wam} &98.5 &94.5 &99.5 &\underline{99.8} &86.9& 96.8 &97.4& 97.6& 73.9& 84.7 & -\\
Epona~\cite{zhang2025epona} & 97.1 &95.7 &99.3 &99.7 &88.6 &96.3 &97.0 &98.0 &67.8 &- & 85.1\\
DiffusionDriveV2~\cite{zou2025diffusiondrivev2}  & 97.7 & 96.6 & 99.2 & \underline{99.8} & 88.9 & 97.2 & 96.0 & 97.8 & \textbf{91.0} & 85.5 & 87.5 \\
Hydra-MDP++~\cite{li2024hydra}  & \underline{98.4} & 98.0 & 99.4 & \underline{99.8} & 87.5 & 97.7 & 95.3 & \textbf{98.3} & 77.4 & 85.1 & - \\
DriveSuprim~\cite{yao2026drivesuprim}& 97.8 & 97.9 & 99.5 & \textbf{99.9} & \underline{90.6} & 97.1 & 96.6 & \textbf{98.3} & 77.9 & 86.0 & - \\
DriveVLA-W0~\cite{li2025drivevla} & \textbf{98.5} &\textbf{99.1} &98.0 &99.7 &86.4 &98.1 &93.2 &97.9 &58.9 &86.1 &-\\
DreamerAD~\cite{yang2026dreamerad} & 98.0 & 97.2 & 99.5 & \underline{99.8} & 87.8 & 97.4 & 97.5 & 98.3 & 72.4 & - &87.7 \\
SparseDriveV2~\cite{sun2026sparsedrivev2} & 98.1 & {98.1} & \underline{99.6} & \underline{99.8} & \textbf{91.1} & 97.3 & \underline{96.9} & \underline{98.2} & 78.4 & \underline{86.7} & \underline{90.1} \\
\midrule
\rowcolor[HTML]{FFE0CC}
\textbf{Discrete-WAM}  & \textbf{98.5} & \underline{98.2} & \textbf{99.7} & \underline{99.8} & {90.5} & \underline{97.9} & \textbf{97.2} & \textbf{98.3} & 78.1 & \textbf{87.0} & \textbf{90.4} \\
\bottomrule
\end{tabular}
}
\vspace{1ex}
\caption{\textbf{Comparison with state-of-the-art methods on the NAVSIM-v2 benchmark.}
We report no collision (NC), drivable area compliance (DAC), driving direction compliance (DDC), traffic light compliance (TLC), ego progress (EP), time-to-collision (TTC), lane keeping (LK), human comfort (HC), ego comfort (EC), EPDMS$^*$ (before benchmark bug fix), and EPDMS. The best results are highlighted in \textbf{bold}, and the second-best result is underlined.}
\label{tab:navsim_extended}
\end{table}

\paragraph{Dataset and benchmarks} We manifest the E2E generation and planning capabilities of \XFM{} on NAVSIM-v1 and v2 benchmark~\cite{dauner2024navsim,cao2025pseudo}, which offers large-scale driving scenarios for end-to-end driving. Following the standard protocol, we evaluate \XFM{} on the navtest split, containing 12k driving scenes sampled at 2 Hz. NAVSIM evaluates planning quality of 4s horizon trajectories with PDMS and the extended EPDMS, where safety- and rule-critical metrics are incorporated as multiplicative constraints, while progress and comfort-related terms are combined through weighted aggregation. The reported metrics include no at-fault collision (NC), drivable area compliance (DAC), driving direction compliance (DDC), traffic light compliance (TLC), ego progress (EP), time-to-collision (TTC), lane keeping (LK), history comfort (HC), and extended comfort (EC). Detailed metric formulations and baselines are referred in the Appendix~\ref{benchmark_detail}.

\paragraph{Implementation details} \XFM{} follows a multi-stage training schedule as previously detailed in Sec.~\ref{training-schedule}. The unified world-policy pretraining with task families is conducted on the full nuPlan training set~\cite{caesar2021nuplan} for 200k steps with a learning rate of $1\times10^{-4}$. After pretraining, \XFM{} receives supervised finetuning on navtrain dataset for 10 epochs. Finally, \XFM{} applies RL post-training for another 2 epochs to further improve planning behavior. Both the SFT and RL post-training stages are performed with LoRA finetuning at a learning rate of $1\times10^{-5}$. All stages are trained on 32 NVIDIA H20 GPUs using AdamW optimizer with a cosine schedule. Model details are referred in the Appendix~\ref{benchmark_detail}.

\subsection{Quantitative Results}

\paragraph{Planning results} On NAVSIM v2, \XFM{} delivers a strong planning result of 90.4 EPDMS, outperforming a series of method jointly built with world modeling, post-training, or generative planers with discretizations. Specifically, \XFM{} improves EPDMS of +2.7 over WAM-Flow~\cite{xu2025wam}, while both improving safety and comfort. Compared with world-model-based methods~\cite{yang2026dreamerad,zhang2025epona,li2025drivevla}, \XFM{} offers a 6.2\% relative improvement over~\cite{zhang2025epona}, and  3.1\% of~\cite{yang2026dreamerad}. Compared with reinforced cognitive planner, \XFM{} performs a 4.1\% relative gain. Consistent performance gains are also observed on the NAVSIM-v1 benchmark, where Discrete-WAM achieves competitive results with a +2.1 PDMS improvement over WAM-Flow~\cite{xu2025wam} and a +7.0 PDMS improvement over world-model-based planners~\cite{zhang2025epona}. These results indicate that both unified world-policy pretraining and the token-editing formulation contribute to the planning performance gains. The former learns aligned vision-action-future representations that support compositional generalization, while the latter provides predictive scenario latents that enhance causal awareness and enable more reliable planning refinement.

\begin{table}[t]
\centering
\setlength{\tabcolsep}{5.8mm}{
\begin{tabular}{l| c c c c c | c}
\toprule
\rowcolor[HTML]{FFE0CC}
\textbf{Method} &
\textbf{NC$\uparrow$} & \textbf{DAC$\uparrow$} & \textbf{TTC$\uparrow$} &
\textbf{Comf.$\uparrow$} & \textbf{EP$\uparrow$} & \textbf{PDMS$\uparrow$} \\
\midrule
VADv2~\cite{chen2024vadv2}              & 97.2 & 89.1 & 91.6 & \textbf{100} & 76.0 & 80.9 \\
UniAD~\cite{hu2023planning}           & 97.8 & 91.9 & 92.9 & \textbf{100} & 78.8 & 83.4 \\
Transfuser~\cite{chitta2022transfuser}  & 97.7 & 92.8 & 92.8 & \textbf{100} & 79.2 & 84.0 \\
PARA-Drive~\cite{weng2024drive}   & 97.9 & 92.4 & 93.0 & 99.8 & 79.3 & 84.0 \\
GoalFlow~\cite{xing2025goalflow}      & 98.3 & 93.8 & 94.3 & \textbf{100} & 79.8 & 85.7 \\
Epona~\cite{zhang2025epona} &97.9 &95.1 &93.8 &\underline{99.9} &80.4 &86.2\\
Hydra-MDP++~\cite{li2024hydra}    & 97.6 & 96.0 & 93.1 & \textbf{100} & 80.4 & 86.6 \\
DiffusionDrive~\cite{liao2025diffusiondrive}  & 98.2 & 96.2 & 94.7 & \textbf{100} & 82.2 & 88.1 \\
WoTE~\cite{li2025end}           & 98.5 & 96.8 & 94.9 & \underline{99.9} & 81.9 & 88.3 \\
DriveSuprim~\cite{yao2026drivesuprim}& 97.8 & 97.3 & 93.6 & \textbf{100} & 86.7 & 89.9 \\
DriveVLA-W0~\cite{li2025drivevla} &98.7 &\textbf{99.1} &\underline{95.3} &99.3 &83.3 &90.2 \\ 
WAM-Flow~\cite{xu2025wam}  &\textbf{99.2} &98.3 &\textbf{97.0} &99.7 &82.3 &90.3 \\
ReCogDrive~\cite{li2025recogdrive} &97.9 &97.3 &94.9 &\textbf{100} &87.3& 90.8\\
ReflectDrive-2~\cite{wang2026reflectdrive} & 97.3 &98.1 &92.5 &\textbf{100}& 89.4 &91.0\\
DiffusionDriveV2~\cite{zou2025diffusiondrivev2} & 98.3 & 97.9 & 94.8 & \underline{99.9} & 87.5 & 91.2 \\
iPad~\cite{guo2025ipad}              & {98.6} & 98.3 & 94.9 & \textbf{100} & 88.0 & 91.7 \\
SparseDrive-V2~\cite{sun2026sparsedrivev2} & 98.5 & \underline{98.4} & {95.0} & \underline{99.9} & \underline{88.6} & \underline{92.0} \\
\midrule
\rowcolor[HTML]{FFE0CC}
\textbf{Discrete-WAM}         & \underline{98.8} & \underline{98.4} & \underline{95.3} & \textbf{100} & \textbf{88.7} & \textbf{92.2} \\
\bottomrule
\end{tabular}
}
\vspace{1ex}
\caption{\textbf{Comparison with state-of-the-art planning methods on NAVSIM-v1.}
All methods are evaluated using the official NAVSIM-v1 metrics: no collision (NC), drivable area compliance (DAC), time-to-collision (TTC), comfort (Comf.), ego progress (EP), and the final planning driving metric score (PDMS). The best result in each column is shown in bold, and the second-best result is underlined.}
\vspace{-5pt}
\label{tab:navsim_comparison}
\end{table}
\begin{table}[!h]
\centering
\setlength{\tabcolsep}{4pt}
\renewcommand{\arraystretch}{1.1}
\resizebox{\linewidth}{!}{%
\begin{tabular}{l|cccccc|c}
\toprule
\cellcolor[HTML]{FFE0CC}\textbf{Metric} 
& \textbf{DriveDreamer~\cite{wang2024drivedreamer}} 
& \textbf{WoVoGen~\cite{lu2024wovogen}} 
& \textbf{Drive-WM~\cite{wang2024driving}} 
& \textbf{GenAD (OpenDV)~\cite{yang2024genad}} 
& \textbf{Vista~\cite{gao2024vista}} 
& \textbf{DrivingWorld~\cite{hu2024drivingworld}} 
& \cellcolor[HTML]{FFE0CC}\textbf{\XFM{}} \\
\midrule
\cellcolor[HTML]{FFE0CC}FID $\downarrow$ 
 & 52.6 & 27.6 & 15.8 & 15.4 & 6.9 & 7.4 & \cellcolor[HTML]{FFE0CC}\textbf{6.6}  \\

\cellcolor[HTML]{FFE0CC}FVD $\downarrow$ 
 & 452.0 & 417.7 & 122.7 & 184.0 & 89.4 & 90.9 & \cellcolor[HTML]{FFE0CC}\textbf{80.0} \\

\cellcolor[HTML]{FFE0CC}Max Duration / Frames$^{*}$ 
& 4s / 48 & 2.5s / 5 & 8s / 16 & 4s / 8 & 15s / 150 & 40s / 400 & \cellcolor[HTML]{FFE0CC}4s / 8 \\
\bottomrule
\end{tabular}
}
\vspace{1ex}
\caption{\textbf{Comparison of generative driving world models.}
We report FID, FVD, and maximum generation duration/frames. The best results are highlighted in \textbf{bold}.}
\label{tab:world_model_comparison}
\end{table}

\paragraph{World generation results} Tab.~\ref{tab:world_model_comparison} quantitatively compares \XFM{} with prior generative driving world models in terms of image and video generation quality. While \XFM{} is primarily designed for short-horizon unified generation to facilitate downstream planning, it achieves the best overall visual fidelity, obtaining an FID of 6.6 and an FVD of 80.0, outperforming existing approaches including Vista~\cite{gao2024vista} and DrivingWorld~\cite{hu2024drivingworld}. Notably, while some prior methods support substantially longer rollouts, their generation quality degrades as the horizon increases. These results demonstrate that the proposed discrete world-action modeling framework can capture future driving dynamics and scene evolution more effectively while maintaining lower generation cost.


\subsection{Analysis}
\label{sec:analysis}
\paragraph{Effect of unified pretraining}
We compare three policy training strategies to isolate the effect of unified pretraining. ``From scratch'' trains the policy model directly on the downstream planning data without any unified pretraining. ``FT'' initializes the model from unified pretraining and then updates all trainable parameters during supervised finetuning. ``LoRA-SFT'' first performs vision-oriented world-policy pretraining, where future actions are provided as teacher-forced conditions and only vision prediction losses are applied, and then finetunes the policy with LoRA adapters. We study the benefit of unified pretraining with different 
\begin{table}[!htbp]
\centering
\resizebox{\linewidth}{!}{
\begin{tabular}{l |c  c c c c c c c c| c c}
\toprule
\rowcolor[HTML]{FFE0CC}
\textbf{Ablations} & 
\textbf{NC$\uparrow$} & \textbf{DAC$\uparrow$} & \textbf{DDC$\uparrow$} &
\textbf{TLC$\uparrow$} & \textbf{EP$\uparrow$} & \textbf{TTC$\uparrow$} &
\textbf{LK$\uparrow$} & \textbf{HC$\uparrow$} & \textbf{EC$\uparrow$} &
\textbf{EPDMS$^*\uparrow$} & \textbf{EPDMS$\uparrow$} \\
\midrule
From scratch  
& 98.5 
& \textbf{98.1} 
& \textbf{99.6} 
& 99.7 
& 90.2 
& \textbf{97.8} 
& \textbf{97.1} 
& 96.8 
& 75.6 
& 86.5 
& 89.8 \\
FT  
& \textbf{98.6} 
& 98.0 
& \textbf{99.6} 
& 99.7 
& \textbf{90.3} 
& \textbf{97.8} 
& \textbf{97.1} 
& 96.4 
& 75.4 
& 86.4 
& 89.7 \\
LoRA-SFT  
& \textbf{98.6} 
& \textbf{98.1} 
& \textbf{99.6} 
& \textbf{99.8} 
& \textbf{90.3} 
& \textbf{97.8} 
& \textbf{97.1} 
& \textbf{97.0} 
& \textbf{76.8} 
& \textbf{86.7} 
& \textbf{90.0} \\
\bottomrule
\end{tabular}
}
\vspace{-5pt}
\caption{\textbf{Effect of policy training strategies on the NAVSIM-v2 benchmark.}
}
\vspace{-5pt}
\label{tab:ablation-pretraining}
\end{table}
\begin{table}[!htbp]
\centering
\resizebox{\linewidth}{!}{
\begin{tabular}{l |c  c c c c c c c c| c c}
\toprule
\rowcolor[HTML]{FFE0CC}
\textbf{Ablations} & 
\textbf{NC$\uparrow$} & \textbf{DAC$\uparrow$} & \textbf{DDC$\uparrow$} &
\textbf{TLC$\uparrow$} & \textbf{EP$\uparrow$} & \textbf{TTC$\uparrow$} &
\textbf{LK$\uparrow$} & \textbf{HC$\uparrow$} & \textbf{EC$\uparrow$} &
\textbf{EPDMS$^*\uparrow$} & \textbf{EPDMS$\uparrow$} \\
\midrule
SFT  
& 98.2 
& \textbf{99.6} 
& 99.5 
& 98.6 
& 90.2 
& 97.6 
& \textbf{97.7} 
& 95.2 
& 73.6 
& 86.5 
& 89.1 \\
SFT-$\mathbf{D}_t$  
& \textbf{98.6} 
& 98.1 
& 99.6 
& \textbf{99.8} 
& 90.3 
& 97.8 
& 97.1 
& 97.0 
& 76.8 
& 86.7 
& 90.0 \\
RL  
& 98.5 
& 98.2 
& \textbf{99.7} 
& \textbf{99.8} 
& \textbf{90.5} 
& \textbf{97.9} 
& 97.2 
& \textbf{98.3} 
& \textbf{78.1} 
& \textbf{87.0} 
& \textbf{90.4} \\
\bottomrule
\end{tabular}
}
\vspace{-5pt}
\caption{\textbf{Effect of post-training on the NAVSIM-v2 benchmark.}}
\vspace{-5pt}
\label{tab:ablation-post-training}
\end{table}
\begin{table}[!htbp]
\centering
\resizebox{\linewidth}{!}{
\begin{tabular}{l |c  c c c c c c c c| c c}
\toprule
\rowcolor[HTML]{FFE0CC}
\textbf{Ablations} & 
\textbf{NC$\uparrow$} & \textbf{DAC$\uparrow$} & \textbf{DDC$\uparrow$} &
\textbf{TLC$\uparrow$} & \textbf{EP$\uparrow$} & \textbf{TTC$\uparrow$} &
\textbf{LK$\uparrow$} & \textbf{HC$\uparrow$} & \textbf{EC$\uparrow$} &
\textbf{EPDMS$^*\uparrow$} & \textbf{EPDMS$\uparrow$} \\
\midrule
Base  & 98.2 & 92.9 & 99.3 & \textbf{99.8} & 86.7 & 97.6 & \textbf{97.7} & 98.3 & 83.3 & 82.9 & 84.7 \\
Base-$\mathbf{D}_t$  & \textbf{{98.8}} & \textbf{{94.2}} & \textbf{{99.5}} & \textbf{{99.8}} & \textbf{{87.3}} & \textbf{{98.2}} & {97.4} & \textbf{98.5} & \textbf{87.7} & \textbf{84.0} & \textbf{87.2} \\
\bottomrule
\end{tabular}
}
\vspace{-5pt}
\caption{\textbf{Effect of decision modeling on the NAVSIM-v2 benchmark.}
}
\vspace{-5pt}
\label{tab:ablation-decision}
\end{table}
adaptation strategies under $\mathbf{D}_t$. As in Tab.~\ref{tab:ablation-pretraining}, training from scratch already gives strong performance, indicating the sufficient fitting capacity for \XFM{}. However, full finetuning does not further improve the result, likely because the orthogonal of planning objective that may overwrite useful pretrained representations. In contrast, LoRA-SFT achieves the best performance, improving EPDMS to 90.0, suggesting that lightweight adaptation better preserves pretrained world-policy knowledge while adapting to planning.

\paragraph{Effect of post training} 
As in Tab.~\ref{tab:ablation-post-training}, compared with based supervised finetuning, leveraging the ground-truth decision condition increases EPDMS from 89.1 to 90.0, confirming that decision-level guidance provides an effective behavioral prior. Nevertheless, this setting still lacks exploration over different decisions and the corresponding trajectory 
refinements. Our RL post-training further improves EPDMS to 90.4 and EPDMS to 87.0, with gains in both comfort and safety. This suggests that the proposed post-training stage jointly benefits high-level decision optimization and low-level token editing, enabling the planner to explore better decision-trajectory combinations rather than only refining trajectories under a fixed decision.

\begin{figure}[!h]
    \centering
    \includegraphics[width=\linewidth]{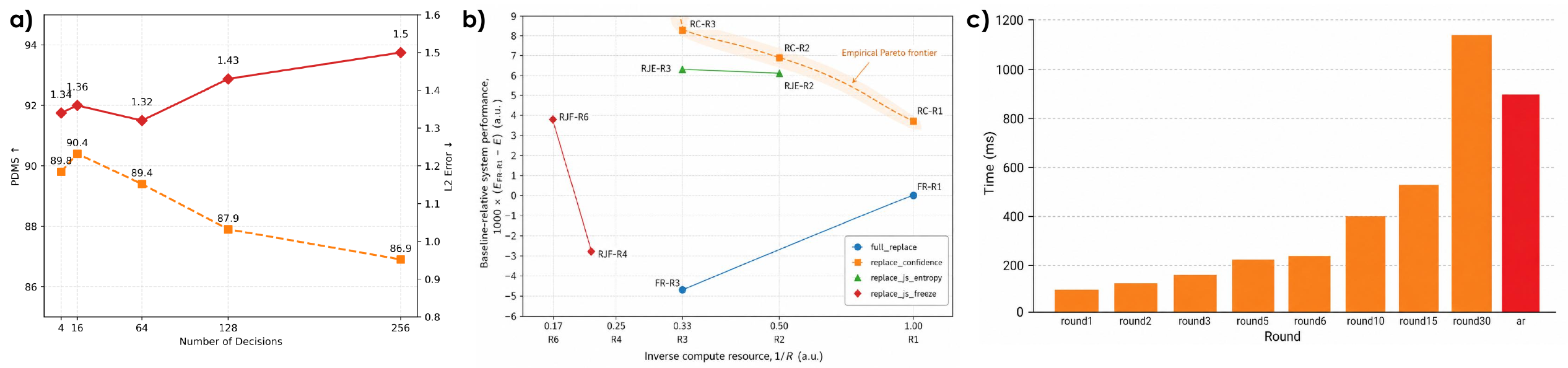}
    \caption{\textbf{Ablation visualization results of \XFM{}.}
    {a) Effect of performance trade-off with decision numbers.} The orange curve denotes the EPDMS with varied decision numbers. The red curve terms for average L2 planning errors under respective number of decisions. {b) Compute performance trade-off of re-edit schedules.} The horizontal axis denotes \(x=1/R\), and the vertical axis denotes \(y=1000(E_{\mathrm{FR},1}-E_{s,R})\).
    {c) Inference latency comparison between discrete diffusion and autoregressive policy decoding.} We compare \texttt{replace\_confidence} discrete diffusion decoding with an autoregressive action decoder.
    }
    \label{fig:viz_pdm_anchor}
\end{figure}

\paragraph{Effect of decision modeling}
Tab.~\ref{tab:ablation-decision} reflects the capabilities enabled by decision learning during pretraining. Introducing $\mathbf{D}_t$ improves EPDMS from 84.7 to 87.2 with consistent gains in sub-metrics. This indicates a strong behavioral prior offered by decision at the pretraining stage that better aligns high-level intent with low-level trajectory generation. The result suggests that the unified world-policy pretraining learns not only scene prediction, but also decision-conditioned planning capability before downstream SFT. Interestingly, performance peaks at $k=16$ and gradually degrades as the anchor set expands. As in Fig.~\ref{fig:viz_pdm_anchor}-a, we attribute this to a trade-off between diversity and optimization difficulty: larger top-$d$ values improve decision coverage but introduce more low-quality anchors, increasing reward variance and weakening the GRPO advantage signal. Consequently, a moderate anchor set provides the most effective policy refinement.

\paragraph{Effect of scheduling strategies}
We evaluate four scheduling strategies for iterative policy-token decoding. 
\texttt{full\_replace} is the baseline scheduler, where all policy tokens are predicted and replaced at every round. 
\texttt{replace\_confidence} accepts only tokens whose prediction confidence exceeds a predefined threshold, while keeping the remaining tokens unchanged for future refinement. 
\texttt{replace\_js\_entropy} further considers cross-round uncertainty and distributional change, updating tokens with high entropy, large Jensen--Shannon divergence, or unstable argmax predictions. 
Finally, \texttt{replace\_js\_freeze} adds a hard-freeze mechanism: tokens that remain stable for consecutive rounds are frozen and no longer updated. 
For each scheduler, we evaluate different scheduling rounds and report the L2 change between the lowest and highest available rounds in Table~\ref{tab:round_l2_delta}.
We report the L2 change between the lowest and highest available rounds for each scheduling strategy, defined as
\(\Delta \mathrm{L2}=\mathrm{L2}_{\mathrm{high\ round}}-\mathrm{L2}_{\mathrm{low\ round}}\).
The results show that increasing the number of rounds does not always improve performance. 
For \texttt{full\_replace}, additional rounds consistently increase L2 error, suggesting that repeatedly overwriting all action tokens can perturb already reasonable predictions. 
This effect is particularly harmful for acceleration-token decoding, where small changes in early acceleration tokens can be amplified through temporal integration into long-horizon position errors. 
In contrast, the selective schedulers improve or preserve L2 performance with more rounds. 
By updating only confident, uncertain, distributionally unstable, or non-frozen tokens, these schedulers use additional computation to refine unresolved parts of the action sequence while retaining stable predictions.



\begin{figure*}[t]
    \centering
    \includegraphics[width=1.0\linewidth]{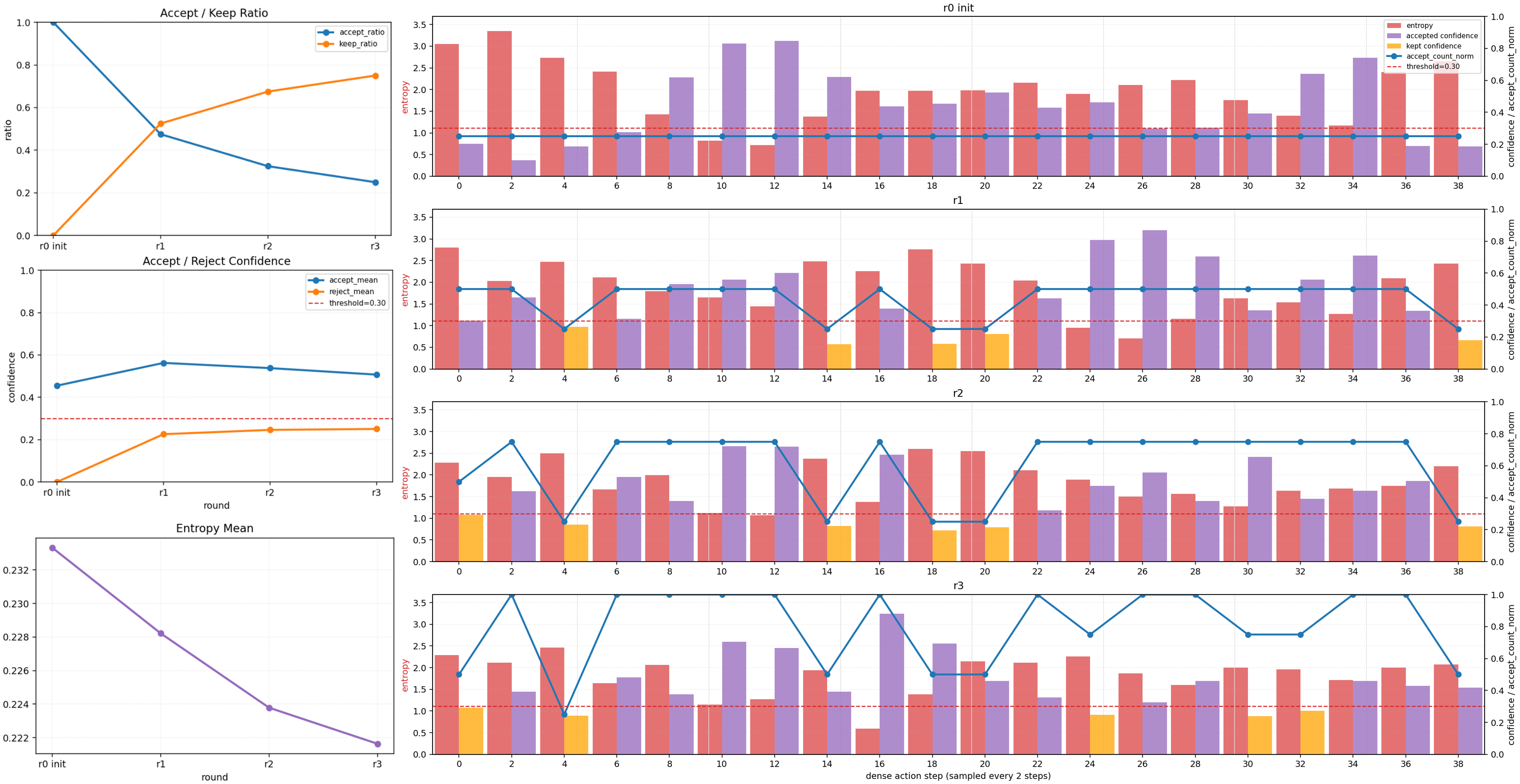}
    \caption{\textbf{Scheduling dynamics of confidence-based token replacement.}
    We visualize \texttt{replace\_confidence} with four editing rounds in a representative left-turn scenario to inspect the detailed multi-round scheduling behavior. 
    The first round is treated as the initialization round, where all tokens are accepted because no previous prediction is available. 
    The left column reports round-level statistics, including the accept/keep ratio, accepted/rejected token confidence, and mean normalized entropy. 
    The right panel shows token-level details across rounds, where each row corresponds to one round and the horizontal axis denotes action-token time steps, with only even steps \(0,2,4,\ldots,38\) shown for readability. 
    Red bars indicate raw entropy, purple/orange bars indicate accepted/rejected tokens, the red dashed line marks the confidence threshold, and the blue curve shows the cumulative accepted-update ratio for each token. 
    The visualization shows that high-confidence tokens are more likely to be accepted and retained, while high-entropy tokens are refined over multiple rounds, leading to progressive entropy reduction.}
    \label{fig:schedule_details}
\end{figure*}

We further analyze the trade-off between scheduling computation and trajectory performance. 
Since we do not strictly measure wall-clock latency or exact FLOPs, we use the scheduling round number \(R\) as a coarse proxy for compute cost and define the inverse-compute coordinate as \(x=1/R\). 
For trajectory performance, we compute the mean L2 error across the four evaluated horizons as \(E_{s,R}=\frac{1}{4}\sum_{t\in\{1,2,3,4\}}\mathrm{L2}_{s,R}@t\), where \(s\) denotes the scheduling strategy and \(R\) denotes the scheduling round. 
For clearer visualization, we use a baseline-relative coordinate with \texttt{full\_replace} at \(R=1\) as the baseline: \(y_{s,R}=1000(E_{\mathrm{FR},1}-E_{s,R})\). 
A positive \(y\) indicates lower mean L2 error than the baseline, while a negative value indicates worse performance. 
The resulting compute-performance coordinates are reported in Table~\ref{tab:compute_performance_coordinates} and visualized in Fig.~\ref{fig:viz_pdm_anchor}-b.

\paragraph{Scheduling dynamics of confidence replacement}
We further analyze the detailed editing dynamics of the \texttt{replace\_confidence} scheduler using a representative left-turn scenario. 
As shown in Fig.~\ref{fig:schedule_details}, the token acceptance ratio gradually decreases as the number of scheduling rounds increases, indicating that more action tokens become stable and no longer require further editing. 
The accepted and rejected tokens also exhibit clearly separated confidence distributions: accepted tokens consistently have higher confidence, while rejected tokens have lower confidence and are preserved for later refinement. 
This shows that confidence provides an effective signal for distinguishing tokens that are ready to be updated from tokens that remain uncertain. 
Meanwhile, the average normalized entropy decreases across scheduling rounds, suggesting that the model prediction becomes progressively sharper during iterative editing.

The token-level visualization further explains this behavior. 
Tokens with high initial confidence tend to have high final retention, which means that once these positions are confidently predicted, they are less likely to be overwritten in later rounds. 
In contrast, tokens with low initial confidence usually have higher entropy, indicating ambiguous action distributions. 
As scheduling proceeds, their confidence gradually increases and entropy decreases, showing a progressive uncertainty reduction process. 
Across different rounds, we also consistently observe an inverse relationship between confidence and entropy: high-confidence tokens usually have low entropy, while high-entropy tokens are more likely to be rejected and refined in subsequent rounds. 
These observations explain why confidence-based selective replacement can improve multi-round policy editing: additional rounds are mainly allocated to uncertain tokens, while stable tokens are preserved.

\paragraph{Inference latency analysis}
We further compare the online inference latency of the discrete diffusion policy decoder and an autoregressive action decoder. 
For the discrete diffusion policy, we use the \texttt{replace\_confidence} scheduling strategy and evaluate different editing rounds, including 1, 2, 3, 5, 6, 10, 15, and 30 rounds. 
For the autoregressive baseline, we implement an AR policy decoder that generates the future action sequence sequentially. 
To make the comparison focus on the intrinsic decoding pattern, we disable engineering acceleration tricks for both methods: the discrete diffusion decoder is evaluated without additional decoding optimizations, and the AR decoder is evaluated without KV-cache acceleration. 
All latency measurements are conducted for online inference on NVIDIA H20 GPUs.

As shown in Fig.~\ref{fig:viz_pdm_anchor}-c, discrete diffusion decoding achieves substantially lower latency than the AR decoder under a moderate number of editing rounds. 
This is because action tokens within each discrete diffusion editing round are decoded in parallel, while the AR decoder must generate the action sequence sequentially. 
Although the latency of discrete diffusion increases with the number of editing rounds, it remains more efficient than AR decoding for common low- and medium-round settings. 
This result highlights the computational advantage of parallel policy-token editing.

\begin{table*}[!h]
\centering

\begin{minipage}[!h]{0.4\textwidth}
\vspace{0pt}
\centering
{
\renewcommand{\arraystretch}{1.25}
\resizebox{\linewidth}{!}{
\begin{tabular}{l | c c c c}
\toprule
\rowcolor[HTML]{FFE0CC}
\textbf{Schedule} 
& \textbf{Round}
& \(\boldsymbol{x=1/R}\)
& \(\boldsymbol{E_{s,R}}\) 
& \(\boldsymbol{y}\) \\
\midrule
Full replace        
& R1 
& \(1.000\) 
& \(0.67325\) 
& \(0.000\) \\
Full replace        
& R3 
& \(0.333\) 
& \(0.67853\) 
& \(-5.275\) \\\midrule
Confidence replace  
& R1 
& \(1.000\) 
& \(0.66950\) 
& \(3.747\) \\
Confidence replace  
& R2 
& \(0.500\) 
& \(0.66643\) 
& \(6.825\) \\
Confidence replace  
& R3 
& \(0.333\) 
& \(\mathbf{0.66488}\) 
& \(\mathbf{8.375}\) \\\midrule
JS-entropy replace 
& R2 
& \(0.500\) 
& \(0.66710\) 
& \(6.150\) \\
JS-entropy replace 
& R3 
& \(0.333\) 
& \(0.66690\) 
& \(6.350\) \\\midrule
JS-freeze replace  
& R4 
& \(0.250\) 
& \(0.67695\) 
& \(-3.700\) \\
JS-freeze replace  
& R6 
& \(0.167\) 
& \(0.66943\) 
& \(3.825\) \\
\bottomrule
\end{tabular}
}

}
\captionof{table}{\textbf{Compute performance of different re-edit schedules.}
The compute coordinate is defined as \(x=1/R\), where \(R\) is the scheduling round. The performance coordinate is the baseline-relative mean-L2 improvement,
\(y=1000(E_{\mathrm{FR},1}-E_{s,R})\), where \(E_{\mathrm{FR},1}\) denotes the mean L2 error of Full replace at \(R=1\), and \(E_{s,R}\) denotes the mean L2 error of schedule \(s\) at round \(R\).}
\label{tab:compute_performance_coordinates}
\end{minipage}
\hfill
\begin{minipage}[!h]{0.58\textwidth}
\vspace{0pt}
\centering
\resizebox{\linewidth}{!}{
\begin{tabular}{l | c c c c c}
\toprule
\rowcolor[HTML]{FFE0CC}
\textbf{Schedule} 
& \textbf{Rounds} 
& \(\boldsymbol{\Delta \mathrm{L2}@1s}\)
& \(\boldsymbol{\Delta \mathrm{L2}@2s}\)
& \(\boldsymbol{\Delta \mathrm{L2}@3s}\)
& \(\boldsymbol{\Delta \mathrm{L2}@4s}\) \\
\midrule
Full replace        
& R3--R1 
& \(+0.0019\) 
& \(+0.0040\) 
& \(+0.0068\) 
& \(+0.0084\) \\
Confidence replace  
& R3--R1 
& \(-0.0001\) 
& \(-0.0007\) 
& \(-0.0014\) 
& \(-0.0040\) \\
JS-entropy replace 
& R3--R2 
& \(-0.0001\) 
& \(-0.0003\) 
& \(-0.0003\) 
& \(-0.0001\) \\
JS-freeze replace  
& R6--R4 
& \(\mathbf{-0.0022}\) 
& \(\mathbf{-0.0055}\) 
& \(\mathbf{-0.0073}\) 
& \(\mathbf{-0.0151}\) \\
\bottomrule
\end{tabular}
}
\captionof{table}{\textbf{Effect of re-edit scheduling rounds on L2 trajectory error.} Negative values indicate that additional edit rounds reduce L2 error.}
\label{tab:round_l2_delta}


\centering
\resizebox{\linewidth}{!}{
\begin{tabular}{l | c c | c c c}
\toprule
\rowcolor[HTML]{FFE0CC}
\textbf{Ablations} &
\textbf{$a_x$ Err. (m/s$^2$)$\downarrow$} &
\textbf{$a_y$ Err. (m/s$^2$)$\downarrow$} &
\textbf{Traj. $x$ Err. (m)$\downarrow$} &
\textbf{Traj. $y$ Err. (m)$\downarrow$} &
\textbf{Traj. Err. (m)$\downarrow$} \\
\midrule
Upper-only
& 0.477
& 0.918
& 0.274
& 0.652
& 0.780 \\
Upper-masked
& 0.476
& 0.916
& 0.268
& 0.647
& 0.772 \\
Full image
& \textbf{0.475}
& \textbf{0.914}
& \textbf{0.267}
& \textbf{0.646}
& \textbf{0.770} \\
\bottomrule
\end{tabular}
}
\captionof{table}{\textbf{Effect of vertical image-region ablations on policy prediction.}
We compare three front-view image settings: Full image keeps the original front-view image unchanged, Upper-only keeps only the upper one-third region, and Upper-masked masks out the upper one-third region. Acceleration errors are reported in m/s$^2$, and trajectory position errors are reported in meters. Lower values indicate better performance.}
\label{tab:image_region_ablation}
\end{minipage}
\end{table*}


We emphasize that this comparison reflects the theoretical decoding-complexity difference between parallel editing and sequential generation under a controlled non-accelerated setting. 
In real deployment, AR decoding can benefit from KV-cache acceleration, whereas discrete diffusion does not use the same acceleration mechanism. 
Therefore, the measured latency should not be interpreted as a complete deployment-level speed comparison, but rather as an analysis of the intrinsic efficiency of the two decoding paradigms.

\subsection{Qualitative Results}
\label{sec:qualitative_results}
\begin{figure}[!h]
    \centering
    \includegraphics[width=\linewidth]{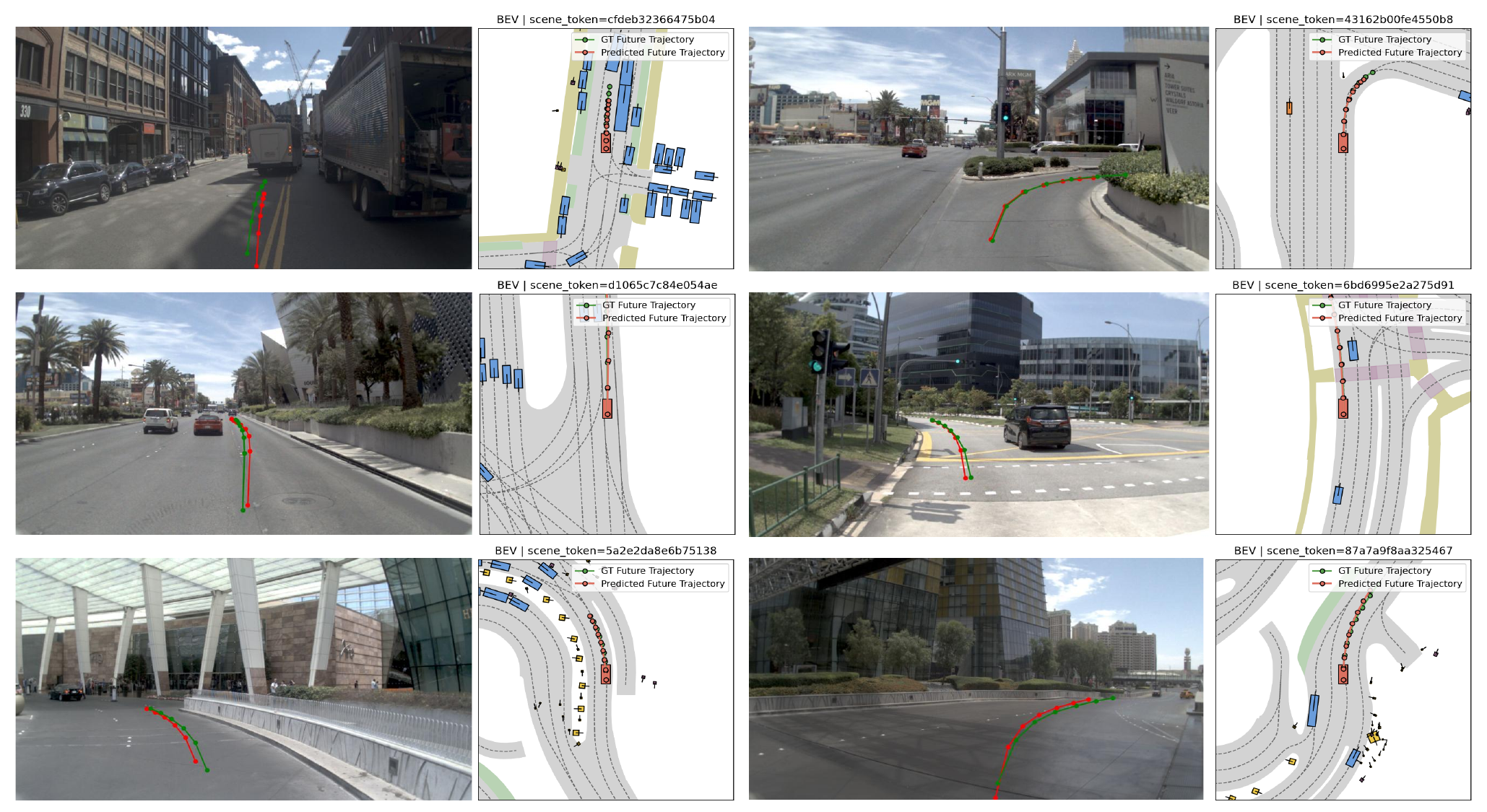}
    \caption{\textbf{Planning Performance of \XFM{}.}
    \XFM{} demonstrate strong planning performance on various driving scenarios, including nudging, lane changing, cruising, or pull-away.}
    \label{fig:planning}
\end{figure}

\paragraph{Planning results} Fig~\ref{fig:planning} presents qualitative planning results of \XFM{}. The predicted trajectories closely follow the expert demonstrations while remaining geometrically consistent with the underlying road topology. In straight-road cruising scenarios, the planner maintains stable lane centering and accurately captures the intended longitudinal progression. In turning and curved-road scenarios,  \XFM{} generates smooth trajectories that align well with lane boundaries and preserve appropriate curvature throughout the maneuver. Notably, in more complex urban scenes such as lane-changing or nudging, \XFM{} produces feasible future plans without explicit rule-based constraints.

\begin{figure}[!h]
    \centering
    \includegraphics[width=\linewidth]{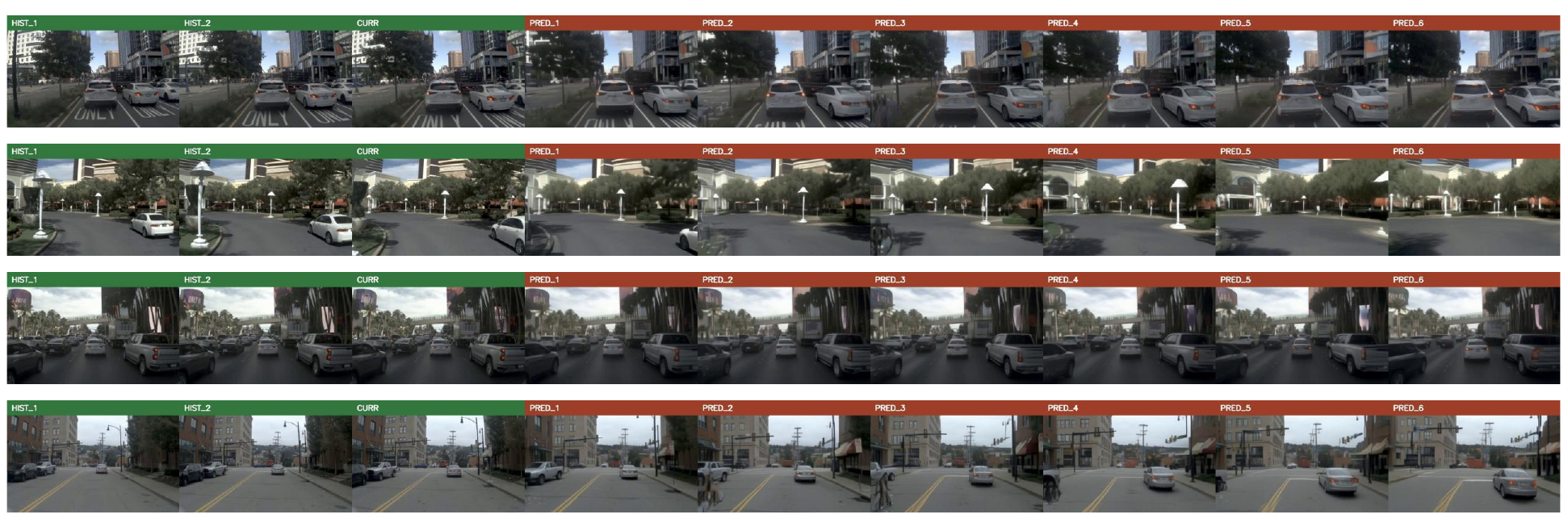}
    \caption{\textbf{World generation result of \XFM{}.}
    \XFM{} delivers coherent generation under a variety of driving scenarios.}
    \label{fig:world_model}
\end{figure}

\paragraph{World generation results} Fig~\ref{fig:world_model} visualizes the future world generation results of \XFM{} across diverse driving scenarios. Given historical observations and the current frame, \XFM{} generates temporally coherent future visual states over multiple prediction steps. The generated sequences preserve scene layout, road geometry, surrounding vehicles, and ego-motion consistency, while capturing realistic forward evolution under different urban driving conditions. These results demonstrate that the proposed discrete vision-action token editing framework can effectively model action-conditioned scene dynamics for world generation.

\paragraph{Attention map analysis}
We analyze the visual grounding behavior of the policy decoder through attention map visualization. 
The model contains 18 Transformer layers, 16 attention heads, and 8 key-value heads. 
All attention maps are computed from the first editing round under the \texttt{replace\_confidence} inference setting. 
Since each policy prediction contains 40 future action tokens, we average the attention maps over policy queries, attention heads, and layers unless otherwise specified.

\begin{figure*}[!h]
    \centering
å    \includegraphics[width=1.0\linewidth]{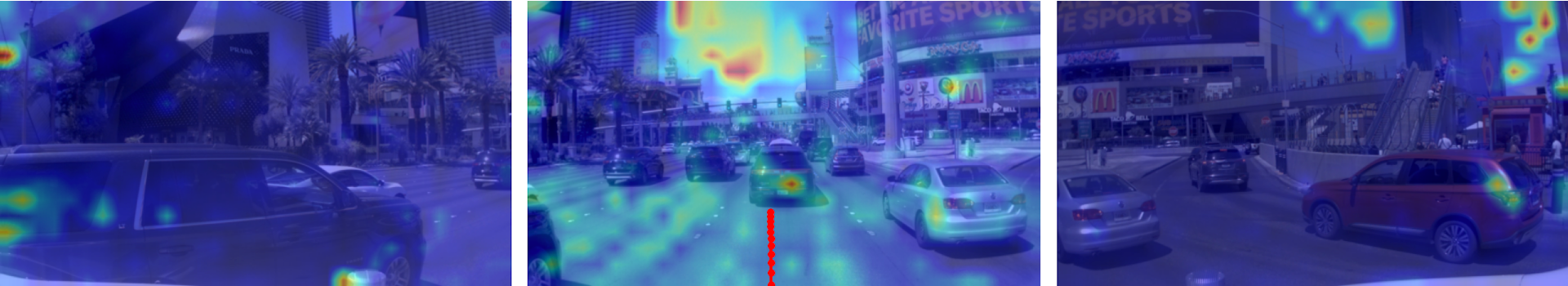}
    \caption{\textbf{Averaged policy attention maps.}
    The three panels show the left-view, front-view, and right-view cameras, respectively. 
    Attention maps are averaged over Transformer layers, attention heads, and policy action queries from the first editing round. 
    The policy decoder attends to driving-relevant regions such as lanes, vehicles, road structures, and traffic signs, while also showing stable activation in upper sky regions.}
    \label{fig:attn_avg}
    \vspace{-0.2cm}
\end{figure*}

Fig.~\ref{fig:attn_avg} shows the averaged attention maps for the front-view and side-view cameras. 
The policy decoder attends to driving-relevant semantic regions, including roads, lane markings, surrounding vehicles, traffic signs, and other dynamic or structural cues. 
This indicates that action-token prediction is grounded in visual scene semantics rather than only low-level image appearance. 
Meanwhile, we observe consistent high attention in sky regions, especially in the front-view image.

\begin{figure*}[!h]
    \centering
    \includegraphics[width=1.0\linewidth]{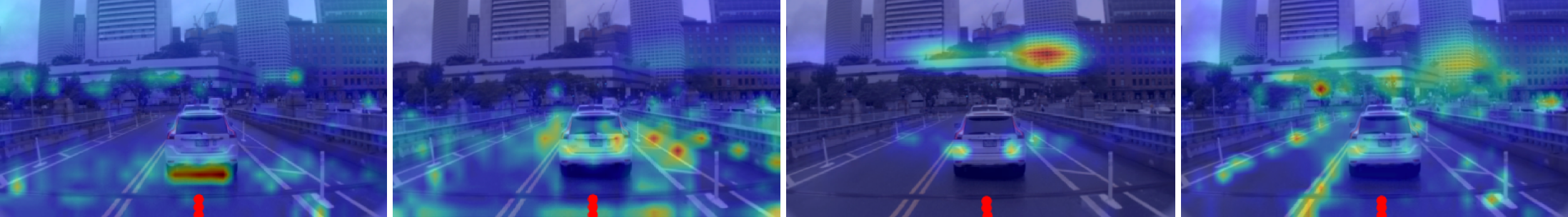}
    \caption{\textbf{Layer-wise policy attention maps.}
    All panels show the front-view camera. 
    We visualize layers 0, 6, 12, and 17 by averaging over attention heads and policy queries. 
    Different layers emphasize different spatial and semantic structures, while the upper-region activation remains visible across multiple layers, suggesting a stable attention pattern rather than an isolated layer-specific artifact.}
    \label{fig:attn_layers}
    \vspace{-0.2cm}
\end{figure*}

\begin{figure}[!h]
    \centering
    \includegraphics[width=\linewidth]{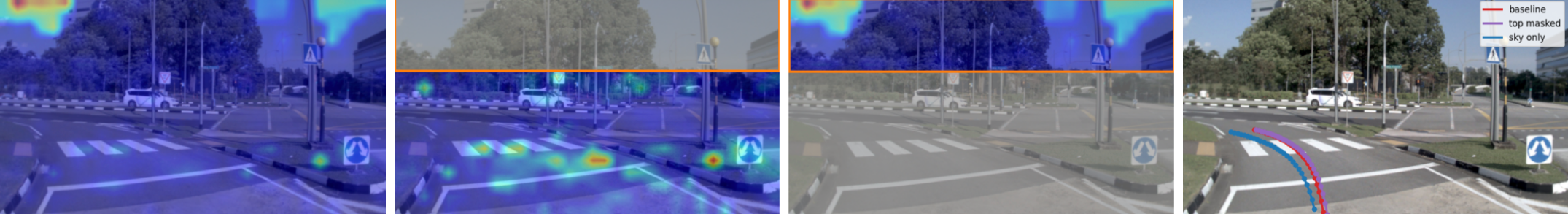}
    \caption{\textbf{Upper-region ablation for policy attention.}
    We compare the original front-view image, an upper-masked image where the top one-third region is masked out, and an upper-only image where only the top one-third region is preserved. Masking the upper region redistributes attention toward local driving semantics, whereas the upper-only setting can still produce plausible but less accurate trajectories, suggesting that upper-region tokens may provide global contextual cues while lower regions remain essential for detailed policy prediction.}
    \label{fig:attn_sky_ablation}
    \vspace{-0.2cm}
\end{figure}

To inspect this behavior across the network depth, we visualize layer-wise attention maps by selecting layers 0, 6, 12, and 17, while averaging over heads and policy queries. 
As shown in Fig.~\ref{fig:attn_layers}, different layers emphasize different levels of spatial and semantic abstraction. 
Early layers show broader attention over the scene, while deeper layers produce more structured responses on lanes, road boundaries, vehicles, and global scene layout. 
The sky activation remains visible across multiple layers, suggesting that it is a stable attention pattern rather than an artifact of a single layer or head.

We hypothesize that sky patches may act as implicit global anchors in the absence of explicit CLS or register tokens. 
Sky regions are spatially stable, visually smooth, and low in local texture, making them suitable locations for absorbing attention mass or organizing global scene context. 
They may also encode weak global cues such as illumination, weather, horizon position, or scene openness. 
Thus, sky attention may reflect a mixture of attention-sink behavior and implicit-register behavior, rather than direct reliance on sky pixels as causal driving evidence.

To examine this hypothesis, we conduct a sky-region ablation on the front-view image. 
Since sky regions typically occupy the upper part of the image, we approximate the sky region with the top one-third of the front-view image. 
We compare three settings: the original image, a top-masked image where the upper one-third region is masked out, and a sky-only image where only the upper one-third region is preserved. 
This is an approximate intervention because the sky is not guaranteed to always lie exactly in the top one-third region, but it provides a controlled way to study the observed attention pattern.

\begin{figure}[!h]
    \centering
    \includegraphics[width=\linewidth]{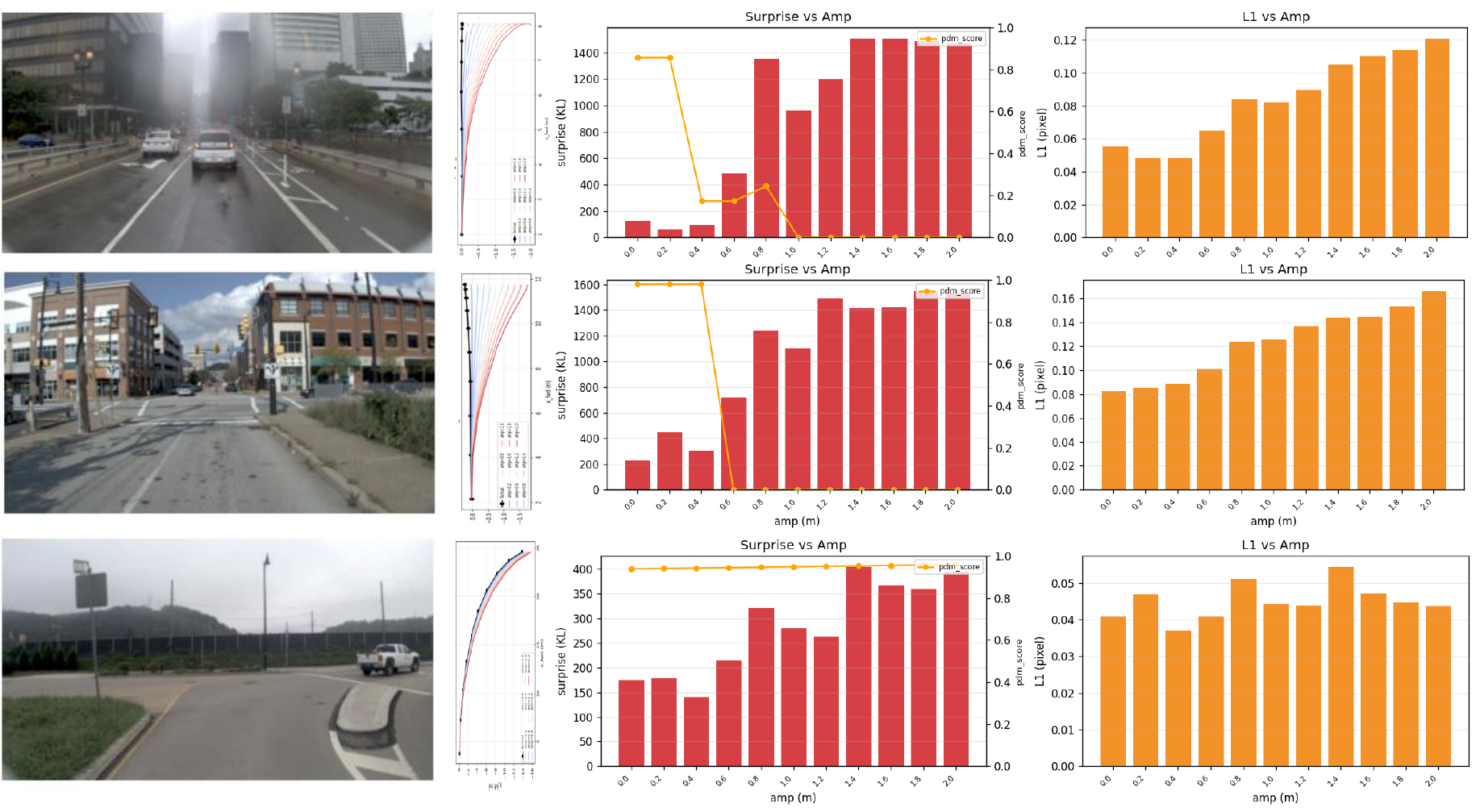}
    \caption{\textbf{Counterfactual result with surprise metric.}
    We compare surprise value $\mathbf{S}$ by both factual and counterfactual world-model generations. A clear correlation is observed between surprise and PDMS, suggesting that \XFM{} captures safety-critical scene dynamics such as collisions and drivable-area violations}
    \label{fig:surprise_ablation}
    \vspace{-0.5cm}
\end{figure}
Fig.~\ref{fig:attn_sky_ablation} shows the ablation results. 
When the upper region is masked, attention is redistributed toward local driving semantics such as road boundaries, lane markings, vehicles, and traffic signs. 
This confirms that the sky region absorbs a non-trivial amount of attention in the original input. 
When only the upper region is preserved, the model can still produce plausible trajectories, but the trajectory quality degrades compared with the full-image baseline. 
This suggests that sky-region tokens provide useful global context or implicit anchoring, while lower image regions provide the detailed local semantics required for accurate policy prediction. The quantitative results in Table~\ref{tab:image_region_ablation} show the same trend: masking the upper region slightly degrades trajectory quality, while using only the upper region gives the worst performance. 
This indicates that sky-region tokens may provide useful global context, but local driving semantics from the lower image region remain essential for accurate policy prediction. Overall, the attention visualizations show that the policy decoder uses both local driving semantics and global contextual regions. 
The sky activation should be interpreted as stable attention-sink or implicit-register behavior, not as direct evidence that sky pixels are strong causal cues for driving. 
Additional attention map examples are provided in Appendix~\ref{additional_qualitative_results}.

\paragraph{Counterfactual results}
To manifest the causal understandability, we evaluate the surprise value of \XFM{} under both factual rollouts and a spectrum of counterfactual world-model generations. 
Specifically, the world model is injected with both ground-truth action $\mathbf{A}_{t:t+H}$ condition or counterfactual ones given a sweep of lateral perturbed actions $\mathbf{\tilde{A}}_{t:t+H}$. Surprise is quantified by $\mathbf{S}=\mathrm{KL}(p_\theta(\cdot|\mathbf{A}_{t:t+H},\mathbf{V}_{t})||p_\theta(\cdot|\mathbf{\tilde{A}}_{t:t+H},\mathbf{V}_{t}))$~\cite{duncan1977quantifying}. We also report the average pixel L1 distance $\Delta_\mathrm{img}$ for reference. A strong negative correlation is observed between surprise and PDMS. As in Fig.\ref{fig:surprise_ablation}, the surge of $\mathbf{S}$ when PDMS drop to zero (collide with static object, drive off the road boundary) indicating that \XFM{} captures action-conditioned scene dynamics and safety-critical outcomes. As driving quality deteriorates due to drivable-area violations, or unsafe interactions, the resulting future observations become increasingly difficult for the world model to predict, leading to higher surprise values. While the gradual increase for $\Delta_\mathrm{img}$ further indicate the casualty learned under prediction error. Conversely, when no hard safety penalties are incurred (PDMS = 1), surprise remains consistently low across both factual and counterfactual rollouts. This suggests that surprise provides a meaningful proxy for model uncertainty and can potentially be leveraged for risk-aware planning and safety evaluation.

\section{Conclusion and Future Directions}

In this work, we present \XFM{}, a unified discrete vision-action world-policy framework for autonomous driving. 
Rather than treating world prediction and policy generation as separate modules, \XFM{} formulates visual observations, future states, driving decisions, and ego actions within a shared discrete token space. 
This design enables representation alignment across modalities through shared sequence modeling, shared Transformer computation, and unified token-level objectives. 
Built on this aligned discrete space, \XFM{} further integrates world modeling, world-policy modeling, and policy modeling through multi-task and multi-stage training, allowing future prediction to be optimized toward policy-relevant action generation rather than isolated visual reconstruction. A key component of \XFM{} is its structured policy modeling design. 
By introducing a high-level decision token as a latent planning skeleton, the model separates multi-modal decision selection from low-level trajectory realization. 
This hierarchical formulation reduces residual dependence among dense future action tokens and makes parallel discrete diffusion editing more suitable for planning. 
Together with confidence-based re-edit scheduling, \XFM{} provides an efficient and structured mechanism for generating temporally consistent future actions. 
Experiments on autonomous-driving benchmarks demonstrate strong planning performance, while additional analyses on world-model surprise, attention grounding, scheduling dynamics, and inference latency show that the learned world-policy representation supports counterfactual evaluation and policy-oriented future reasoning.

More broadly, we view \XFM{} as an initial validation of a physical AI design paradigm based on discrete representation alignment, unified world-policy training, and hierarchical policy construction with parallel token editing. Future work will incorporate language-based reasoning, explore adaptive serial-parallel generation schedules, and leverage intrinsic value functions learned from world models for self-supervised reinforcement learning. We will further extend this paradigm to broader autonomous-driving and robotic datasets, including larger-scale in-house driving data, multi-modal reasoning, and interactive field tests. We also plan to release additional evaluation results as these deployments mature, enabling a more comprehensive study of how discrete world-policy alignment can support scalable embodied decision-making.
\clearpage

\bibliographystyle{plainnat}
\bibliography{main}

\begin{thebibliography}{103}
\providecommand{\natexlab}[1]{#1}
\providecommand{\url}[1]{\texttt{#1}}
\expandafter\ifx\csname urlstyle\endcsname\relax
  \providecommand{\doi}[1]{doi: #1}\else
  \providecommand{\doi}{doi: \begingroup \urlstyle{rm}\Url}\fi

\bibitem[Austin et~al.(2021)Austin, Johnson, Ho, Tarlow, and Van Den~Berg]{austin2021structured}
Jacob Austin, Daniel~D Johnson, Jonathan Ho, Daniel Tarlow, and Rianne Van Den~Berg.
\newblock Structured denoising diffusion models in discrete state-spaces.
\newblock \emph{Advances in neural information processing systems}, 34:\penalty0 17981--17993, 2021.

\bibitem[Azzolini et~al.(2025)Azzolini, Bai, Brandon, Cao, Chattopadhyay, Chen, Chu, Cui, Diamond, Ding, et~al.]{azzolini2025cosmos}
Alisson Azzolini, Junjie Bai, Hannah Brandon, Jiaxin Cao, Prithvijit Chattopadhyay, Huayu Chen, Jinju Chu, Yin Cui, Jenna Diamond, Yifan Ding, et~al.
\newblock Cosmos-reason1: From physical common sense to embodied reasoning.
\newblock \emph{arXiv preprint arXiv:2503.15558}, 2025.

\bibitem[Bartoccioni et~al.(2025)Bartoccioni, Ramzi, Besnier, Venkataramanan, Vu, Xu, Chambon, Gidaris, Odabas, Hurych, et~al.]{bartoccioni2025vavim}
Florent Bartoccioni, Elias Ramzi, Victor Besnier, Shashanka Venkataramanan, Tuan-Hung Vu, Yihong Xu, Loick Chambon, Spyros Gidaris, Serkan Odabas, David Hurych, et~al.
\newblock Vavim and vavam: Autonomous driving through video generative modeling.
\newblock \emph{arXiv preprint arXiv:2502.15672}, 2025.

\bibitem[Ben-Hamu et~al.(2025)Ben-Hamu, Gat, Severo, Nolte, and Karrer]{benhamu2025accelerated}
Heli Ben-Hamu, Itai Gat, Daniel Severo, Niklas Nolte, and Brian Karrer.
\newblock Accelerated sampling from masked diffusion models via entropy bounded unmasking.
\newblock In \emph{Advances in Neural Information Processing Systems}, 2025.

\bibitem[Bi et~al.(2026)Bi, Tan, Xie, Wang, Huang, Liu, Zhao, Feng, Xiang, Rong, et~al.]{bi2026motus}
Hongzhe Bi, Hengkai Tan, Shenghao Xie, Zeyuan Wang, Shuhe Huang, Haitian Liu, Ruowen Zhao, Yao Feng, Chendong Xiang, Yinze Rong, et~al.
\newblock Motus: A unified latent action world model.
\newblock In \emph{Proceedings of the IEEE/CVF Conference on Computer Vision and Pattern Recognition}, pages 35101--35113, 2026.

\bibitem[Caesar et~al.(2021)Caesar, Kabzan, Tan, Fong, Wolff, Lang, Fletcher, Beijbom, and Omari]{caesar2021nuplan}
Holger Caesar, Juraj Kabzan, Kok~Seang Tan, Whye~Kit Fong, Eric Wolff, Alex Lang, Luke Fletcher, Oscar Beijbom, and Sammy Omari.
\newblock nuplan: A closed-loop ml-based planning benchmark for autonomous vehicles.
\newblock \emph{arXiv preprint arXiv:2106.11810}, 2021.

\bibitem[Cai and Li(2026)]{cai2026confidence}
Changxiao Cai and Gen Li.
\newblock Confidence-based decoding is provably efficient for diffusion language models.
\newblock \emph{arXiv preprint arXiv:2603.22248}, 2026.

\bibitem[Cao et~al.(2025)Cao, Hallgarten, Li, Dauner, Gu, Wang, Miron, Aiello, Li, Gilitschenski, et~al.]{cao2025pseudo}
Wei Cao, Marcel Hallgarten, Tianyu Li, Daniel Dauner, Xunjiang Gu, Caojun Wang, Yakov Miron, Marco Aiello, Hongyang Li, Igor Gilitschenski, et~al.
\newblock Pseudo-simulation for autonomous driving.
\newblock \emph{arXiv preprint arXiv:2506.04218}, 2025.

\bibitem[Cen et~al.(2025)Cen, Yu, Yuan, Jiang, Huang, Guo, Li, Song, Luo, Wang, et~al.]{cen2025worldvla}
Jun Cen, Chaohui Yu, Hangjie Yuan, Yuming Jiang, Siteng Huang, Jiayan Guo, Xin Li, Yibing Song, Hao Luo, Fan Wang, et~al.
\newblock Worldvla: Towards autoregressive action world model.
\newblock \emph{arXiv preprint arXiv:2506.21539}, 2025.

\bibitem[Chen et~al.(2026)Chen, Liu, Yan, Han, Zhang, Gu, Gao, Guo, Qian, Wang, et~al.]{chen2026last}
Hao Chen, Jiaming Liu, Zhonghao Yan, Nuowei Han, Renrui Zhang, Chenyang Gu, Jialin Gao, Ziyu Guo, Siyuan Qian, Yinxi Wang, et~al.
\newblock Last-r1: Reinforcing action via adaptive physical latent reasoning for vla models.
\newblock \emph{arXiv preprint arXiv:2604.28192}, 2026.

\bibitem[Chen et~al.(2024{\natexlab{a}})Chen, Wu, Chitta, Jaeger, Geiger, and Li]{chen2024end}
Li~Chen, Penghao Wu, Kashyap Chitta, Bernhard Jaeger, Andreas Geiger, and Hongyang Li.
\newblock End-to-end autonomous driving: Challenges and frontiers.
\newblock \emph{IEEE Transactions on Pattern Analysis and Machine Intelligence}, 46\penalty0 (12):\penalty0 10164--10183, 2024{\natexlab{a}}.

\bibitem[Chen et~al.(2022)Chen, Li, Huang, Li, Xing, Tian, Li, Hu, Na, Li, et~al.]{chen2022milestones}
Long Chen, Yuchen Li, Chao Huang, Bai Li, Yang Xing, Daxin Tian, Li~Li, Zhongxu Hu, Xiaoxiang Na, Zixuan Li, et~al.
\newblock Milestones in autonomous driving and intelligent vehicles: Survey of surveys.
\newblock \emph{IEEE Transactions on Intelligent Vehicles}, 8\penalty0 (2):\penalty0 1046--1056, 2022.

\bibitem[Chen et~al.(2024{\natexlab{b}})Chen, Jiang, Gao, Liao, Xu, Zhang, Huang, Liu, and Wang]{chen2024vadv2}
Shaoyu Chen, Bo~Jiang, Hao Gao, Bencheng Liao, Qing Xu, Qian Zhang, Chang Huang, Wenyu Liu, and Xinggang Wang.
\newblock Vadv2: End-to-end vectorized autonomous driving via probabilistic planning.
\newblock \emph{arXiv preprint arXiv:2402.13243}, 2024{\natexlab{b}}.

\bibitem[Chen et~al.(2025)Chen, Cong, and Li]{chen2025optimal}
Sitan Chen, Kevin Cong, and Jerry Li.
\newblock Optimal inference schedules for masked diffusion models.
\newblock \emph{arXiv preprint arXiv:2511.04647}, 2025.

\bibitem[Chen et~al.(2024{\natexlab{c}})Chen, Ye, Xu, Cao, and Chen]{chen2024ppad}
Zhili Chen, Maosheng Ye, Shuangjie Xu, Tongyi Cao, and Qifeng Chen.
\newblock Ppad: Iterative interactions of prediction and planning for end-to-end autonomous driving.
\newblock In \emph{European Conference on Computer Vision}, pages 239--256. Springer, 2024{\natexlab{c}}.

\bibitem[Chitta et~al.(2022)Chitta, Prakash, Jaeger, Yu, Renz, and Geiger]{chitta2022transfuser}
Kashyap Chitta, Aditya Prakash, Bernhard Jaeger, Zehao Yu, Katrin Renz, and Andreas Geiger.
\newblock Transfuser: Imitation with transformer-based sensor fusion for autonomous driving.
\newblock \emph{IEEE transactions on pattern analysis and machine intelligence}, 45\penalty0 (11):\penalty0 12878--12895, 2022.

\bibitem[Dauner et~al.(2024)Dauner, Hallgarten, Li, Weng, Huang, Yang, Li, Gilitschenski, Ivanovic, Pavone, et~al.]{dauner2024navsim}
Daniel Dauner, Marcel Hallgarten, Tianyu Li, Xinshuo Weng, Zhiyu Huang, Zetong Yang, Hongyang Li, Igor Gilitschenski, Boris Ivanovic, Marco Pavone, et~al.
\newblock Navsim: Data-driven non-reactive autonomous vehicle simulation and benchmarking.
\newblock \emph{Advances in Neural Information Processing Systems}, 37:\penalty0 28706--28719, 2024.

\bibitem[Duncan-Johnson and Donchin(1977)]{duncan1977quantifying}
Carolyn~C. Duncan-Johnson and Emanuel Donchin.
\newblock On quantifying surprise: The variation of event-related potentials with subjective probability.
\newblock \emph{Psychophysiology}, 14\penalty0 (5):\penalty0 456--467, 1977.

\bibitem[Feng et~al.(2025)Feng, Geng, Guan, Wu, Wang, and He]{feng2025theoretical}
Guhao Feng, Yihan Geng, Jian Guan, Wei Wu, Liwei Wang, and Di~He.
\newblock Theoretical benefit and limitation of diffusion language model.
\newblock In \emph{Advances in Neural Information Processing Systems}, 2025.

\bibitem[Gao et~al.(2026{\natexlab{a}})Gao, Chen, Jiang, Liao, Shi, Guo, Pu, Li, Liu, Zhang, et~al.]{gao2026rad}
Hao Gao, Shaoyu Chen, Bo~Jiang, Bencheng Liao, Yiang Shi, Xiaoyang Guo, Yuechuan Pu, Xiangyu Li, Wenyu Liu, Qian Zhang, et~al.
\newblock Rad: Training an end-to-end driving policy via large-scale 3dgs-based reinforcement learning.
\newblock \emph{Advances in Neural Information Processing Systems}, 38:\penalty0 32551--32576, 2026{\natexlab{a}}.

\bibitem[Gao et~al.(2024)Gao, Yang, Chen, Chitta, Qiu, Geiger, Zhang, and Li]{gao2024vista}
Shenyuan Gao, Jiazhi Yang, Li~Chen, Kashyap Chitta, Yihang Qiu, Andreas Geiger, Jun Zhang, and Hongyang Li.
\newblock Vista: A generalizable driving world model with high fidelity and versatile controllability.
\newblock \emph{Advances in Neural Information Processing Systems}, 37:\penalty0 91560--91596, 2024.

\bibitem[Gao et~al.(2026{\natexlab{b}})Gao, Liang, Zheng, Malik, Ye, Yu, Tseng, Dong, Mo, Lin, et~al.]{gao2026dreamdojo}
Shenyuan Gao, William Liang, Kaiyuan Zheng, Ayaan Malik, Seonghyeon Ye, Sihyun Yu, Wei-Cheng Tseng, Yuzhu Dong, Kaichun Mo, Chen-Hsuan Lin, et~al.
\newblock Dreamdojo: A generalist robot world model from large-scale human videos.
\newblock \emph{arXiv preprint arXiv:2602.06949}, 2026{\natexlab{b}}.

\bibitem[Guan et~al.(2024)Guan, Liao, Li, Hu, Yuan, Zhang, and Xu]{guan2024world}
Yanchen Guan, Haicheng Liao, Zhenning Li, Jia Hu, Runze Yuan, Guohui Zhang, and Chengzhong Xu.
\newblock World models for autonomous driving: An initial survey.
\newblock \emph{IEEE Transactions on Intelligent Vehicles}, 2024.

\bibitem[Guo et~al.(2025)Guo, Liu, Wu, Pan, and Lv]{guo2025ipad}
Ke~Guo, Haochen Liu, Xiaojun Wu, Jia Pan, and Chen Lv.
\newblock ipad: Iterative proposal-centric end-to-end autonomous driving.
\newblock \emph{arXiv preprint arXiv:2505.15111}, 2025.

\bibitem[Hagedorn et~al.(2024)Hagedorn, Hallgarten, Stoll, and Condurache]{hagedorn2024integration}
Steffen Hagedorn, Marcel Hallgarten, Martin Stoll, and Alexandru~Paul Condurache.
\newblock The integration of prediction and planning in deep learning automated driving systems: A review.
\newblock \emph{IEEE Transactions on Intelligent Vehicles}, 10\penalty0 (5):\penalty0 3626--3643, 2024.

\bibitem[Hu et~al.(2023{\natexlab{a}})Hu, Russell, Yeo, Murez, Fedoseev, Kendall, Shotton, and Corrado]{hu2023gaia}
Anthony Hu, Lloyd Russell, Hudson Yeo, Zak Murez, George Fedoseev, Alex Kendall, Jamie Shotton, and Gianluca Corrado.
\newblock Gaia-1: A generative world model for autonomous driving.
\newblock \emph{arXiv preprint arXiv:2309.17080}, 2023{\natexlab{a}}.

\bibitem[Hu et~al.(2024)Hu, Yin, Jia, Deng, Guo, Zhang, Long, and Tan]{hu2024drivingworld}
Xiaotao Hu, Wei Yin, Mingkai Jia, Junyuan Deng, Xiaoyang Guo, Qian Zhang, Xiaoxiao Long, and Ping Tan.
\newblock Drivingworld: Constructing world model for autonomous driving via video gpt.
\newblock \emph{arXiv preprint arXiv:2412.19505}, 2024.

\bibitem[Hu et~al.(2023{\natexlab{b}})Hu, Yang, Chen, Li, Sima, Zhu, Chai, Du, Lin, Wang, et~al.]{hu2023planning}
Yihan Hu, Jiazhi Yang, Li~Chen, Keyu Li, Chonghao Sima, Xizhou Zhu, Siqi Chai, Senyao Du, Tianwei Lin, Wenhai Wang, et~al.
\newblock Planning-oriented autonomous driving.
\newblock In \emph{Proceedings of the IEEE/CVF conference on computer vision and pattern recognition}, pages 17853--17862, 2023{\natexlab{b}}.

\bibitem[Huang et~al.(2026{\natexlab{a}})Huang, Zhang, Huang, Wang, Mao, Chua, Chen, Chen, and Lv]{huang2026automot}
Wenhui Huang, Songyan Zhang, Qihang Huang, Zhidong Wang, Zhiqi Mao, Collister Chua, Zhan Chen, Long Chen, and Chen Lv.
\newblock Automot: A unified vision-language-action model with asynchronous mixture-of-transformers for end-to-end autonomous driving.
\newblock \emph{arXiv preprint arXiv:2603.14851}, 2026{\natexlab{a}}.

\bibitem[Huang et~al.(2026{\natexlab{b}})Huang, Zhu, Lu, Huang, Zhang, Chen, Dai, Xie, and Li]{huang2026mindvla}
Yuzhou Huang, Benjin Zhu, Hengtong Lu, Victor Shea-Jay Huang, Haiming Zhang, Wei Chen, Jifeng Dai, Yan Xie, and Hongsheng Li.
\newblock Mindvla-u1: Vla beats va with unified streaming architecture for autonomous driving.
\newblock \emph{arXiv preprint arXiv:2605.12624}, 2026{\natexlab{b}}.

\bibitem[Huang et~al.(2023)Huang, Liu, and Lv]{huang2023gameformer}
Zhiyu Huang, Haochen Liu, and Chen Lv.
\newblock Gameformer: Game-theoretic modeling and learning of transformer-based interactive prediction and planning for autonomous driving.
\newblock In \emph{Proceedings of the IEEE/CVF International Conference on Computer Vision}, pages 3903--3913, 2023.

\bibitem[Jiang et~al.(2025{\natexlab{a}})Jiang, Gao, Wang, Sun, Wang, Heng, Sun, Tang, Zhu, Chai, et~al.]{jiang2025irlvla}
Anqing Jiang, Yu~Gao, Yiru Wang, Zhigang Sun, Shuo Wang, Yuwen Heng, Hao Sun, Shichen Tang, Lijuan Zhu, Jinhao Chai, et~al.
\newblock Irl-vla: Training an vision-language-action policy via reward world model.
\newblock \emph{arXiv preprint arXiv:2508.06571}, 2025{\natexlab{a}}.

\bibitem[Jiang et~al.(2025{\natexlab{b}})Jiang, Huang, Qian, Luo, Zhu, Zhong, Tang, Kong, Wang, Jiao, et~al.]{jiang2025survey}
Sicong Jiang, Zilin Huang, Kangan Qian, Ziang Luo, Tianze Zhu, Yang Zhong, Yihong Tang, Menglin Kong, Yunlong Wang, Siwen Jiao, et~al.
\newblock A survey on vision-language-action models for autonomous driving.
\newblock In \emph{Proceedings of the IEEE/CVF International Conference on Computer Vision}, pages 4524--4536, 2025{\natexlab{b}}.

\bibitem[Karkus et~al.(2025)Karkus, Igl, Chen, Chitta, Packer, Douillard, Tian, Naumann, Garcia-Cobo, Tan, et~al.]{KarkusIglEtAl2025}
Peter Karkus, Maximilian Igl, Yuxiao Chen, Kashyap Chitta, Jef Packer, Bertrand Douillard, Ran Tian, Alexander Naumann, Guillermo Garcia-Cobo, Shuhan Tan, et~al.
\newblock Beyond behavior cloning in autonomous driving: a survey of closed-loop training techniques.
\newblock \emph{Authorea Preprints}, 2025.

\bibitem[Kim et~al.(2025)Kim, Shah, Kontonis, Kakade, and Chen]{kim2025train}
Jaeyeon Kim, Kulin Shah, Vasilis Kontonis, Sham~M. Kakade, and Sitan Chen.
\newblock Train for the worst, plan for the best: Understanding token ordering in masked diffusions.
\newblock In \emph{Proceedings of the 42nd International Conference on Machine Learning}, volume 267 of \emph{Proceedings of Machine Learning Research}, pages 30749--30768. PMLR, 2025.

\bibitem[Kim et~al.(2026)Kim, Gao, Lin, Lin, Ge, Lam, Liang, Song, Liu, Finn, et~al.]{kim2026cosmos}
Moo~Jin Kim, Yihuai Gao, Tsung-Yi Lin, Yen-Chen Lin, Yunhao Ge, Grace Lam, Percy Liang, Shuran Song, Ming-Yu Liu, Chelsea Finn, et~al.
\newblock Cosmos policy: Fine-tuning video models for visuomotor control and planning.
\newblock \emph{arXiv preprint arXiv:2601.16163}, 2026.

\bibitem[Kong et~al.(2025)Kong, Yang, Mei, Liu, Liang, Zhu, Lu, Yin, Hu, Jia, et~al.]{kong20253d}
Lingdong Kong, Wesley Yang, Jianbiao Mei, Youquan Liu, Ao~Liang, Dekai Zhu, Dongyue Lu, Wei Yin, Xiaotao Hu, Mingkai Jia, et~al.
\newblock 3d and 4d world modeling: A survey.
\newblock \emph{arXiv preprint arXiv:2509.07996}, 2025.

\bibitem[Lavenant and Zanella(2025)]{lavenant2025error}
Hugo Lavenant and Giacomo Zanella.
\newblock Error bounds and optimal schedules for masked diffusions with factorized approximations.
\newblock \emph{arXiv preprint arXiv:2510.25544}, 2025.

\bibitem[LeCun et~al.(2022)]{lecun2022path}
Yann LeCun et~al.
\newblock A path towards autonomous machine intelligence version 0.9. 2, 2022-06-27.
\newblock \emph{Open Review}, 62\penalty0 (1):\penalty0 1--62, 2022.

\bibitem[Li and Cai(2025)]{li2025convergence}
Gen Li and Changxiao Cai.
\newblock A convergence theory for diffusion language models: An information-theoretic perspective.
\newblock \emph{arXiv preprint arXiv:2505.21400}, 2025.

\bibitem[Li et~al.(2025{\natexlab{a}})Li, Zheng, Wang, Wang, Zhao, Liu, Zhan, Zhan, and Lang]{li2025discrete}
Pengxiang Li, Yinan Zheng, Yue Wang, Huimin Wang, Hang Zhao, Jingjing Liu, Xianyuan Zhan, Kun Zhan, and Xianpeng Lang.
\newblock Discrete diffusion for reflective vision-language-action models in autonomous driving.
\newblock \emph{arXiv preprint arXiv:2509.20109}, 2025{\natexlab{a}}.

\bibitem[Li et~al.(2025{\natexlab{b}})Li, Shang, Liu, Zhan, Wang, Wang, Chen, Wang, An, Tang, et~al.]{li2025drivevla}
Yingyan Li, Shuyao Shang, Weisong Liu, Bing Zhan, Haochen Wang, Yuqi Wang, Yuntao Chen, Xiaoman Wang, Yasong An, Chufeng Tang, et~al.
\newblock Drivevla-w0: World models amplify data scaling law in autonomous driving.
\newblock \emph{arXiv preprint arXiv:2510.12796}, 2025{\natexlab{b}}.

\bibitem[Li et~al.(2025{\natexlab{c}})Li, Wang, Liu, He, Fan, and Zhang]{li2025end}
Yingyan Li, Yuqi Wang, Yang Liu, Jiawei He, Lue Fan, and Zhaoxiang Zhang.
\newblock End-to-end driving with online trajectory evaluation via bev world model.
\newblock In \emph{Proceedings of the IEEE/CVF International Conference on Computer Vision}, pages 27137--27146, 2025{\natexlab{c}}.

\bibitem[Li et~al.(2025{\natexlab{d}})Li, Xiong, Guo, Li, Yan, Xu, Zhou, Chen, Sun, Wang, et~al.]{li2025recogdrive}
Yongkang Li, Kaixin Xiong, Xiangyu Guo, Fang Li, Sixu Yan, Gangwei Xu, Lijun Zhou, Long Chen, Haiyang Sun, Bing Wang, et~al.
\newblock Recogdrive: A reinforced cognitive framework for end-to-end autonomous driving.
\newblock \emph{arXiv preprint arXiv:2506.08052}, 2025{\natexlab{d}}.

\bibitem[Li et~al.(2026)Li, Zhou, Yan, Liao, Yan, Xiong, Chen, Xie, Wang, Chen, et~al.]{li2026unidrivevla}
Yongkang Li, Lijun Zhou, Sixu Yan, Bencheng Liao, Tianyi Yan, Kaixin Xiong, Long Chen, Hongwei Xie, Bing Wang, Guang Chen, et~al.
\newblock Unidrivevla: Unifying understanding, perception, and action planning for autonomous driving.
\newblock \emph{arXiv preprint arXiv:2604.02190}, 2026.

\bibitem[Li et~al.(2024)Li, Li, Wang, Lan, Yu, Ji, Li, Zhu, Kautz, Wu, et~al.]{li2024hydra}
Zhenxin Li, Kailin Li, Shihao Wang, Shiyi Lan, Zhiding Yu, Yishen Ji, Zhiqi Li, Ziyue Zhu, Jan Kautz, Zuxuan Wu, et~al.
\newblock Hydra-mdp: End-to-end multimodal planning with multi-target hydra-distillation.
\newblock \emph{arXiv preprint arXiv:2406.06978}, 2024.

\bibitem[Li et~al.(2025{\natexlab{e}})Li, Wang, Lan, Yu, Wu, and Alvarez]{Li_2025_ICCV}
Zhenxin Li, Shihao Wang, Shiyi Lan, Zhiding Yu, Zuxuan Wu, and Jose~M Alvarez.
\newblock Hydra-next: Robust closed-loop driving with open-loop training.
\newblock In \emph{Proceedings of the IEEE/CVF International Conference on Computer Vision}, pages 27305--27314, 2025{\natexlab{e}}.

\bibitem[Liao et~al.(2025)Liao, Chen, Yin, Jiang, Wang, Yan, Zhang, Li, Zhang, Zhang, et~al.]{liao2025diffusiondrive}
Bencheng Liao, Shaoyu Chen, Haoran Yin, Bo~Jiang, Cheng Wang, Sixu Yan, Xinbang Zhang, Xiangyu Li, Ying Zhang, Qian Zhang, et~al.
\newblock Diffusiondrive: Truncated diffusion model for end-to-end autonomous driving.
\newblock In \emph{Proceedings of the Computer Vision and Pattern Recognition Conference}, pages 12037--12047, 2025.

\bibitem[Lin et~al.(2025{\natexlab{a}})Lin, Zhang, Ding, Wu, and Zhao]{lin2025model}
Haohong Lin, Yunzhi Zhang, Wenhao Ding, Jiajun Wu, and Ding Zhao.
\newblock Model-based policy adaptation for closed-loop end-to-end autonomous driving.
\newblock In \emph{Workshop on Foundation Models Meet Embodied Agents at CVPR 2025}, 2025{\natexlab{a}}.

\bibitem[Lin et~al.(2025{\natexlab{b}})Lin, Yang, Zhang, Zheng, Feng, Wang, Wang, Chen, Wang, Zhang, et~al.]{lin2025futurex}
Hongbin Lin, Yiming Yang, Yifan Zhang, Chaoda Zheng, Jie Feng, Sheng Wang, Zhennan Wang, Shijia Chen, Boyang Wang, Yu~Zhang, et~al.
\newblock Futurex: Enhance end-to-end autonomous driving via latent chain-of-thought world model.
\newblock \emph{arXiv preprint arXiv:2512.11226}, 2025{\natexlab{b}}.

\bibitem[Liu et~al.(2024{\natexlab{a}})Liu, Chen, Qiao, Lv, and Li]{liu2024reasoning}
Haochen Liu, Li~Chen, Yu~Qiao, Chen Lv, and Hongyang Li.
\newblock Reasoning multi-agent behavioral topology for interactive autonomous driving.
\newblock \emph{Advances in Neural Information Processing Systems}, 37:\penalty0 92605--92637, 2024{\natexlab{a}}.

\bibitem[Liu et~al.(2025)Liu, Huang, Huang, Yang, Mo, and Lv]{liu2025hybrid}
Haochen Liu, Zhiyu Huang, Wenhui Huang, Haohan Yang, Xiaoyu Mo, and Chen Lv.
\newblock Hybrid-prediction integrated planning for autonomous driving.
\newblock \emph{IEEE Transactions on Pattern Analysis and Machine Intelligence}, 47\penalty0 (4):\penalty0 2597--2614, 2025.

\bibitem[Liu et~al.(2026)Liu, Li, Yang, Chen, Wang, Guo, Tian, Li, Li, and Lv]{liu2026r2se}
Haochen Liu, Tianyu Li, Haohan Yang, Li~Chen, Caojun Wang, Ke~Guo, Haochen Tian, Hongchen Li, Hongyang Li, and Chen Lv.
\newblock Reinforced refinement with self-aware expansion for end-to-end autonomous driving.
\newblock \emph{IEEE Transactions on Pattern Analysis and Machine Intelligence}, 2026.

\bibitem[Liu et~al.(2024{\natexlab{b}})Liu, Nam, Campbell, St{"a}rk, Xu, Jaakkola, and G{'o}mez-Bombarelli]{liu2024think}
Sulin Liu, Juno Nam, Andrew Campbell, Hannes St{"a}rk, Yilun Xu, Tommi Jaakkola, and Rafael G{'o}mez-Bombarelli.
\newblock Think while you generate: Discrete diffusion with planned denoising.
\newblock \emph{arXiv preprint arXiv:2410.06264}, 2024{\natexlab{b}}.

\bibitem[Locatello et~al.(2020)Locatello, Weissenborn, Unterthiner, Mahendran, Heigold, Uszkoreit, Dosovitskiy, and Kipf]{locatello2020object}
Francesco Locatello, Dirk Weissenborn, Thomas Unterthiner, Aravindh Mahendran, Georg Heigold, Jakob Uszkoreit, Alexey Dosovitskiy, and Thomas Kipf.
\newblock Object-centric learning with slot attention.
\newblock \emph{Advances in neural information processing systems}, 33:\penalty0 11525--11538, 2020.

\bibitem[Lu et~al.(2024)Lu, Huang, Yang, Zhang, and Zhang]{lu2024wovogen}
Jiachen Lu, Ze~Huang, Zeyu Yang, Jiahui Zhang, and Li~Zhang.
\newblock Wovogen: World volume-aware diffusion for controllable multi-camera driving scene generation.
\newblock In \emph{European conference on computer vision}, pages 329--345. Springer, 2024.

\bibitem[Luxembourg et~al.(2025)Luxembourg, Permuter, and Nachmani]{luxembourg2025plan}
Omer Luxembourg, Haim Permuter, and Eliya Nachmani.
\newblock Plan for speed: Dilated scheduling for masked diffusion language models.
\newblock \emph{arXiv preprint arXiv:2506.19037}, 2025.

\bibitem[Ma et~al.(2026)Ma, Zheng, Wang, Jiang, Cui, Liang, and Yang]{ma2026dit4dit}
Teli Ma, Jia Zheng, Zifan Wang, Chunli Jiang, Andy Cui, Junwei Liang, and Shuo Yang.
\newblock Dit4dit: Jointly modeling video dynamics and actions for generalizable robot control.
\newblock \emph{arXiv preprint arXiv:2603.10448}, 2026.

\bibitem[Ma et~al.(2025)Ma, Cao, Ding, Zhang, Wang, Ivanovic, Jiang, Pavone, and Xiao]{ma2025dvlm}
Yingzi Ma, Yulong Cao, Wenhao Ding, Shuibai Zhang, Yan Wang, Boris Ivanovic, Ming Jiang, Marco Pavone, and Chaowei Xiao.
\newblock dvlm-ad: Enhance diffusion vision-language-model for driving via controllable reasoning.
\newblock \emph{arXiv preprint arXiv:2512.04459}, 2025.

\bibitem[Park et~al.(2025)Park, Lai, Hayakawa, Takida, and Mitsufuji]{park2025jump}
Yong-Hyun Park, Chieh-Hsin Lai, Satoshi Hayakawa, Yuhta Takida, and Yuki Mitsufuji.
\newblock Jump your steps: Optimizing sampling schedule of discrete diffusion models.
\newblock In \emph{International Conference on Learning Representations}, volume 2025, pages 96272--96300, 2025.

\bibitem[Peng et~al.(2025)Peng, Bezemek, Patel, Rector-Brooks, Yao, Tong, and Chatterjee]{peng2025path}
Fred~Zhangzhi Peng, Zachary Bezemek, Sawan Patel, Jarrid Rector-Brooks, Sherwood Yao, Alexander Tong, and Pranam Chatterjee.
\newblock Path planning for masked diffusion model sampling.
\newblock \emph{arXiv preprint arXiv:2502.03540}, 2025.

\bibitem[Schiff et~al.(2026)Schiff, Belhasin, Uziel, Wang, Arriola, Turok, Elad, and Kuleshov]{schiff2026learn}
Yair Schiff, Omer Belhasin, Roy Uziel, Guanghan Wang, Marianne Arriola, Gilad Turok, Michael Elad, and Volodymyr Kuleshov.
\newblock Learn from your mistakes: Self-correcting masked diffusion models.
\newblock \emph{arXiv preprint arXiv:2602.11590}, 2026.

\bibitem[Shang et~al.(2026)Shang, Chen, Wang, Li, and ZHANG]{shang2026drivedpo}
Shuyao Shang, Yuntao Chen, Yuqi Wang, Yingyan Li, and ZHAO-XIANG ZHANG.
\newblock Drivedpo: Policy learning via safety dpo for end-to-end autonomous driving.
\newblock \emph{Advances in Neural Information Processing Systems}, 38:\penalty0 81565--81585, 2026.

\bibitem[Shao et~al.(2024)Shao, Wang, Zhu, Xu, Song, Bi, Zhang, Zhang, Li, Wu, et~al.]{shao2024deepseekmath}
Zhihong Shao, Peiyi Wang, Qihao Zhu, Runxin Xu, Junxiao Song, Xiao Bi, Haowei Zhang, Mingchuan Zhang, YK~Li, Yang Wu, et~al.
\newblock Deepseekmath: Pushing the limits of mathematical reasoning in open language models.
\newblock \emph{arXiv preprint arXiv:2402.03300}, 2024.

\bibitem[Shi et~al.(2026)Shi, Xu, Shi, Sheng, Zhang, and Jiang]{shi2026drivewam}
Chen Shi, Jinrui Xu, Shaoshuai Shi, Kehua Sheng, Bo~Zhang, and Li~Jiang.
\newblock Drivewam: Video generative priors enable scalable world-action modeling for autonomous driving.
\newblock \emph{arXiv preprint arXiv:2605.28544}, 2026.

\bibitem[Sim{\'e}oni et~al.(2025)Sim{\'e}oni, Vo, Seitzer, Baldassarre, Oquab, Jose, Khalidov, Szafraniec, Yi, Ramamonjisoa, et~al.]{simeoni2025dinov3}
Oriane Sim{\'e}oni, Huy~V Vo, Maximilian Seitzer, Federico Baldassarre, Maxime Oquab, Cijo Jose, Vasil Khalidov, Marc Szafraniec, Seungeun Yi, Micha{\"e}l Ramamonjisoa, et~al.
\newblock Dinov3.
\newblock \emph{arXiv preprint arXiv:2508.10104}, 2025.

\bibitem[Song et~al.(2025)Song, Liu, Pan, Liao, Guo, Yang, Zhang, Xu, Jia, and Luo]{song2025diver}
Ziying Song, Lin Liu, Hongyu Pan, Bencheng Liao, Mingzhe Guo, Lei Yang, Yongchang Zhang, Shaoqing Xu, Caiyan Jia, and Yadan Luo.
\newblock Diver: Reinforced diffusion breaks imitation bottlenecks in end-to-end autonomous driving.
\newblock \emph{arXiv preprint arXiv:2507.04049}, 2025.

\bibitem[Sun et~al.(2026)Sun, Lin, Chen, Pei, Li, Shi, and Zheng]{sun2026sparsedrivev2}
Wenchao Sun, Xuewu Lin, Keyu Chen, Zixiang Pei, Xiang Li, Yining Shi, and Sifa Zheng.
\newblock Sparsedrivev2: Scoring is all you need for end-to-end autonomous driving.
\newblock \emph{arXiv preprint arXiv:2603.29163}, 2026.

\bibitem[Tian et~al.(2025)Tian, Li, Liu, Yang, Qiu, Li, Wang, Gao, Zhang, Wang, et~al.]{tian2025simscale}
Haochen Tian, Tianyu Li, Haochen Liu, Jiazhi Yang, Yihang Qiu, Guang Li, Junli Wang, Yinfeng Gao, Zhang Zhang, Liang Wang, et~al.
\newblock Simscale: Learning to drive via real-world simulation at scale.
\newblock \emph{arXiv preprint arXiv:2511.23369}, 2025.

\bibitem[Tong et~al.(2023)Tong, Sima, Wang, Chen, Wu, Deng, Gu, Lu, Luo, Lin, et~al.]{tong2023scene}
Wenwen Tong, Chonghao Sima, Tai Wang, Li~Chen, Silei Wu, Hanming Deng, Yi~Gu, Lewei Lu, Ping Luo, Dahua Lin, et~al.
\newblock Scene as occupancy.
\newblock In \emph{Proceedings of the IEEE/CVF International Conference on Computer Vision}, pages 8406--8415, 2023.

\bibitem[van~den Oord et~al.(2017)van~den Oord, Vinyals, and Kavukcuoglu]{vanDenOord2017vqvae}
Aaron van~den Oord, Oriol Vinyals, and Koray Kavukcuoglu.
\newblock Neural discrete representation learning.
\newblock In \emph{Advances in Neural Information Processing Systems}, 2017.

\bibitem[Wang et~al.(2026{\natexlab{a}})Wang, Wang, Cui, Li, Lu, Wang, Wang, Tang, Zhang, and Zhan]{wang2026reflectdrive}
Huimin Wang, Yue Wang, Bihao Cui, Pengxiang Li, Ben Lu, Mingqian Wang, Tong Wang, Chuan Tang, Teng Zhang, and Kun Zhan.
\newblock Reflectdrive-2: Reinforcement-learning-aligned self-editing for discrete diffusion driving.
\newblock \emph{arXiv preprint arXiv:2605.04647}, 2026{\natexlab{a}}.

\bibitem[Wang et~al.(2026{\natexlab{b}})Wang, Zheng, Chen, Li, Zhang, Xing, Zhang, Li, Qian, Yang, et~al.]{wang2026latentwam}
Linbo Wang, Yupeng Zheng, Qiang Chen, Shiwei Li, Yichen Zhang, Zebin Xing, Qichao Zhang, Xiang Li, Deheng Qian, Pengxuan Yang, et~al.
\newblock Latent-wam: Latent world action modeling for end-to-end autonomous driving.
\newblock \emph{arXiv preprint arXiv:2603.24581}, 2026{\natexlab{b}}.

\bibitem[Wang et~al.(2026{\natexlab{c}})Wang, Yang, Bai, Zhang, Liu, Zheng, Long, Lu, and Lu]{wang2026drive}
Linhan Wang, Zichong Yang, Chen Bai, Guoxiang Zhang, Xiaotong Liu, Xiaoyin Zheng, Xiao-Xiao Long, Chang-Tien Lu, and Cheng Lu.
\newblock Drive-jepa: Video jepa meets multimodal trajectory distillation for end-to-end driving.
\newblock \emph{arXiv preprint arXiv:2601.22032}, 2026{\natexlab{c}}.

\bibitem[Wang et~al.(2022)Wang, Wang, Zhang, Liu, and Sun]{wang2022social}
Wenshuo Wang, Letian Wang, Chengyuan Zhang, Changliu Liu, and Lijun Sun.
\newblock Social interactions for autonomous driving: A review and perspectives.
\newblock \emph{Foundations and Trends{\textregistered} in Robotics}, 10\penalty0 (3-4):\penalty0 198--377, 2022.

\bibitem[Wang et~al.(2024{\natexlab{a}})Wang, Zhu, Huang, Chen, Zhu, and Lu]{wang2024drivedreamer}
Xiaofeng Wang, Zheng Zhu, Guan Huang, Xinze Chen, Jiagang Zhu, and Jiwen Lu.
\newblock Drivedreamer: Towards real-world-drive world models for autonomous driving.
\newblock In \emph{European conference on computer vision}, pages 55--72. Springer, 2024{\natexlab{a}}.

\bibitem[Wang et~al.(2024{\natexlab{b}})Wang, He, Fan, Li, Chen, and Zhang]{wang2024driving}
Yuqi Wang, Jiawei He, Lue Fan, Hongxin Li, Yuntao Chen, and Zhaoxiang Zhang.
\newblock Driving into the future: Multiview visual forecasting and planning with world model for autonomous driving.
\newblock In \emph{Proceedings of the IEEE/CVF Conference on Computer Vision and Pattern Recognition}, pages 14749--14759, 2024{\natexlab{b}}.

\bibitem[Wei et~al.(2024)Wei, Yuan, Li, Hu, Gan, and Ding]{wei2024occllama}
Julong Wei, Shanshuai Yuan, Pengfei Li, Qingda Hu, Zhongxue Gan, and Wenchao Ding.
\newblock Occllama: An occupancy-language-action generative world model for autonomous driving.
\newblock \emph{arXiv preprint arXiv:2409.03272}, 2024.

\bibitem[Weng et~al.(2024)Weng, Ivanovic, Wang, Wang, and Pavone]{weng2024drive}
Xinshuo Weng, Boris Ivanovic, Yan Wang, Yue Wang, and Marco Pavone.
\newblock Para-drive: Parallelized architecture for real-time autonomous driving.
\newblock In \emph{Proceedings of the IEEE/CVF Conference on Computer Vision and Pattern Recognition}, pages 15449--15458, 2024.

\bibitem[Xia et~al.(2026)Xia, Li, Zhou, Yao, Xiong, Sun, Wang, Ma, Chen, Ye, et~al.]{xia2026drivelaw}
Tianze Xia, Yongkang Li, Lijun Zhou, Jingfeng Yao, Kaixin Xiong, Haiyang Sun, Bing Wang, Kun Ma, Guang Chen, Hangjun Ye, et~al.
\newblock Drivelaw: Unifying planning and video generation in a latent driving world.
\newblock In \emph{Proceedings of the IEEE/CVF Conference on Computer Vision and Pattern Recognition}, pages 39701--39712, 2026.

\bibitem[Xing et~al.(2025)Xing, Zhang, Hu, Jiang, He, Zhang, Long, and Yin]{xing2025goalflow}
Zebin Xing, Xingyu Zhang, Yang Hu, Bo~Jiang, Tong He, Qian Zhang, Xiaoxiao Long, and Wei Yin.
\newblock Goalflow: Goal-driven flow matching for multimodal trajectories generation in end-to-end autonomous driving.
\newblock In \emph{Proceedings of the Computer Vision and Pattern Recognition Conference}, pages 1602--1611, 2025.

\bibitem[Xu et~al.(2025)Xu, Cui, Cai, Zhu, Shang, Luan, Xu, Zhang, Li, Cai, et~al.]{xu2025wam}
Yifang Xu, Jiahao Cui, Feipeng Cai, Zhihao Zhu, Hanlin Shang, Shan Luan, Mingwang Xu, Neng Zhang, Yaoyi Li, Jia Cai, et~al.
\newblock Wam-flow: Parallel coarse-to-fine motion planning via discrete flow matching for autonomous driving.
\newblock \emph{arXiv preprint arXiv:2512.06112}, 2025.

\bibitem[Xu et~al.(2024)Xu, Zhang, Xie, Zhao, Guo, Wong, Li, and Zhao]{xu2024drivegpt4}
Zhenhua Xu, Yujia Zhang, Enze Xie, Zhen Zhao, Yong Guo, Kwan-Yee~K Wong, Zhenguo Li, and Hengshuang Zhao.
\newblock Drivegpt4: Interpretable end-to-end autonomous driving via large language model.
\newblock \emph{IEEE Robotics and Automation Letters}, 2024.

\bibitem[Yan et~al.(2026)Yan, Tang, Gui, Li, Zheng, Huang, Kong, Han, Zhou, Zhang, et~al.]{yan2026ad}
Tianyi Yan, Tao Tang, Xingtai Gui, Yongkang Li, Jiasen Zheng, Weiyao Huang, Lingdong Kong, Wencheng Han, Xia Zhou, Xueyang Zhang, et~al.
\newblock Ad-r1: Closed-loop reinforcement learning for end-to-end autonomous driving with impartial world models.
\newblock In \emph{Proceedings of the IEEE/CVF Conference on Computer Vision and Pattern Recognition}, pages 1085--1095, 2026.

\bibitem[Yang et~al.(2025)Yang, Li, Yang, Zhang, Hui, Zheng, Yu, Gao, Huang, Lv, et~al.]{yang2025qwen3}
An~Yang, Anfeng Li, Baosong Yang, Beichen Zhang, Binyuan Hui, Bo~Zheng, Bowen Yu, Chang Gao, Chengen Huang, Chenxu Lv, et~al.
\newblock Qwen3 technical report.
\newblock \emph{arXiv preprint arXiv:2505.09388}, 2025.

\bibitem[Yang et~al.(2024{\natexlab{a}})Yang, Gao, Qiu, Chen, Li, Dai, Chitta, Wu, Zeng, Luo, et~al.]{yang2024genad}
Jiazhi Yang, Shenyuan Gao, Yihang Qiu, Li~Chen, Tianyu Li, Bo~Dai, Kashyap Chitta, Penghao Wu, Jia Zeng, Ping Luo, et~al.
\newblock Genad: Generalized predictive model for autonomous driving.
\newblock \emph{arXiv preprint arXiv:2403.09630}, 2024{\natexlab{a}}.

\bibitem[Yang et~al.(2024{\natexlab{b}})Yang, Gao, Qiu, Chen, Li, Dai, Chitta, Wu, Zeng, Luo, et~al.]{yang2024generalized}
Jiazhi Yang, Shenyuan Gao, Yihang Qiu, Li~Chen, Tianyu Li, Bo~Dai, Kashyap Chitta, Penghao Wu, Jia Zeng, Ping Luo, et~al.
\newblock Generalized predictive model for autonomous driving.
\newblock In \emph{Proceedings of the IEEE/CVF Conference on Computer Vision and Pattern Recognition}, pages 14662--14672, 2024{\natexlab{b}}.

\bibitem[Yang et~al.(2026{\natexlab{a}})Yang, Lin, Li, Zhang, Lin, Wu, Su, Zhao, Zhang, Chen, et~al.]{yang2026rise}
Jiazhi Yang, Kunyang Lin, Jinwei Li, Wencong Zhang, Tianwei Lin, Longyan Wu, Zhizhong Su, Hao Zhao, Ya-Qin Zhang, Li~Chen, et~al.
\newblock Rise: Self-improving robot policy with compositional world model.
\newblock \emph{arXiv preprint arXiv:2602.11075}, 2026{\natexlab{a}}.

\bibitem[Yang et~al.(2026{\natexlab{b}})Yang, Zheng, Qian, Xing, Zhang, Wang, Zhang, Guo, Xia, Chen, et~al.]{yang2026dreamerad}
Pengxuan Yang, Yupeng Zheng, Deheng Qian, Zebin Xing, Qichao Zhang, Linbo Wang, Yichen Zhang, Shaoyu Guo, Zhongpu Xia, Qiang Chen, et~al.
\newblock Dreamerad: Efficient reinforcement learning via latent world model for autonomous driving.
\newblock \emph{arXiv preprint arXiv:2603.24587}, 2026{\natexlab{b}}.

\bibitem[Yao et~al.(2026{\natexlab{a}})Yao, Li, Lan, Wang, Sun, Alvarez, and Wu]{yao2026drivesuprim}
Wenhao Yao, Zhenxin Li, Shiyi Lan, Zi~Wang, Xinglong Sun, Jose~M Alvarez, and Zuxuan Wu.
\newblock Drivesuprim: Towards precise trajectory selection for end-to-end planning.
\newblock In \emph{Proceedings of the AAAI Conference on Artificial Intelligence}, volume~40, pages 11910--11918, 2026{\natexlab{a}}.

\bibitem[Yao et~al.(2026{\natexlab{b}})Yao, Zhu, Jiang, Guo, and Zhao]{yao2026unifieddrivingtokensrepresentation}
Ziyang Yao, Zeyu Zhu, YunCheng Jiang, Zibin Guo, and Huijing Zhao.
\newblock Unified driving tokens: Representation- and geometry-guided discrete tokenizer for driving world models and planning.
\newblock \emph{arXiv preprint arXiv:2606.01935}, 2026{\natexlab{b}}.

\bibitem[Ye et~al.(2025)Ye, Zhang, and Zhao]{ye2025dap}
Bowen Ye, Bin Zhang, and Hang Zhao.
\newblock Dap: A discrete-token autoregressive planner for autonomous driving.
\newblock \emph{arXiv preprint arXiv:2511.13306}, 2025.

\bibitem[Ye et~al.(2026)Ye, Ge, Zheng, Gao, Yu, Kurian, Indupuru, Tan, Zhu, Xiang, et~al.]{ye2026world}
Seonghyeon Ye, Yunhao Ge, Kaiyuan Zheng, Shenyuan Gao, Sihyun Yu, George Kurian, Suneel Indupuru, You~Liang Tan, Chuning Zhu, Jiannan Xiang, et~al.
\newblock World action models are zero-shot policies.
\newblock \emph{arXiv preprint arXiv:2602.15922}, 2026.

\bibitem[Yuan et~al.(2026)Yuan, Dong, Liu, and Zhao]{yuan2026fast}
Tianyuan Yuan, Zibin Dong, Yicheng Liu, and Hang Zhao.
\newblock Fast-wam: Do world action models need test-time future imagination?
\newblock \emph{arXiv preprint arXiv:2603.16666}, 2026.

\bibitem[Zeng and Dong(2026)]{zeng2026latent}
Rongxiang Zeng and Yongqi Dong.
\newblock Latent world models for automated driving: A unified taxonomy, evaluation framework, and open challenges.
\newblock \emph{arXiv preprint arXiv:2603.09086}, 2026.

\bibitem[Zhang et~al.(2025)Zhang, Tang, Hu, Pan, Guo, Liu, Huang, Yuan, Zhang, Long, et~al.]{zhang2025epona}
Kaiwen Zhang, Zhenyu Tang, Xiaotao Hu, Xingang Pan, Xiaoyang Guo, Yuan Liu, Jingwei Huang, Li~Yuan, Qian Zhang, Xiao-Xiao Long, et~al.
\newblock Epona: Autoregressive diffusion world model for autonomous driving.
\newblock In \emph{Proceedings of the IEEE/CVF International Conference on Computer Vision}, pages 27220--27230, 2025.

\bibitem[Zhang et~al.(2026)Zhang, Wang, Gao, Wu, Cao, Han, Ivanovic, Liu, Pavone, Han, et~al.]{zhang2026fastddrive}
Kewei Zhang, Jin Wang, Sensen Gao, Chengyue Wu, Yulong Cao, Songyang Han, Boris Ivanovic, Langechuan Liu, Marco Pavone, Song Han, et~al.
\newblock Fast-ddrive: Efficient block-diffusion vlm for autonomous driving.
\newblock \emph{arXiv preprint arXiv:2605.23163}, 2026.

\bibitem[Zhang and Syed(2025)]{zhang2025cosine}
Leo Zhang and Saifuddin Syed.
\newblock The cosine schedule is fisher-rao-optimal for masked discrete diffusion models.
\newblock \emph{arXiv preprint arXiv:2508.04884}, 2025.

\bibitem[Zhao et~al.(2026)Zhao, Gupta, Zheng, and Grover]{zhao2026d1}
Siyan Zhao, Devaansh Gupta, Qinqing Zheng, and Aditya Grover.
\newblock d1: Scaling reasoning in diffusion large language models via reinforcement learning.
\newblock \emph{Advances in Neural Information Processing Systems}, 38:\penalty0 56729--56762, 2026.

\bibitem[Zhao et~al.(2024)Zhao, Shi, Chen, Druckmann, Mackey, and Linderman]{zhao2024informed}
Yixiu Zhao, Jiaxin Shi, Feng Chen, Shaul Druckmann, Lester Mackey, and Scott Linderman.
\newblock Informed correctors for discrete diffusion models.
\newblock \emph{arXiv preprint arXiv:2407.21243}, 2024.

\bibitem[Zhou et~al.(2025)Zhou, Cai, Zhao, Zhang, Huang, Zhou, and Ma]{zhou2025autovla}
Zewei Zhou, Tianhui Cai, Seth~Z Zhao, Yun Zhang, Zhiyu Huang, Bolei Zhou, and Jiaqi Ma.
\newblock Autovla: A vision-language-action model for end-to-end autonomous driving with adaptive reasoning and reinforcement fine-tuning.
\newblock \emph{arXiv preprint arXiv:2506.13757}, 2025.

\bibitem[Zhu et~al.(2025)Zhu, Yu, Feng, Burchfiel, Shah, and Gupta]{zhu2025unified}
Chuning Zhu, Raymond Yu, Siyuan Feng, Benjamin Burchfiel, Paarth Shah, and Abhishek Gupta.
\newblock Unified world models: Coupling video and action diffusion for pretraining on large robotic datasets.
\newblock \emph{arXiv preprint arXiv:2504.02792}, 2025.

\bibitem[Zou et~al.(2025)Zou, Chen, Liao, Zheng, Song, Zhang, Zhang, Liu, and Wang]{zou2025diffusiondrivev2}
Jialv Zou, Shaoyu Chen, Bencheng Liao, Zhiyu Zheng, Yuehao Song, Lefei Zhang, Qian Zhang, Wenyu Liu, and Xinggang Wang.
\newblock Diffusiondrivev2: Reinforcement learning-constrained truncated diffusion modeling in end-to-end autonomous driving.
\newblock \emph{arXiv preprint arXiv:2512.07745}, 2025.

\end{thebibliography}
\clearpage

\clearpage
\clearpage
\onecolumn

{\renewcommand{\thefootnote}{\fnsymbol{footnote}}

\section{Contributions and Acknowledgments}

\noindent
\begin{minipage}[t]{0.48\textwidth}
\subsubsection*{Core Contributors}
\begin{itemize}
    \item Ziyang Yao\footnotemark[1]
    \item Haochen Liu\footnotemark[1]\footnotemark[2]
    \item Yuncheng Jiang\footnotemark[1]\footnotemark[2]\footnotemark[4]
    \item Zeyu Zhu\footnotemark[3]
    \item Zibin Guo
    \item Jingwei Zhao
    \item Guang Chen
    \item Hangjun Ye
\end{itemize}
\end{minipage}
\hfill
\begin{minipage}[t]{0.48\textwidth}
\subsubsection*{Contributors}
\begin{itemize}
    \item Jingru Wang
    \item Tianle Liu
    \item Jianwei Cui
    \item Kuiyuan Yang
    \item Hongwei Xie
\end{itemize}
\end{minipage}

\footnotetext[1]{Equal contribution.}
\footnotetext[2]{Project lead.}
\footnotetext[3]{Work done at Xiaomi EV.}
\footnotetext[4]{Corresponding author.}
}
\clearpage

\section{Appendix}

\subsection{Related Work}

\subsubsection{World Policy Modeling}
World models (WMs) have emerged as a central paradigm for physical intelligence, aiming to model how the external world evolves under agent actions~\cite{guan2024world}. In autonomous driving, early world-model-inspired approaches were often coupled with planning through intermediate representations, such as occupancy forecasting~\cite{wei2024occllama,tong2023scene,liu2025hybrid} or joint motion prediction~\cite{liu2024reasoning,chen2024ppad,huang2023gameformer,hagedorn2024integration}, where future dynamics were learned primarily to support downstream decision making. With the emergence of large-scale vision foundation models~\cite{kong20253d}, recent research has increasingly shifted toward future visual supervision~\cite{gao2024vista,lu2024wovogen,yang2024generalized,wang2024drivedreamer}, leveraging image and video generation as rich supervisory signals for representation learning~\cite{wang2026drive}, planning-oriented reasoning~\cite{yang2026dreamerad}, and even explicit or implicit reward modeling~\cite{wang2024driving,li2025end}. In parallel, another line of research focuses on simulation and data generation engines. These approaches construct interactive or long-tail scenarios through structured world modeling~\cite{yang2026rise}, such as behavior simulation~\cite{lin2025model}, 3D scene reconstruction~\cite{tian2025simscale,gao2026rad}, or generative video models~\cite{wang2024drivedreamer,yang2024genad}. By synthesizing diverse future outcomes and counterfactual interactions, they provide scalable environments for policy evaluation, closed-loop training, and safety validation. Despite their success, most existing approaches still treat world modeling and planning as loosely coupled components, where the world model primarily serves as an auxiliary predictor or simulator rather than being jointly optimized with decision making under a unified causal formulation.

This limitation has motivated the recent development of world-action modeling (WAM)~\cite{zhu2025unified}, which jointly formulates future observation and policy generation as a unified learning problem~\cite{cen2025worldvla,li2025drivevla}. Some works model future worlds and agent actions as a single generative process~\cite{ma2026dit4dit,ye2026world,xia2026drivelaw}, while others emphasize unified pretraining~\cite{yuan2026fast} of shared world-policy representations~\cite{bi2026motus}, latent action modeling~\cite{gao2026dreamdojo}, and value-aware world models~\cite{kim2026cosmos} to reduce the redundancy of explicit observation generation and strengthen the coupling between environment dynamics and control. However, existing formulations are predominantly built upon continuous latent spaces, which often suffer from representation ambiguity~\cite{shi2026drivewam,zhang2025epona}. These limitations have recently inspired the adoption of parallel generative paradigms based on discrete token spaces, including mask modeling and discrete diffusion~\cite{wang2026reflectdrive,xu2025wam,zhang2026fastddrive}. While such methods improve generation efficiency and enable iterative refinement, most of them still formulate decision making as a direct observation-to-action mapping without explicitly learning a unified generative prior over both future worlds and policies. In contrast, \XFM{} formulates observations and actions within a shared discrete representation space. Unified generative pretraining establishes a common prior over world evolution and policy generation, while the joint discrete diffusion formulation provides a unified framework for world modeling and decision making.

\subsubsection{Discrete Diffusion Scheduling}
Sampling schedules play a central role in discrete and masked diffusion models because they determine not only the number of reverse steps, but also which tokens are decoded jointly at each step. In discrete diffusion models, early work such as D3PM formalizes discrete corruption kernels, including absorbing-state processes, which provide the basis for mask-based reverse generation \cite{austin2021structured}. Subsequent work studies scheduling from several complementary perspectives.

One line of work focuses on time-grid or noise-level scheduling. JYS optimizes the sampling time grid by minimizing a path-space KL upper bound, showing that non-uniform schedules can reduce discretization error in discrete diffusion sampling \cite{park2025jump}. Related information-geometric analysis further argues that the commonly used cosine schedule can be interpreted as Fisher--Rao optimal for masked discrete diffusion under specific geometric assumptions \cite{zhang2025cosine}.

A second line studies how many tokens should be decoded per iteration. \cite{li2025convergence} derive an information-theoretic convergence bound for diffusion language models and show that balanced block schedules yield an O(1/T)-type reduction of KL error under general assumptions . \cite{chen2025optimal} further characterize the optimal inference schedule as the best step-function approximation to an information curve, making explicit that the optimal block sizes depend on the dependency structure of the data distribution . \cite{lavenant2025error} also analyze the error induced by factorized approximations and derive asymptotically optimal schedules from an information-profile viewpoint .

A third line uses confidence or entropy to adapt the unmasking budget. EB-Sampler decomposes sampling error into denoiser model error and joint-dependence error, and uses entropy-bounded unmasking to accelerate masked diffusion sampling while controlling the risk of decoding too many uncertain tokens simultaneously \cite{benhamu2025accelerated}. \cite{cai2026confidence} provide a provable analysis of confidence-based decoding and show that entropy-sum stopping rules can achieve KL-accurate sampling with an expected number of steps depending on the intrinsic entropy of the data distribution .

A fourth line studies which positions should be decoded together. DUS proposes a dilated unmasking schedule that selects separated positions in early iterations to reduce the entropy gap among jointly decoded tokens, followed by finer local decoding in later iterations \cite{luxembourg2025plan}. This is particularly suitable when dependencies are local or fast-mixing, but it may be less reliable for tasks dominated by global consistency constraints.

Finally, several works extend scheduling with learned planners or correction mechanisms. DDPD and P2 decouple position planning from token denoising, allowing the model to learn nontrivial unmasking paths \cite{liu2024think,peng2025path}. Informed correctors and self-correcting masked diffusion further introduce correction or remasking mechanisms to revise earlier decoding errors \cite{zhao2024informed,schiff2026learn}. These methods are practically useful, but their theoretical guarantees are generally weaker than the information-theoretic schedule analyses above. Recent theory also shows that although masked diffusion can be efficient for token-level quality, sequence-level correctness may still require a number of steps scaling with sequence length in worst-case settings \cite{feng2025theoretical,kim2025train}. Therefore, scheduling should be understood as a mechanism for reducing discretization error and parallel-dependence error, rather than as a universal solution to all long-range dependency constraints.

\subsubsection{Policy Post-training}
\label{related:post-training}
While large-scale pretraining has significantly improved the generalization capability of E2E autonomous driving systems~\cite{chen2024end}, it is fundamentally optimized under an open-loop objective and therefore cannot fully address distribution shifts induced by closed-loop interactions~\cite{wang2022social}. In particular, behavior cloning learns to match demonstrated actions but does not directly optimize the sequential objectives used in closed-loop evaluation, resulting in a persistent objective gap~\cite{KarkusIglEtAl2025}. To mitigate this discrepancy, studies have explored offline reinforcement learning and post-training strategies for policy alignment~\cite{Li_2025_ICCV,song2025diver}. Early approaches construct positive and negative trajectory pairs and apply preference optimization, such as DPO-style ranking~\cite{shang2026drivedpo} or inverse reinforcement learning~\cite{jiang2025irlvla}, to encourage preferred behaviors. However, these methods remain optimizing relative preferences rather than long-horizon driving objective. More recently, policy-gradient-based approaches such as GRPO~\cite{shao2024deepseekmath} have been introduced to directly optimize trajectory-level rewards and closed-loop metrics~\cite{li2025recogdrive,liu2026r2se,li2025end}. Nevertheless, existing methods typically operate on complete trajectories and therefore only improve trajectory generation conditioned on a fixed decision, without explicitly optimizing the high-level decision-making process itself. A parallel research direction leverages 3DGS-based simulators or learned world models to enable online RL through interaction~\cite{gao2026rad,tian2025simscale,yan2026ad,lin2025model}. Although these methods provide a mechanism for closed-loop optimization, their performance is inherently limited by accumulated simulation and model errors, which introduce additional bias into policy learning. Recent approaches attempt to alleviate this issue by performing reinforcement learning directly in latent transition spaces~\cite{yang2026dreamerad,wang2026reflectdrive,chen2026last,wang2026latentwam}, avoiding explicit simulator rollouts while still improving long-horizon reasoning capabilities. In contrast, our framework jointly optimizes both decision selection and trajectory planning through a unified GRPO objective. By performing policy optimization over a structured decision-planning hierarchy, our method not only improves planning quality under closed-loop metrics but also explicitly refines the decision space itself, enabling more effective optimization of long-horizon driving behavior.

\subsection{Additional Implementation Details}
\label{implementation_details}

\subsubsection{Token Design}
\label{token_design}
\paragraph{Vision token configuration} We employ the codebook size of $K_V=16384$ for quantizer, and $H_V,W_V=16$ for image patch. The rest of the pipeline are aligned with our previous setup.

\paragraph{Decision token configuration}
We parameterize high-level driving decisions with a discrete decision vocabulary constructed from lateral path candidates and longitudinal speed profiles. Specifically, let $\mathcal{P}=\{p_i\}_{i=1}^{N_{\mathrm{lat}}}$ denote the set of lateral path primitives and $\mathcal{S}=\{s_j\}_{j=1}^{N_{\mathrm{lon}}} $denote the set of longitudinal speed profiles. Their Cartesian product defines a decision token space:
$\mathcal{D}=\mathcal{P}\times\mathcal{S}, |\mathcal{D}|=N_{\mathrm{lat}}N_{\mathrm{lon}}=400,$
where each decision token $d_{i,j}\in\mathcal{D}$ specifies a coarse behavior prior combining path topology and speed evolution. During pretraining, we assign supervision by evaluating all candidate decisions with a winner-take-all criterion based on the EPDMS sub-scores. The selected decision is
$d^\star=\arg\max_{d\in\mathcal{D}} R_{\mathrm{EPDMS}}(d),$
where $R_{\mathrm{EPDMS}}$ aggregates safety, progress, comfort, and rule-compliance sub-scores. The model is then trained to predict $d^\star$ from the current observation and context using a cross-entropy objective:
$\mathcal{L}_{\mathrm{dec}}=-\log p_\theta(d^\star \mid \mathbf{o}, \mathbf{c})+\mathcal{L}_{\text{score}}.$

For SFT and post-training, we further restrict the decision space to the top-$D$ candidates ranking, forming a compact decision set $\mathcal{D}_\mathrm{top} \subset \mathcal{D}$. Subsequent supervised fine-tuning and RL optimization are performed within $\mathcal{D}_{\mathrm{top}}$, which preserves high-quality behavioral diversity while avoiding exploration over suboptimal decisions.

\paragraph{Vocabulary configuration for action and auxiliary position supervision}
For the acceleration-token vocabulary, we use a two-dimensional grid over ego-centric longitudinal and lateral accelerations. Both acceleration components are clipped to the valid range $[-4,4]\,\mathrm{m/s^2}$. The longitudinal and lateral acceleration dimensions are uniformly partitioned into $N_x=N_y=60$ bins, forming a grid-structured acceleration vocabulary with $60\times 60=3600$ prototypes. Each vocabulary entry corresponds to a 2D acceleration prototype
\[
\mathbf{v}_{ij}=(c^x_i,c^y_j),
\qquad
i,j\in\{1,\ldots,60\},
\]
where $c^x_i$ and $c^y_j$ denote the uniformly spaced bin centers along the longitudinal and lateral acceleration dimensions. Continuous accelerations are represented by soft labels over the four neighboring prototypes through bilinear interpolation, as described in Sec.~\ref{sec:action_tokenization}.

In addition to acceleration-token classification, we introduce an auxiliary position classification task to supervise the trajectory obtained after integrating the predicted accelerations. To avoid the prohibitive vocabulary size induced by a Cartesian-product 2D position grid, we factorize position supervision into two independent one-dimensional classification tasks over ego-centric longitudinal and lateral positions. The longitudinal position range is $[-2,65]\,\mathrm{m}$, and the lateral position range is $[-25,25]\,\mathrm{m}$. Both dimensions use a grid-cell resolution of $0.02\,\mathrm{m}$, resulting in
\[
N^p_x = \frac{65-(-2)}{0.02}=3350,
\qquad
N^p_y = \frac{25-(-25)}{0.02}=2500.
\]
The auxiliary position heads therefore predict two categorical distributions, one over $3350$ longitudinal bins and the other over $2500$ lateral bins, rather than a single $3350\times2500$ joint vocabulary. This factorized position supervision provides fine-grained trajectory-level spatial constraints while keeping the classification space computationally tractable.

\paragraph{Auxiliary factorized position classification loss}
Let $\mathbf{p}_h=(x_h,y_h)$ denote the ground-truth future ego position at horizon step $h$. The auxiliary position classification loss is defined as the sum of the longitudinal and lateral cross-entropy losses:
\[
\mathcal{L}_{\mathrm{pos}}
=
\sum_{h=1}^{H}
\left[
\mathrm{CE}\left(p^{x,\ast}_h, q^x_{\theta,h}\right)
+
\mathrm{CE}\left(p^{y,\ast}_h, q^y_{\theta,h}\right)
\right],
\]
where $p^{x,\ast}_h$ and $p^{y,\ast}_h$ denote the target distributions over the discretized $x$ and $y$ position bins, and $q^x_{\theta,h}$ and $q^y_{\theta,h}$ are the corresponding predicted distributions. This factorized formulation avoids the cost of a joint 2D position vocabulary while still providing direct supervision on the integrated trajectory.

\subsubsection{Detailed Token-Editing Objective}
\label{app:token_editing_objective}

We provide the detailed mathematical formulation of the token-editing objective used in unified pretraining. At time step $t$, the scene context is denoted as $\mathbf{C}_t$, which contains historical visual tokens, ego-state tokens, and navigation tokens. The model may also condition on a high-level decision token sequence $\mathbf{D}_t\subset\mathcal{D}_\mathrm{top}$.

Let $\mathbf{X}=\{x_j\}_{j=1}^{N}$ denote a clean target token sequence. Depending on the task, $\mathbf{X}$ can be a future visual token sequence $\mathbf{V}_{t+1:t+H}$ or a future action token sequence $\mathbf{A}_{t+1:t+H}$. The corrupted version of the target sequence is denoted as $\tilde{\mathbf{X}}=\{\tilde{x}_j\}_{j=1}^{N}$. The effective model input is the concatenation of the scene context, optional decision tokens, and corrupted target tokens:
$[\mathbf{C}_t, \mathbf{D}_t, \tilde{\mathbf{X}}].$
Unlike masked diffusion methods that introduce a special mask token, our formulation corrupts tokens within the original discrete vocabulary. Let $\mathcal{M}^X_\gamma$ denote the corrupted token positions under corruption ratio $\gamma\in[0,1]$. The vocabulary associated with $\mathbf{X}$ is
\begin{equation}
\mathcal{B}_X =
\begin{cases}
\mathcal{V}, & \mathbf{X}=\mathbf{V}_{t+1:t+H},\\
\mathcal{A}, & \mathbf{X}=\mathbf{A}_{t+1:t+H},
\end{cases}
\end{equation}
where $\mathcal{V}$ and $\mathcal{A}$ denote the visual and action vocabularies, respectively. The corrupted sequence is constructed as
\begin{equation}
\tilde{x}_j =
\begin{cases}
\eta_j, & j\in\mathcal{M}^X_\gamma,\\
x_j, & j\notin\mathcal{M}^X_\gamma,
\end{cases}
\quad
\eta_j\sim\mathrm{Unif}(\mathcal{B}_X),
\end{equation}
where $\mathrm{Unif}(\mathcal{B}_X)$ denotes the uniform distribution over valid tokens in $\mathcal{B}_X$.

The token-editing loss is applied to all editable positions, including both corrupted and clean tokens:
\begin{equation}
\mathcal{L}_{\mathrm{edit}}(\mathbf{X})
=
-\frac{1}{N}
\sum_{j=1}^{N}
\log p_\theta(x_j \mid \mathbf{C}_t, \mathbf{D}_t, \tilde{\mathbf{X}}, \gamma).
\end{equation}
For corrupted positions, this loss trains the model to recover the clean target tokens. For clean positions, it encourages an identity mapping, teaching the model to preserve tokens that are already correct and providing an implicit stopping signal for token editing.

The corruption pattern differs across modalities. For visual prediction, corrupted image-token positions are sampled uniformly from all subsets with cardinality $\lfloor\gamma N_v\rfloor$:
\begin{equation}
\mathcal{M}^{v}_\gamma \sim
\mathrm{Unif}\left(
\left\{
\mathcal{M}\subseteq\{1,\ldots,N_v\}:
|\mathcal{M}|=\lfloor\gamma N_v\rfloor
\right\}
\right).
\end{equation}
For action prediction, corruption follows a causal suffix pattern:
\begin{equation}
h_\gamma=\lfloor(1-\gamma)H\rfloor,
\quad
\mathcal{M}^{a}_\gamma=\{h_\gamma+1,\ldots,H\}.
\end{equation}
Thus, the first $h_\gamma$ action tokens remain clean, while the remaining suffix is corrupted. This design respects the temporal dependency of acceleration-based action tokens.

The final training objective combines token classification losses and continuous motion regression losses:
\begin{equation}
\mathcal{L}
=
\lambda_v \mathcal{L}_{v}^{\mathrm{cls}}
+
\lambda_a \mathcal{L}_{a}^{\mathrm{cls}}
+
\lambda_{\mathrm{acc}} \mathcal{L}_{\mathrm{acc}}
+
\lambda_{\mathrm{traj}} \mathcal{L}_{\mathrm{traj}}
+
\lambda_s \mathcal{L}_{s}^{\mathrm{cls}}
+
\lambda_\mathrm{dec} \mathcal{L}_\mathrm{dec}.
\end{equation}
Here, $\mathcal{L}_{v}^{\mathrm{cls}}$ and $\mathcal{L}_{a}^{\mathrm{cls}}$ are cross-entropy losses over the visual and action vocabularies. $\mathcal{L}_{\mathrm{acc}}$ is the acceleration-level regression loss, and $\mathcal{L}_{\mathrm{traj}}$ is the trajectory-level regression loss obtained after integrating predicted accelerations into future ego trajectories. $\mathcal{L}_{s}^{\mathrm{cls}}$ denotes the classification loss for auxiliary special tokens. Different tasks activate different subsets of these losses according to their prediction targets.

\subsubsection{Model Structure Design}
\label{model_design}
Our model is built upon a decoder-only Transformer~\cite{yang2025qwen3} with hidden dimension $d=2048$, 18 Transformer layers, 16 attention heads, and 8 key-value heads  of grouped-query attention. Rotary positional embeddings follow the Qwen3-MRoPE scheme with $\theta=10^6$ and multi-axis section splits $[24,20,20]$. The total parameter count is approximately 1B. Parameters for LoRA finetuning is about 30M.

\subsubsection{Benchmark Details}
\label{benchmark_detail}
\paragraph{Evaluation metrics}
We evaluate planning performance using the Predictive Driver Model Score (PDMS) and its extended version EPDMS used in NAVSIM. PDMS was introduced in NAVSIM v1~\cite{dauner2024navsim} as a simulation-based open-loop metric, where the predicted 4-second trajectory is unrolled in a BEV simulator and scored by combining multiplicative safety constraints with weighted planning-quality subscores. It is defined as
\begin{equation}
\mathrm{PDMS}
=
\left(
\prod_{m \in \{\mathrm{NC}, \mathrm{DAC}\}}
m(\mathrm{agent})
\right)
\cdot
\frac{
\sum_{m \in \{\mathrm{TTC}, \mathrm{EP}, \mathrm{C}\}}
w_m m(\mathrm{agent})
}{
\sum_{m \in \{\mathrm{TTC}, \mathrm{EP}, \mathrm{C}\}} w_m
},
\end{equation}
where NC denotes no at-fault collision, DAC denotes drivable-area compliance, TTC denotes time-to-collision, EP denotes ego progress, and C denotes comfort. NAVSIM v2~\cite{cao2025pseudo} further adopts the Extended Predictive Driver Model Score (EPDMS), which adds driving-direction compliance (DDC), traffic-light compliance (TLC) for multiplicative scoring, and adding lane keeping (LK), history comfort (HC), and extended comfort (EC) in weighted scores. EPDMS further adds filter with
\begin{equation}
\mathrm{filter}_m(\mathrm{agent}, \mathrm{human})
=
\begin{cases}
1.0, & \text{if } m(\mathrm{human}) = 0, \\
m(\mathrm{agent}), & \text{otherwise}.
\end{cases}
\end{equation}
This filtering avoids penalizing the planner when the same violation is also present in the human demonstration, which can occur due to annotation noise or contextually valid maneuvers. The weighted EPDMS terms use $w_{\mathrm{EP}}=5$, $w_{\mathrm{TTC}}=5$, $w_{\mathrm{LK}}=2$, $w_{\mathrm{HC}}=2$, and $w_{\mathrm{EC}}=2$.

\paragraph{Baselines}
We compare Discrete-WAM with a diverse set of state-of-the-art autonomous driving systems, including modular end-to-end plannersr~\cite{chitta2022transfuser,chen2024vadv2,hu2023planning}, generative planners~\cite{li2024hydra,sun2026sparsedrivev2,guo2025ipad,liao2025diffusiondrive,zou2025diffusiondrivev2,li2025recogdrive}, world-model-based methods~\cite{zhang2025epona,li2025drivevla}, world-action policies~\cite{xu2025wam,yang2026dreamerad,wang2026latentwam}, and vision-language-action (VLA) approaches~\cite{li2025drivevla,jiang2025irlvla}. 

\subsection{Analytical Results}
\label{app:analytical_results}

This section provides the theoretical analysis used in Sec.~\ref{sec:unified_pretraining}. 
We use a generic notation where $C$ denotes the context, $Z$ denotes a latent skeleton, and $\mathbf{Y}_U=\{Y_i:i\in U\}$ denotes a group of future tokens. 
For policy modeling, $Z$ corresponds to the decision skeleton $\mathbf{D}_t$, and $\mathbf{Y}_U$ corresponds to a subset of future action tokens.

\subsubsection{One-step KL Decomposition}
\label{app:analytical_one_step}

For a token group $\mathbf{Y}_U$, the conditional total correlation is defined as
\begin{equation}
\mathrm{TC}(\mathbf{Y}_U\mid C)
=
D_{\mathrm{KL}}
\left(
q(\mathbf{Y}_U\mid C)
\Vert
\prod_{i\in U}q(Y_i\mid C)
\right).
\end{equation}
Equivalently,
\begin{equation}
\mathrm{TC}(\mathbf{Y}_U\mid C)
=
\sum_{i\in U}H(Y_i\mid C)-H(\mathbf{Y}_U\mid C).
\end{equation}

Assume a parallel token predictor factorizes the token group as
\begin{equation}
p_\theta(\mathbf{Y}_U\mid C)=\prod_{i\in U}p_\theta(Y_i\mid C).
\end{equation}
Then the one-step KL error decomposes as
\begin{equation}
\begin{aligned}
&D_{\mathrm{KL}}
\left(
q(\mathbf{Y}_U\mid C)
\Vert
\prod_{i\in U}p_\theta(Y_i\mid C)
\right) \\
&=
\mathrm{TC}(\mathbf{Y}_U\mid C)
+
\sum_{i\in U}
D_{\mathrm{KL}}
\left(
q(Y_i\mid C)
\Vert
p_\theta(Y_i\mid C)
\right).
\end{aligned}
\end{equation}

\begin{proof}
By expanding the KL divergence,
\begin{equation}
\begin{aligned}
&D_{\mathrm{KL}}
\left(
q(\mathbf{Y}_U\mid C)
\Vert
\prod_{i\in U}p_\theta(Y_i\mid C)
\right) \\
&=
\mathbb{E}_{q}
\left[
\log q(\mathbf{Y}_U\mid C)
-
\sum_{i\in U}\log p_\theta(Y_i\mid C)
\right].
\end{aligned}
\end{equation}
Adding and subtracting $\sum_{i\in U}\log q(Y_i\mid C)$ gives
\begin{equation}
\begin{aligned}
&\mathbb{E}_{q}
\left[
\log \frac{q(\mathbf{Y}_U\mid C)}
{\prod_{i\in U}q(Y_i\mid C)}
\right]
+
\sum_{i\in U}
\mathbb{E}_{q}
\left[
\log \frac{q(Y_i\mid C)}
{p_\theta(Y_i\mid C)}
\right],
\end{aligned}
\end{equation}
which is exactly the sum of conditional total correlation and token-level model errors.
\end{proof}

\subsubsection{Latent Skeleton Decomposition}
\label{app:analytical_latent_decomp}

We define the redundancy gain of a latent skeleton $Z$ as
\begin{equation}
R_Z(U\mid C)
=
\sum_{i\in U} I(Y_i;Z\mid C)
-
I(\mathbf{Y}_U;Z\mid C).
\end{equation}
Then the skeleton-conditioned residual total correlation satisfies
\begin{equation}
\mathbb{E}_{Z}
\mathrm{TC}(\mathbf{Y}_U\mid C,Z)
=
\mathrm{TC}(\mathbf{Y}_U\mid C)
-
R_Z(U\mid C).
\end{equation}

\begin{proof}
Using the entropy form of total correlation,
\begin{equation}
\mathbb{E}_{Z}\mathrm{TC}(\mathbf{Y}_U\mid C,Z)
=
\sum_{i\in U}H(Y_i\mid C,Z)-H(\mathbf{Y}_U\mid C,Z).
\end{equation}
Since
\begin{equation}
H(Y_i\mid C,Z)=H(Y_i\mid C)-I(Y_i;Z\mid C)
\end{equation}
and
\begin{equation}
H(\mathbf{Y}_U\mid C,Z)=H(\mathbf{Y}_U\mid C)-I(\mathbf{Y}_U;Z\mid C),
\end{equation}
we obtain
\begin{equation}
\begin{aligned}
\mathbb{E}_{Z}\mathrm{TC}(\mathbf{Y}_U\mid C,Z)
&=
\sum_{i\in U}H(Y_i\mid C)-H(\mathbf{Y}_U\mid C) \\
&\quad -
\left[
\sum_{i\in U}I(Y_i;Z\mid C)-I(\mathbf{Y}_U;Z\mid C)
\right] \\
&=
\mathrm{TC}(\mathbf{Y}_U\mid C)-R_Z(U\mid C).
\end{aligned}
\end{equation}
\end{proof}

This identity shows that introducing a latent skeleton does not automatically reduce token dependence. 
The residual total correlation decreases only when $R_Z(U\mid C)>0$. 
This is expected when $Z$ is an upstream common-cause skeleton of the fine tokens, but may fail when $Z$ is a downstream collider or a synergistic summary of the token group.

\subsubsection{Positive Redundancy Gain under Residual Mixing}
\label{app:analytical_positive_gain}

We now state sufficient conditions under which a latent skeleton yields positive redundancy gain.

\textbf{Assumption 1: Upstream common-cause skeleton.}
Given context $C$, the latent skeleton $Z$ is an upstream low-frequency variable that conditions the generation of the fine token group $\mathbf{Y}_U$:
\begin{equation}
C \rightarrow Z \rightarrow \mathbf{Y}_U.
\end{equation}
This excludes downstream trajectory summaries, evaluation labels, or selection variables that are computed after observing the full token group.

\textbf{Assumption 2: Pre-skeleton dependence lower bound.}
There exists $\kappa(U)>0$ such that
\begin{equation}
\mathrm{TC}(\mathbf{Y}_U\mid C)\ge \kappa(U).
\end{equation}
This means that the token group contains non-trivial group-level dependence before conditioning on the skeleton.

\textbf{Assumption 3: Residual mixing after skeleton conditioning.}
After conditioning on $(C,Z)$, the residual dependence between fine tokens decays with their distance:
\begin{equation}
I(Y_i;Y_j\mid C,Z)
\le
\beta
\exp\left(-\frac{d(i,j)}{\ell_z}\right),
\end{equation}
where $\beta$ measures residual local coupling strength, $d(i,j)$ is the distance between tokens $i$ and $j$, and $\ell_z$ is the residual correlation length. 
Under this assumption, the skeleton-conditioned residual total correlation is upper bounded by
\begin{equation}
\mathbb{E}_{Z}\mathrm{TC}(\mathbf{Y}_U\mid C,Z)
\le
\varepsilon_z(U),
\end{equation}
where
\begin{equation}
\varepsilon_z(U)
=
\mathbb{E}_{Z}
\left[
\sum_{\{i,j\}\subset U}
\beta
\exp\left(-\frac{d(i,j)}{\ell_z}\right)
\right].
\end{equation}

Combining the latent skeleton decomposition with the above assumptions gives
\begin{equation}
\begin{aligned}
R_Z(U\mid C)
&=
\mathrm{TC}(\mathbf{Y}_U\mid C)
-
\mathbb{E}_{Z}\mathrm{TC}(\mathbf{Y}_U\mid C,Z) \\
&\ge
\kappa(U)-\varepsilon_z(U).
\end{aligned}
\end{equation}
Therefore, if
\begin{equation}
\kappa(U)>\varepsilon_z(U),
\end{equation}
then
\begin{equation}
R_Z(U\mid C)>0.
\end{equation}
Equivalently, under the residual mixing bound, a verifiable sufficient condition is
\begin{equation}
\kappa(U)>
\mathbb{E}_{Z}
\left[
\sum_{\{i,j\}\subset U}
\beta
\exp\left(-\frac{d(i,j)}{\ell_z}\right)
\right]
\Longrightarrow
R_Z(U\mid C)>0.
\end{equation}

This condition has a direct interpretation for trajectory planning. 
The skeleton $Z$ is useful when it explains the strong low-frequency dependence shared by multiple future action tokens, leaving only weak local residual dependence after conditioning. 
If $Z$ is hard to predict, acts as a downstream collider, or fails to reduce residual dependence, the positive-gain condition may not hold.

\subsubsection{KL Upper Bound with Model Error and Re-edit Schedule}
\label{app:analytical_kl_upper_bound}

We finally incorporate token-level model error, initial proposal mismatch, skeleton prediction error, and re-edit scheduling. 
Let $\pi$ denote an iterative token-editing schedule with $R$ rounds. 
At round $r$, let $A_r$ be the active edit set and let $S_r$ denote the editing state before updating the active tokens. 
The token-level model error is defined as
\begin{equation}
\delta_{i,r}(S_r,Z)
=
D_{\mathrm{KL}}
\left(
q(Y^{(r)}_i\mid S_r)
\Vert
p_\theta(Y^{(r)}_i\mid S_r)
\right).
\end{equation}

We also define the initial proposal mismatch as
\begin{equation}
\delta_{\mathrm{init}}
=
\mathbb{E}_{q(C,Z)}
D_{\mathrm{KL}}
\left(
q_0(\mathbf{Y}^{(0)}\mid C,Z)
\Vert
p_0(\mathbf{Y}^{(0)}\mid C,Z)
\right),
\end{equation}
where $q_0$ is the reference initial proposal distribution and $p_0$ is the initial proposal distribution used by the model. 
If both processes start from the same proposal distribution, then $\delta_{\mathrm{init}}=0$.

For a given skeleton $Z$, applying the one-step KL decomposition at each edit round gives
\begin{equation}
\begin{aligned}
&D_{\mathrm{KL}}
\left(
q(\mathbf{Y}^{(R)}\mid C,Z)
\Vert
p_{\theta,\pi}(\mathbf{Y}^{(R)}\mid C,Z)
\right) \\
&\le
\delta_{\mathrm{init}}
+
\sum_{r=1}^{R}
\mathbb{E}
\left[
\sum_{i\in A_r}\delta_{i,r}(S_r,Z)
+
\mathrm{TC}(\mathbf{Y}_{A_r}^{(r)}\mid S_r,Z)
\right].
\end{aligned}
\end{equation}
Using the residual mixing bound on each active edit set,
\begin{equation}
\mathrm{TC}(\mathbf{Y}_{A_r}^{(r)}\mid S_r,Z)
\le
\sum_{\{i,j\}\subset A_r}
\beta_r
\exp\left(-\frac{d_r(i,j)}{\ell_r}\right),
\end{equation}
we obtain
\begin{equation}
\begin{aligned}
&D_{\mathrm{KL}}
\left(
q(\mathbf{Y}^{(R)}\mid C,Z)
\Vert
p_{\theta,\pi}(\mathbf{Y}^{(R)}\mid C,Z)
\right) \\
&\le
\delta_{\mathrm{init}}
+
\sum_{r=1}^{R}
\mathbb{E}
\left[
\sum_{i\in A_r}\delta_{i,r}(S_r,Z)
+
\sum_{\{i,j\}\subset A_r}
\beta_r
\exp\left(-\frac{d_r(i,j)}{\ell_r}\right)
\right].
\end{aligned}
\end{equation}

In our hierarchical policy model, the latent skeleton $Z$ is instantiated as the decision token $\mathbf{D}_t$ and is predicted from the scene context. 
Therefore, we introduce a skeleton prediction error to measure the mismatch between the target skeleton distribution and the model-predicted skeleton distribution:
\begin{equation}
\delta_Z
=
D_{\mathrm{KL}}
\left(
q(Z\mid C)
\Vert
p_\psi(Z\mid C)
\right).
\end{equation}
The full KL upper bound is then
\begin{equation}
D_{\mathrm{KL}}
\left(
q(\mathbf{Y}\mid C)
\Vert
p_{\psi,\theta,\pi}(\mathbf{Y}\mid C)
\right)
\le
\delta_Z+\Lambda_z(\pi),
\end{equation}
where
\begin{equation}
\Lambda_z(\pi)
=
\delta_{\mathrm{init}}
+
\sum_{r=1}^{R}
\mathbb{E}
\left[
\sum_{i\in A_r}\delta_{i,r}(S_r,Z)
+
\sum_{\{i,j\}\subset A_r}
\beta_r
\exp\left(-\frac{d_r(i,j)}{\ell_r}\right)
\right].
\end{equation}

Equivalently, define
\begin{equation}
\mathcal{B}_{\mathrm{model}}(\pi)
=
\sum_{r=1}^{R}
\mathbb{E}
\left[
\sum_{i\in A_r}\delta_{i,r}(S_r,Z)
\right],
\end{equation}
and
\begin{equation}
\mathcal{U}_{\mathrm{dep}}(\pi)
=
\sum_{r=1}^{R}
\mathbb{E}
\left[
\sum_{\{i,j\}\subset A_r}
\beta_r
\exp\left(-\frac{d_r(i,j)}{\ell_r}\right)
\right].
\end{equation}
Then
\begin{equation}
D_{\mathrm{KL}}
\left(
q(\mathbf{Y}\mid C)
\Vert
p_{\psi,\theta,\pi}(\mathbf{Y}\mid C)
\right)
\le
\delta_Z
+
\delta_{\mathrm{init}}
+
\mathcal{B}_{\mathrm{model}}(\pi)
+
\mathcal{U}_{\mathrm{dep}}(\pi).
\end{equation}

This bound shows that the latent skeleton improves the overall generation risk only when its prediction cost is small and the reduction in residual dependence outweighs the additional skeleton prediction error.

\paragraph{Implication for policy modeling}
In our policy modeling task, the decision token $\mathbf{D}_t$ plays the role of the latent skeleton $Z$. 
A valid decision token should represent upstream low-frequency planning structure, such as maneuver intent, coarse reference motion, target lane, or speed trend, rather than a downstream trajectory-quality label computed from the final action sequence. 
Under this interpretation, $\mathbf{D}_t$ explains global multi-modal driving choices before fine action-token editing, while the remaining action tokens mainly model local residual corrections. 
Therefore, the hierarchical factorization
\[
p_{\psi,\theta}(\mathbf{A}_{t+1:t+H},\mathbf{D}_t\mid \mathbf{C}_t)
=
p_\psi(\mathbf{D}_t\mid \mathbf{C}_t)
p_\theta(\mathbf{A}_{t+1:t+H}\mid \mathbf{C}_t,\mathbf{D}_t)
\]
is theoretically justified when $\mathbf{D}_t$ reduces residual action-token dependence enough to offset its own prediction error.

\subsubsection{Exact Reconstruction under Soft-label Interpolation}
\label{app:soft_label_exact_reconstruction}

We show that the proposed soft-label action representation removes the deterministic hard-quantization error within each acceleration grid cell. 
Let the acceleration vocabulary be defined by the Cartesian product of the bin centers along the longitudinal and lateral acceleration dimensions. 
Denote the bin centers of $a_x$ by $\{c^x_i\}_{i=1}^{N_x}$ and the bin centers of $a_y$ by $\{c^y_j\}_{j=1}^{N_y}$. 
Consider a continuous acceleration vector $\mathbf{a}=(a_x,a_y)$ that lies in the grid cell spanned by four neighboring prototypes:
\begin{equation}
\begin{aligned}
\mathbf{v}_{ij} &= (c^x_i,c^y_j), 
&
\mathbf{v}_{i+1,j} &= (c^x_{i+1},c^y_j), \\
\mathbf{v}_{i,j+1} &= (c^x_i,c^y_{j+1}), 
&
\mathbf{v}_{i+1,j+1} &= (c^x_{i+1},c^y_{j+1}).
\end{aligned}
\label{eq:soft_label_neighboring_prototypes}
\end{equation}
That is,
\begin{equation}
a_x \in [c^x_i,c^x_{i+1}], 
\qquad
a_y \in [c^y_j,c^y_{j+1}].
\label{eq:soft_label_cell_condition}
\end{equation}
We define the interpolation coefficients
\begin{equation}
\lambda_x=\frac{a_x-c^x_i}{c^x_{i+1}-c^x_i}, 
\qquad
\lambda_y=\frac{a_y-c^y_j}{c^y_{j+1}-c^y_j}.
\label{eq:soft_label_interpolation_coefficients}
\end{equation}
The soft target distribution $p^\ast(\cdot \mid \mathbf{a})$ is nonzero only on the four neighboring prototypes, with weights
\begin{equation}
\begin{aligned}
p^\ast_{ij} &= (1-\lambda_x)(1-\lambda_y), \\
p^\ast_{i+1,j} &= \lambda_x(1-\lambda_y), \\
p^\ast_{i,j+1} &= (1-\lambda_x)\lambda_y, \\
p^\ast_{i+1,j+1} &= \lambda_x\lambda_y.
\end{aligned}
\label{eq:soft_label_2d_weights}
\end{equation}
These weights are non-negative and sum to one. 
The acceleration reconstructed from the soft target is
\begin{equation}
\bar{\mathbf{a}}
=
\sum_{m\in\{i,i+1\}}
\sum_{n\in\{j,j+1\}}
p^\ast_{mn}\mathbf{v}_{mn}.
\label{eq:soft_label_reconstructed_acceleration}
\end{equation}
For the longitudinal component, we have
\begin{equation}
\bar{a}_x
=
p^\ast_{ij}c^x_i
+
p^\ast_{i+1,j}c^x_{i+1}
+
p^\ast_{i,j+1}c^x_i
+
p^\ast_{i+1,j+1}c^x_{i+1}.
\label{eq:soft_label_reconstructed_ax_expand}
\end{equation}
Substituting the interpolation weights gives
\begin{equation}
\bar{a}_x
=
(1-\lambda_x)c^x_i+\lambda_x c^x_{i+1}
=
a_x.
\label{eq:soft_label_reconstructed_ax}
\end{equation}
Similarly, for the lateral component,
\begin{equation}
\bar{a}_y
=
(1-\lambda_y)c^y_j+\lambda_y c^y_{j+1}
=
a_y.
\label{eq:soft_label_reconstructed_ay}
\end{equation}
Therefore,
\begin{equation}
\bar{\mathbf{a}}=\mathbf{a}.
\label{eq:soft_label_exact_reconstruction}
\end{equation}

During training, the action head predicts a distribution $q_\theta(\cdot \mid C_t)$ over the acceleration vocabulary and is optimized using soft-label cross-entropy:
\begin{equation}
\mathcal{L}^{\mathrm{cls}}_a
=
-
\sum_{k=1}^{N_xN_y}
p^\ast_k(\mathbf{a})
\log q_{\theta,k}(C_t).
\label{eq:soft_label_ce_loss}
\end{equation}
Since
\begin{equation}
\mathcal{L}^{\mathrm{cls}}_a
=
H(p^\ast)
+
\mathrm{KL}(p^\ast\Vert q_\theta),
\label{eq:soft_label_ce_kl_decomposition}
\end{equation}
the ideal optimum under sufficient model capacity and optimization satisfies
\begin{equation}
q_\theta(\cdot \mid C_t)=p^\ast(\cdot \mid \mathbf{a}).
\label{eq:soft_label_ideal_prediction}
\end{equation}
At inference time, if the continuous acceleration is decoded by the vocabulary expectation
\begin{equation}
\hat{\mathbf{a}}
=
\sum_{k=1}^{N_xN_y}
q_{\theta,k}(C_t)\mathbf{v}_k,
\label{eq:soft_label_expectation_decoding}
\end{equation}
then under the ideal prediction condition $q_\theta=p^\ast$, we obtain
\begin{equation}
\hat{\mathbf{a}}
=
\sum_{k=1}^{N_xN_y}
p^\ast_k(\mathbf{a})\mathbf{v}_k
=
\mathbf{a}.
\label{eq:soft_label_inference_exact_reconstruction}
\end{equation}
Thus, under exact recovery of the soft target distribution, the proposed soft-label interpolation yields zero deterministic quantization error within the acceleration grid cell.

\subsubsection{Consistency Bound for Continuous Motion Supervision}
\label{app:continuous_motion_consistency_bound}

\paragraph{Acceleration error under distribution prediction mismatch}
We next characterize the remaining reconstruction error when the predicted action distribution does not exactly match the soft target distribution. 
Let $p^\ast(\cdot \mid \mathbf{a})$ denote the interpolation target distribution for the continuous acceleration $\mathbf{a}$, and let $q_\theta(\cdot \mid C_t)$ denote the predicted action distribution. 
The decoded continuous acceleration is
\begin{equation}
\hat{\mathbf{a}}
=
\sum_{k=1}^{N_xN_y}
q_{\theta,k}(C_t)\mathbf{v}_k.
\label{eq:predicted_acceleration_expectation}
\end{equation}
Since the soft target distribution exactly reconstructs $\mathbf{a}$ under grid interpolation,
\begin{equation}
\mathbf{a}
=
\sum_{k=1}^{N_xN_y}
p^\ast_k(\mathbf{a})\mathbf{v}_k.
\label{eq:target_acceleration_expectation}
\end{equation}
Therefore, the acceleration reconstruction error can be written as
\begin{equation}
\hat{\mathbf{a}}-\mathbf{a}
=
\sum_{k=1}^{N_xN_y}
\left(
q_{\theta,k}(C_t)-p^\ast_k(\mathbf{a})
\right)
\mathbf{v}_k.
\label{eq:acceleration_error_decomposition}
\end{equation}
Assume the acceleration vocabulary is bounded by
\begin{equation}
\|\mathbf{v}_k\|_2 \le V_{\max},
\qquad
\forall k\in\{1,\ldots,N_xN_y\}.
\label{eq:acceleration_vocab_bound}
\end{equation}
Then, by the triangle inequality,
\begin{equation}
\|\hat{\mathbf{a}}-\mathbf{a}\|_2
\le
\sum_{k=1}^{N_xN_y}
\left|
q_{\theta,k}(C_t)-p^\ast_k(\mathbf{a})
\right|
\|\mathbf{v}_k\|_2.
\label{eq:acceleration_error_triangle}
\end{equation}
Using the boundedness of the vocabulary prototypes, we obtain
\begin{equation}
\|\hat{\mathbf{a}}-\mathbf{a}\|_2
\le
V_{\max}
\|q_\theta(\cdot \mid C_t)-p^\ast(\cdot \mid \mathbf{a})\|_1.
\label{eq:acceleration_error_l1_bound}
\end{equation}
By Pinsker's inequality,
\begin{equation}
\|q_\theta(\cdot \mid C_t)-p^\ast(\cdot \mid \mathbf{a})\|_1
\le
\sqrt{
2\mathrm{KL}
\left(
p^\ast(\cdot \mid \mathbf{a})
\Vert
q_\theta(\cdot \mid C_t)
\right)
}.
\label{eq:pinsker_soft_label}
\end{equation}
Therefore,
\begin{equation}
\|\hat{\mathbf{a}}-\mathbf{a}\|_2
\le
V_{\max}
\sqrt{
2\mathrm{KL}
\left(
p^\ast(\cdot \mid \mathbf{a})
\Vert
q_\theta(\cdot \mid C_t)
\right)
}.
\label{eq:acceleration_error_kl_bound}
\end{equation}
Since the soft-label cross-entropy satisfies
\begin{equation}
\mathcal{L}^{\mathrm{cls}}_a
=
H(p^\ast)
+
\mathrm{KL}
\left(
p^\ast(\cdot \mid \mathbf{a})
\Vert
q_\theta(\cdot \mid C_t)
\right),
\label{eq:soft_label_ce_kl_bound}
\end{equation}
we further have
\begin{equation}
\|\hat{\mathbf{a}}-\mathbf{a}\|_2
\le
V_{\max}
\sqrt{
2\left(
\mathcal{L}^{\mathrm{cls}}_a
-
H(p^\ast)
\right)
}.
\label{eq:acceleration_error_ce_bound}
\end{equation}
This shows that, after replacing hard one-hot quantization with soft-label interpolation, the remaining acceleration reconstruction error is controlled by the distribution prediction error rather than by deterministic nearest-bin quantization error.

\paragraph{Mode-aware decoding for continuous motion supervision}
The soft-label interpolation analysis above concerns the construction of the action-token target distribution and the deterministic error introduced by hard quantization. 
In contrast, the mode-aware decoding strategy is used for a different purpose: it converts the predicted categorical action distribution into a continuous acceleration for the auxiliary continuous motion losses. 
Its goal is not to preserve the full-distribution expectation, but to avoid averaging incompatible action modes.

Let $q_\theta(\cdot \mid C_t)$ denote the predicted categorical distribution over the acceleration vocabulary $\{\mathbf{v}_k\}_{k=1}^{K}$. 
Full-distribution expectation decoding gives
\begin{equation}
\bar{\mathbf{a}}
=
\sum_{k=1}^{K}
q_{\theta,k}(C_t)\mathbf{v}_k.
\label{eq:full_distribution_expectation_decoding}
\end{equation}
When $q_\theta$ is multi-modal, $\bar{\mathbf{a}}$ can lie between several plausible modes and may not correspond to any physically meaningful driving behavior. 
To avoid this decoding-induced mode averaging, we fit $q_\theta$ with a Gaussian mixture over the acceleration vocabulary, select one mode support $\mathcal{S}_{m^\star}$, and form the normalized local distribution
\begin{equation}
\tilde{q}_{\theta,k}(C_t)
=
\frac{
q_{\theta,k}(C_t)\mathbf{1}[k\in \mathcal{S}_{m^\star}]
}{
\sum_{\ell\in \mathcal{S}_{m^\star}}
q_{\theta,\ell}(C_t)
}.
\label{eq:mode_aware_local_distribution}
\end{equation}
The mode-aware decoded acceleration is then
\begin{equation}
\hat{\mathbf{a}}_{\mathrm{mode}}
=
\sum_{k=1}^{K}
\tilde{q}_{\theta,k}(C_t)\mathbf{v}_k.
\label{eq:mode_aware_decoded_acceleration}
\end{equation}

This operation intentionally replaces the full predicted distribution $q_\theta$ with a mode-conditioned local distribution $\tilde{q}_\theta$. 
Therefore, it introduces a mode-selection approximation while avoiding the mode-averaging effect of full-distribution expectation decoding. 
Let $p^\ast(\cdot \mid \mathbf{a})$ denote the soft interpolation target of the ground-truth acceleration. 
The reconstruction error of mode-aware decoding can be decomposed as
\begin{equation}
\begin{aligned}
\hat{\mathbf{a}}_{\mathrm{mode}}-\mathbf{a}
&=
\sum_{k=1}^{K}
\left(
\tilde{q}_{\theta,k}(C_t)-p^\ast_k(\mathbf{a})
\right)\mathbf{v}_k \\
&=
\sum_{k=1}^{K}
\left(
q_{\theta,k}(C_t)-p^\ast_k(\mathbf{a})
\right)\mathbf{v}_k
+
\sum_{k=1}^{K}
\left(
\tilde{q}_{\theta,k}(C_t)-q_{\theta,k}(C_t)
\right)\mathbf{v}_k .
\end{aligned}
\label{eq:mode_aware_error_decomposition}
\end{equation}
The first term corresponds to the distribution prediction error with respect to the soft-label target, while the second term corresponds to the additional mode-selection error introduced by replacing $q_\theta$ with $\tilde{q}_\theta$. 
Assuming $\|\mathbf{v}_k\|_2\le V_{\max}$ for all vocabulary prototypes, we obtain
\begin{equation}
\begin{aligned}
\|\hat{\mathbf{a}}_{\mathrm{mode}}-\mathbf{a}\|_2
&\le
V_{\max}
\|q_\theta(\cdot\mid C_t)-p^\ast(\cdot\mid \mathbf{a})\|_1 \\
&\quad+
V_{\max}
\|\tilde{q}_\theta(\cdot\mid C_t)-q_\theta(\cdot\mid C_t)\|_1 .
\end{aligned}
\label{eq:mode_aware_error_bound}
\end{equation}
Using Pinsker's inequality for the first term gives
\begin{equation}
\begin{aligned}
\|\hat{\mathbf{a}}_{\mathrm{mode}}-\mathbf{a}\|_2
&\le
V_{\max}
\sqrt{
2\mathrm{KL}
\left(
p^\ast(\cdot\mid \mathbf{a})
\Vert
q_\theta(\cdot\mid C_t)
\right)
} \\
&\quad+
V_{\max}
\|\tilde{q}_\theta(\cdot\mid C_t)-q_\theta(\cdot\mid C_t)\|_1 .
\end{aligned}
\label{eq:mode_aware_kl_bound}
\end{equation}
This bound separates the error caused by imperfect distribution prediction from the error introduced by mode selection. 
In practice, the mode-aware decoding is used only for continuous acceleration and trajectory regression losses. 
It prevents the regression loss from forcing a multi-modal categorical prediction into its global mean, thereby reducing mode collapse while preserving the discrete action distribution for token-level policy prediction.

\begin{figure}[!h]
    \centering
    \includegraphics[width=\linewidth]{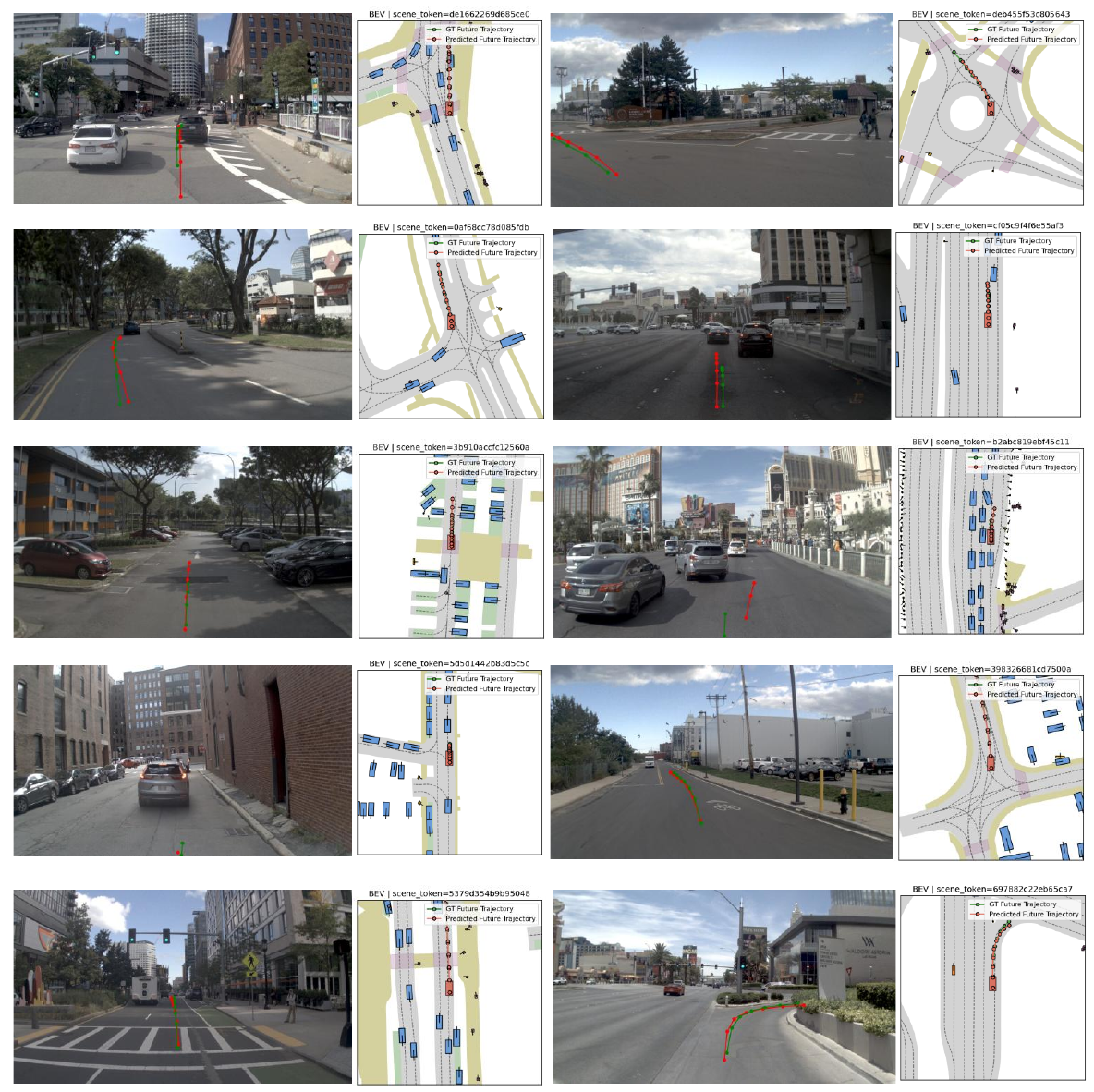}
    \caption{\textbf{Additional planning results in navhard subset.}
     \XFM{} consistently handles complex interactions, road geometries, and hazard situations, producing safe and goal-directed behaviors across diverse environments.}
    \label{fig:planning_appendix}
\end{figure}

The first term is the prediction mismatch of the original categorical action distribution, while the second term is the approximation introduced by mode selection and re-normalization. If the acceleration vocabulary is bounded by $\|\mathbf{v}_k\|_2\le V_{\max}$, then
\[
\|\hat{\mathbf{a}}_{\mathrm{mode}}-\mathbf{a}\|_2
\le
V_{\max}
\|q_\theta(\cdot \mid C_t)-p^\ast(\cdot \mid \mathbf{a})\|_1
+
V_{\max}
\|\tilde{q}_\theta(\cdot \mid C_t)-q_\theta(\cdot \mid C_t)\|_1.
\]
Using Pinsker's inequality for the first term gives
\[
\|\hat{\mathbf{a}}_{\mathrm{mode}}-\mathbf{a}\|_2
\le
V_{\max}
\sqrt{
2\mathrm{KL}
\left(
p^\ast(\cdot \mid \mathbf{a})
\Vert
q_\theta(\cdot \mid C_t)
\right)
}
+
V_{\max}
\|\tilde{q}_\theta(\cdot \mid C_t)-q_\theta(\cdot \mid C_t)\|_1.
\]

\begin{figure}[!h]
    \centering
    \includegraphics[width=\linewidth]{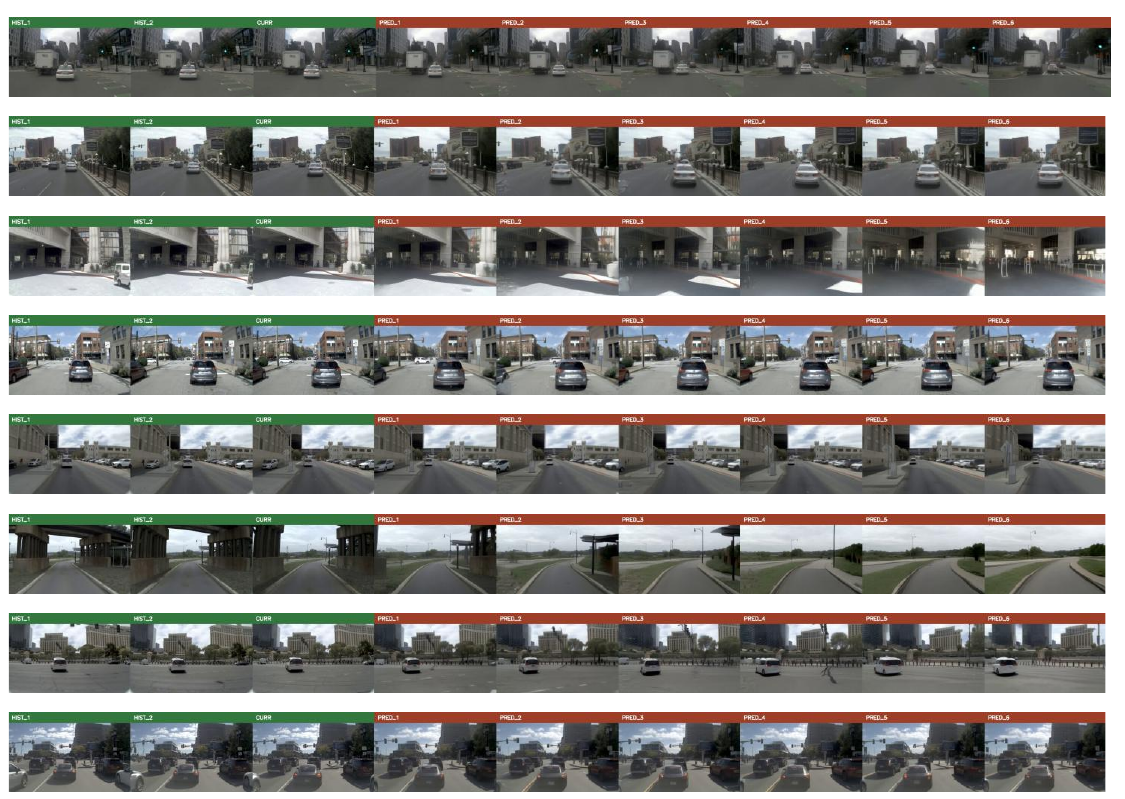}
    \caption{\textbf{Additional world model generation results.}}
    \label{fig:world_model_appendix}
\end{figure}

\paragraph{Implication for proposed supervision} This bound separates the original distribution prediction error from the additional mode-selection approximation. Unlike the soft-label interpolation result, mode-aware decoding is not claimed to be error-free. Instead, it trades the unbiased full-distribution expectation for a mode-conditioned acceleration that is more physically consistent for multi-modal action prediction.

\subsection{Additional Qualitative Results}
\label{additional_qualitative_results}

\paragraph{Planning}
We further visualize the iterative scheduling process of policy-token decoding. Specifically, \XFM{} is further evaluated on navhard, a curated subset of navtest consisting of challenging and safety-critical driving scenarios, as shown in Fig.~\ref{fig:planning_appendix}.
The qualitative results show that selective re-editing gradually reduces the uncertainty of the predicted action distribution across scheduling rounds. 
At early rounds, high-entropy tokens often appear around decision-sensitive or dynamically constrained segments, where multiple future actions may still be plausible. 
As scheduling proceeds, the entropy of these tokens decreases, indicating that the model progressively resolves ambiguous action choices and converges to a more stable policy.

Importantly, the entropy reduction is not achieved by blindly overwriting the entire action sequence. 
Instead, the scheduler selectively updates tokens that remain uncertain or distributionally unstable, while preserving tokens that have already become confident. 
This behavior is consistent with the quantitative results in Sec.~\ref{sec:analysis}: selective replacement improves with additional rounds, whereas full replacement may perturb stable tokens and degrade long-horizon trajectory accuracy. 
The visualization therefore provides qualitative evidence that the proposed scheduling strategy performs uncertainty-aware iterative refinement rather than repeated full-sequence resampling.

\begin{figure}[!h]
    \centering
    \includegraphics[width=\linewidth]{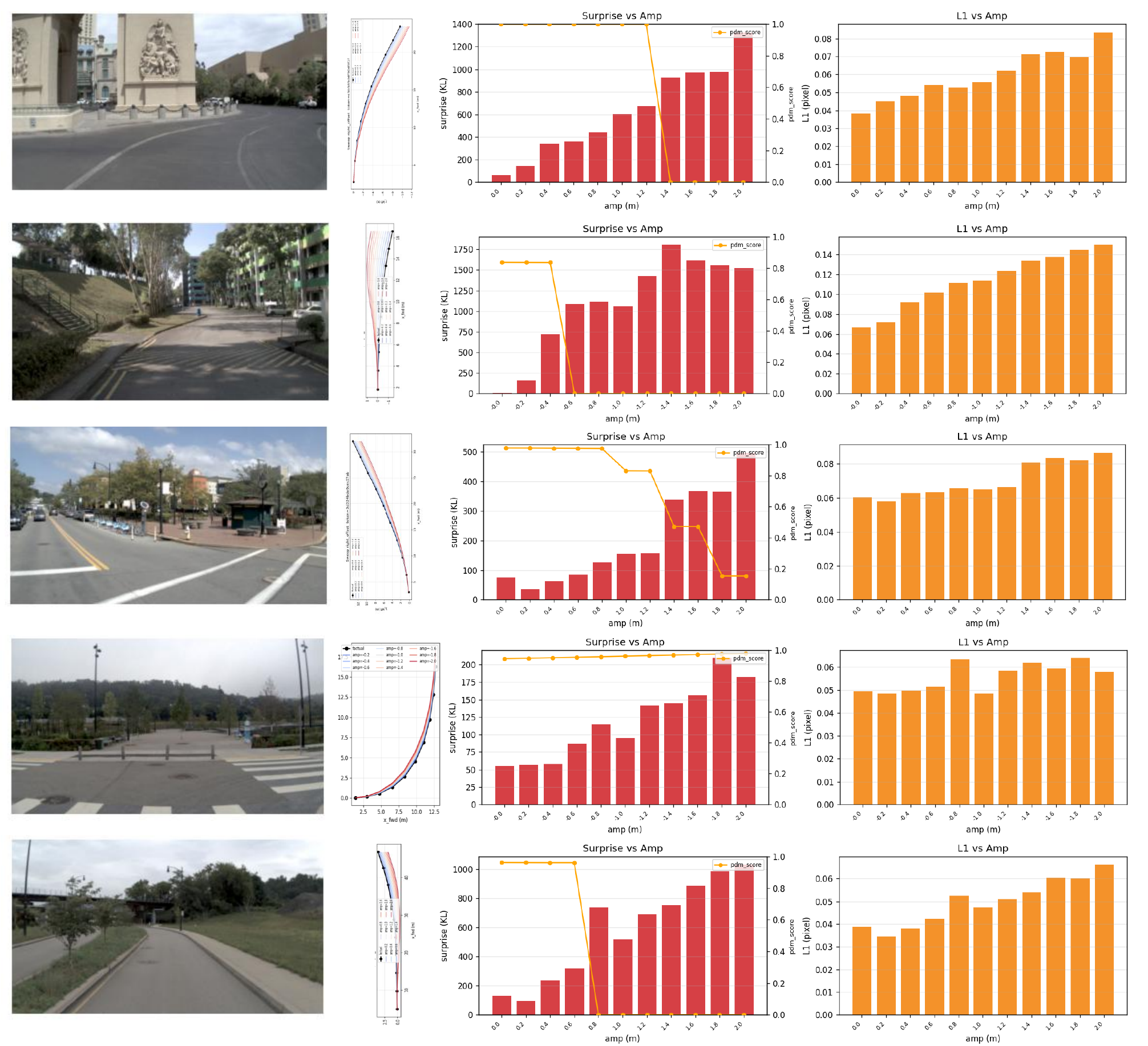}
    \caption{\textbf{Additional counterfactual result with surprise metric.}
    We compare surprise value $\mathbf{S}$ by both factual and counterfactual world-model generations by larger sweep of lateral counterfactual actions under havhard subset. A clear correlation persists of \XFM{} with causal understanding.}
    \label{fig:counterfactual_appendix}
\end{figure}

\paragraph{World modeling} Figure~\ref{fig:world_model_appendix} presents additional qualitative world-generation examples across a diverse set of urban driving scenarios. Given two historical frames and the current observation, Discrete-WAM generates future visual observations over multiple prediction horizons. Across intersections, urban roads, highway ramps, underpasses, and open-road environments, the generated futures remain temporally coherent and geometrically consistent with the underlying scene structure. In particular, Discrete-WAM accurately preserves static elements such as road boundaries, lane layouts, buildings, and traffic infrastructure, while simultaneously modeling the motion of surrounding vehicles and the ego-induced viewpoint changes.

\begin{figure}[t]
    \centering
    \includegraphics[width=\linewidth]{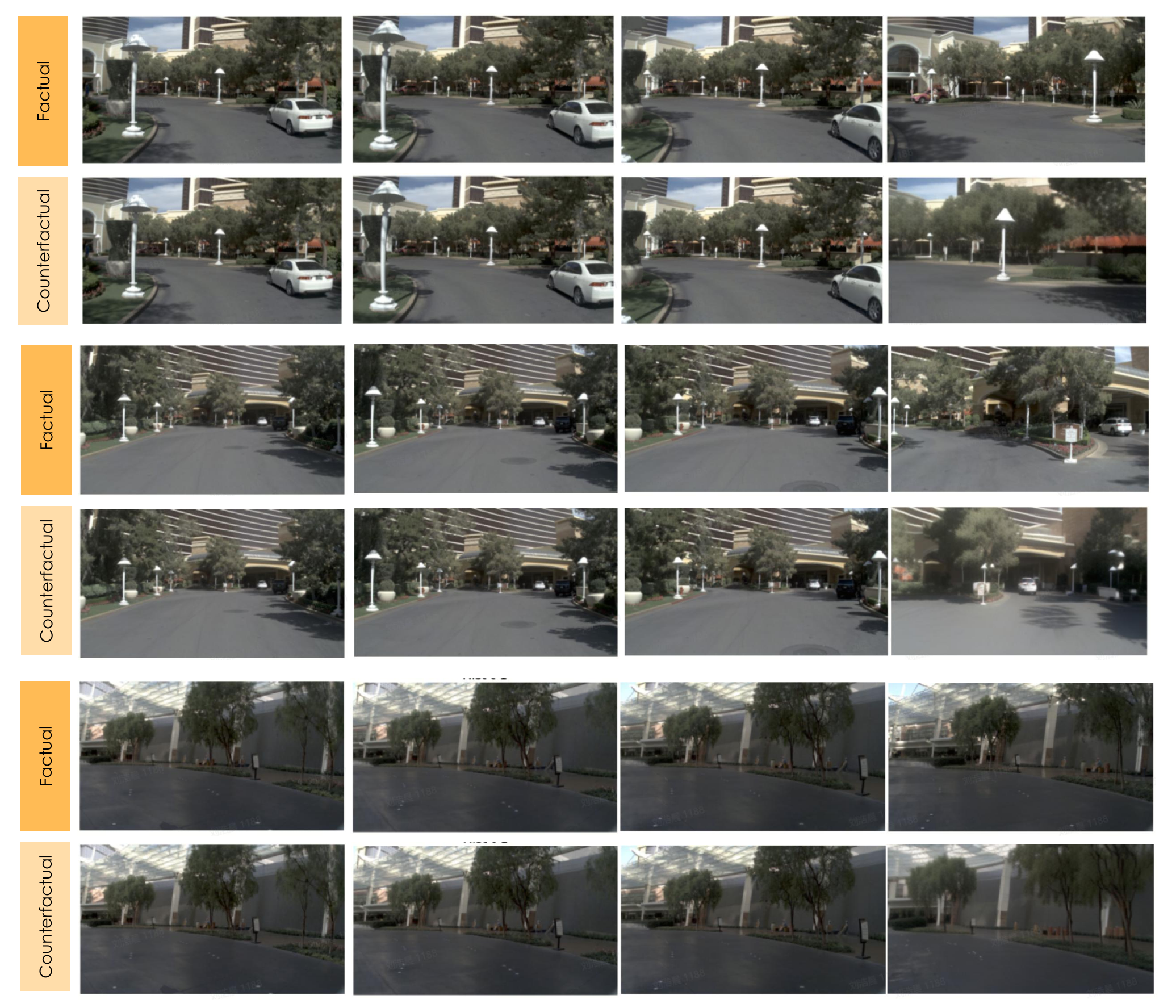}
    \caption{\textbf{Additional world model counterfactual prediction results.}}
    \label{fig:counterfactual_pred_appendix}
\end{figure}

\paragraph{Counterfactual inference}
Fig.~\ref{fig:counterfactual_appendix} presents additional counterfactual evaluations under diverse hazard scenarios by progressively increasing the magnitude of lateral action perturbations. Across all examples, the surprise metric exhibits a clear monotonic relationship with the severity of counterfactual interventions. When the perturbation remains small and the generated future is still physically plausible, the surprise value stays low and the planning score remains largely unaffected. As the counterfactual action increasingly deviates from the factual behavior, the generated future begins to violate scene constraints, leading to off-road behaviors, unsafe interactions, or imminent collisions. Correspondingly, the surprise value rises substantially while the PDM score drops sharply. We further observe that the pixel-level reconstruction error (L1) increases much more gradually than surprise, suggesting that surprise captures semantically meaningful deviations beyond low-level appearance differences. This consistent trend across multiple scenes indicates that the world model has learned action-conditioned causal dynamics rather than merely modeling visual statistics. The strong negative correlation between surprise and planning quality demonstrates that surprise can serve as an effective indicator of causal inconsistency and unsafe future outcomes in generated driving scenarios. Further generation results are provided in Fig.~\ref{fig:counterfactual_pred_appendix}.

\paragraph{Attention map visualization}
We provide additional attention map visualizations to complement the analysis in the main text. 
These examples include averaged policy attention maps across camera views, layer-wise front-view attention maps, and upper-region ablation results. 
They are intended to show that the observed attention patterns, including attention to driving-relevant semantics and stable activation in upper image regions, are not limited to a single example.

\begin{figure*}[t]
    \centering
    \includegraphics[width=1.0\linewidth,
                    height=0.95\textheight,
                    keepaspectratio,
                    trim=0 11cm 0 0,
                    clip]{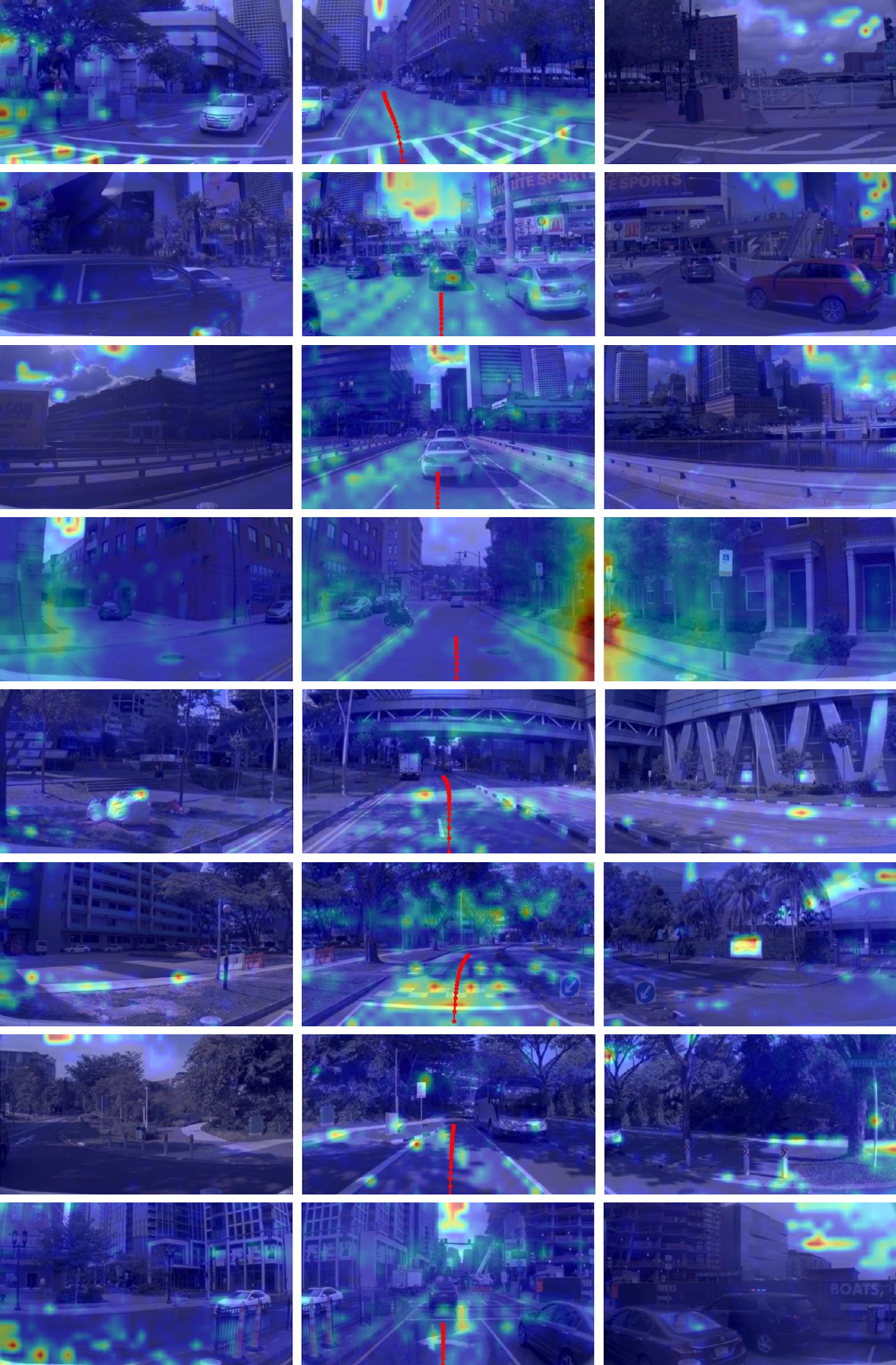}
    \caption{\textbf{Additional averaged policy attention maps.}
    Supplementary examples of policy attention averaged over Transformer layers, attention heads, and policy action queries from the first editing round. 
    The three panels in each example correspond to the left-view, front-view, and right-view cameras.}
    \label{fig:app_attn_avg}
\end{figure*}

\begin{figure*}[t]
    \centering
    \includegraphics[width=1.0\linewidth,
                    height=0.93\textheight,
                    keepaspectratio,
                    trim=0 5cm 0 0,
                    clip]{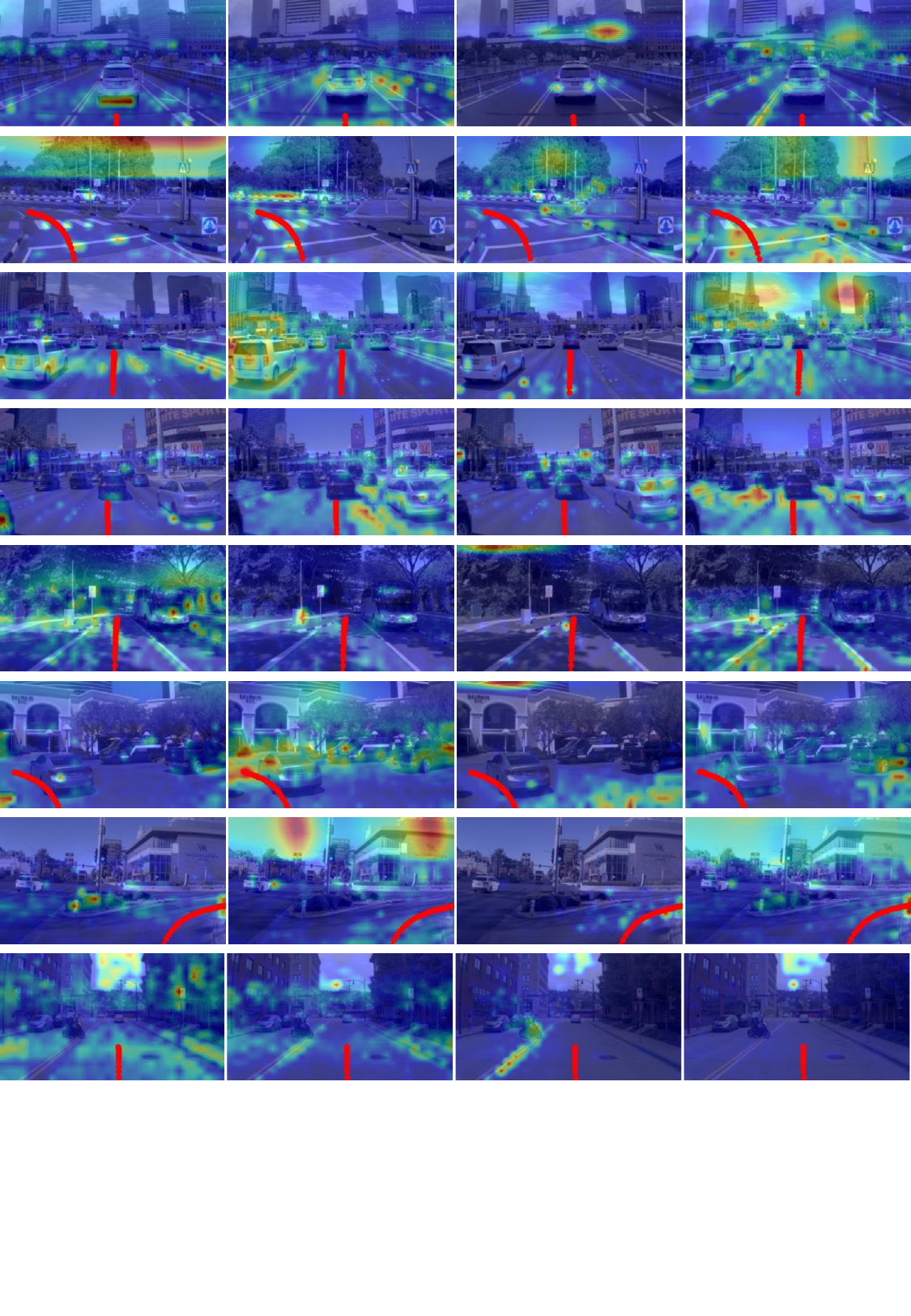}
    \caption{\textbf{Additional layer-wise policy attention maps.}
    Supplementary front-view examples of layer-wise policy attention. 
    Layers 0, 6, 12, and 17 are visualized by averaging over attention heads and policy queries.}
    \label{fig:app_attn_layers}
\end{figure*}

\begin{figure}[t]
    \centering
    \includegraphics[width=1.0\linewidth,
                    height=0.9\textheight,
                    keepaspectratio,
                    trim=0 1cm 0 0,
                    clip]{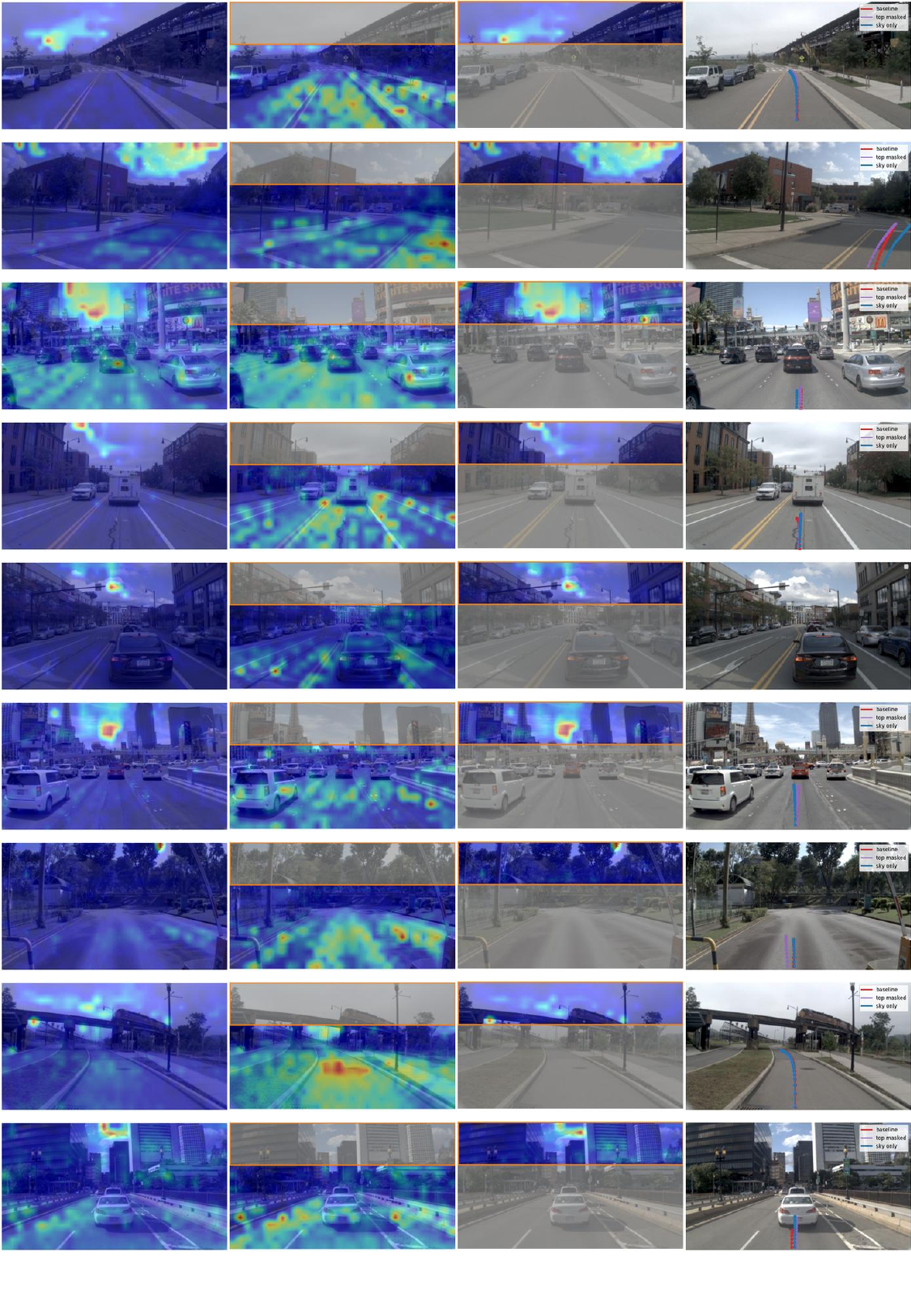}
    \caption{\textbf{Additional upper-region ablation examples.}
    Supplementary front-view examples comparing the original image, the upper-masked setting, and the upper-only setting. 
    These examples provide additional qualitative evidence for the attention redistribution behavior discussed in the main text.}
    \label{fig:app_attn_sky_ablation}
\end{figure}


\end{document}